\newif\ifcolm
\newif\ifcolmsubmission
\colmfalse 

\documentclass{article}

\usepackage{microtype}
\usepackage{graphicx}
\usepackage{subcaption}
\usepackage{booktabs} 

\usepackage{hyperref}

\usepackage{nicefrac}
\usepackage{xspace}
\usepackage{enumitem}
\usepackage{wrapfig}
\usepackage[ruled,vlined]{algorithm2e}
\usepackage[dvipsnames]{xcolor}

\ifcolm
  \usepackage[submission]{colm2025_conference}
\else
  \usepackage[accepted]{icml2025}
\fi

\usepackage{microtype}
\usepackage{hyperref}
\usepackage{url}
\usepackage{booktabs}

\usepackage{lineno}

\definecolor{darkblue}{rgb}{0, 0, 0.5}
\hypersetup{colorlinks=true, citecolor=darkblue, linkcolor=darkblue, urlcolor=darkblue}

\usepackage{amsmath}
\usepackage{amssymb}
\usepackage{mathtools}
\usepackage{amsthm}
\usepackage{arydshln}

\usepackage[capitalize,noabbrev]{cleveref}

\theoremstyle{plain}

\theoremstyle{definition}

\theoremstyle{remark}

\usepackage[textsize=tiny]{todonotes}
\usepackage{placeins}

\newcommand{\nbr}{noise-batch ratio\xspace}
\newcommand{\NBR}{Noise-Batch Ratio\xspace}

\newcommand{\users}{\ensuremath{N}}
\newcommand{\batch}{\ensuremath{B}}

\usepackage[scaled=0.95]{inconsolata}
\usepackage[T1]{fontenc}
\usepackage[scaled=0.95]{biolinum}
\newcommand{\dpsgd}{\textsf{DP-SGD}\xspace}
\newcommand{\dpadam}{\textsf{DP-Adam}\xspace}
\newcommand{\optupdate}{\textsf{OptimizerUpdate}\xspace}
\newcommand{\bert}{\textsf{BERT}\xspace}
\newcommand{\berttiny}{\textsf{BertTiny}\xspace}
\newcommand{\bertmini}{\textsf{BertMini}\xspace}
\newcommand{\bertsmall}{\textsf{BertSmall}\xspace}
\newcommand{\bertmedium}{\textsf{BertMedium}\xspace}
\newcommand{\bertbase}{\textsf{BertBase}\xspace}
\newcommand{\bertlarge}{\textsf{BertLarge}\xspace}
\newcommand{\bertmega}{\textsf{BertMega}\xspace}
\newcommand{\dpaccounting}{\textsf{dp\_accounting}\xspace}

\newcommand{\tradeoff}{compute-privacy-utility tradeoff\xspace}
\newcommand{\tradeoffs}{compute-privacy-utility tradeoffs\xspace}

\newcommand{\numsamples}{\ensuremath{N}}
\newcommand{\batchsize}{\ensuremath{B}}
\newcommand{\privacybudget}{\ensuremath{\epsilon}}

\newcommand{\modelsize}{\ensuremath{M}}
\newcommand{\iters}{\ensuremath{T}}
\newcommand{\slen}{\ensuremath{S}}

\newcommand{\computebudget}{\ensuremath{C}}
\newcommand{\nbratio}{\bar{\sigma}}
\newcommand{\mul}{\! \cdot \!}

\usepackage[]{mdframed}

\newcommand*{\eg}{e.g.,\@\xspace}
\newcommand*{\ie}{i.e.,\@\xspace}

\newcommand{\pmask}{\ensuremath{p_{\mathrm{mask}}}\xspace}
\newcommand{\mask}{\textsf{[MASK]}}
\newcommand{\xmask}{\ensuremath{\bar{\seq}}}
\newcommand{\masked}{\ensuremath{\mathcal{M}}\xspace}
\newcommand{\seq}{\ensuremath{\textbf{x}}}

\begin{document}

\ifcolm
    \title{Scaling Laws for Differentially Private Language Models}
    \maketitle
    \ifcolmsubmission
    \linenumbers
    \fi
\fi

\ifcolm
    {}
\else
\twocolumn[
    \icmltitle{Scaling Laws for Differentially Private Language Models}
    \icmlsetsymbol{equal}{*}
    
    \begin{icmlauthorlist}
    \icmlauthor{Ryan McKenna}{google}
    \icmlauthor{Yangsibo Huang}{google}
    \icmlauthor{Amer Sinha}{google}
    \icmlauthor{Borja Balle}{deepmind}
    \icmlauthor{Zachary Charles}{google}
    \icmlauthor{Christopher A. Choquette-Choo}{deepmind}
    \icmlauthor{Badih Ghazi}{google}
    \icmlauthor{George Kaissis}{deepmind}
    \icmlauthor{Ravi Kumar}{google}
    \icmlauthor{Ruibo Liu}{deepmind}
    \icmlauthor{Da Yu}{google}
    \icmlauthor{Chiyuan Zhang}{google}
    \end{icmlauthorlist}
    
    \icmlaffiliation{google}{Google Research}
    \icmlaffiliation{deepmind}{Google DeepMind}
    
    \icmlcorrespondingauthor{Ryan McKenna}{mckennar@google.com}
    
    \icmlkeywords{Differential Privacy, Scaling Laws, DP-SGD}
    
    \vskip 0.3in
]
    \printAffiliationsAndNotice{}  
\fi

\setlength{\textfloatsep}{4pt}

\begin{abstract}
Scaling laws have emerged as important components of large language model (LLM) training as they can predict performance gains through scale, and provide guidance on important hyper-parameter choices that would otherwise be expensive.
LLMs also rely on large, high-quality training datasets, like those sourced from (sometimes sensitive) user data. Training models on this sensitive user data requires careful privacy protections like differential privacy (DP). However, the dynamics of DP training are significantly different, and consequently their scaling laws are not yet fully understood.
In this work, we establish scaling laws that accurately model the intricacies of DP LLM training, providing a complete picture of the \tradeoffs and the optimal training configurations in many settings.
\end{abstract}
\section{Introduction}
Large language models (LLMs) are revolutionizing how we interact with technology, powering everything from instant translations and concise summaries to complex reasoning and creative content generation~\citep{achiam2023gpt,team2023gemini}.  
Training increasingly large models over increasingly large datasets has been a key driver of success for these LLMs, with frontier models being trained for millions of GPU-hours \cite{anil2023palm} and increasingly many trillions of tokens~\citep{team2024gemma,team2024gemma2}. Scaling laws for neural language models have been crucial because they provide a framework for understanding and predicting the performance gains achievable with increased computational resources, and importantly, guide the optimal allocation of that compute budget between model size and dataset size \cite{kaplan2020scaling,hoffmann2022training}.


The scale of data driving LLM progress also creates a critical privacy challenge.  State-of-the-art models train on massive, diverse datasets~\citep{dubey2024llama,team2024gemma} that are also distributed~\citep{carlini2024poisoning} making it difficult to exclude inadvertently shared personal information.  Paradoxically, user data, a key privacy concern, is also crucial for advancing LLM capabilities.  User interactions provide invaluable feedback for generating realistic synthetic data~\citep{microsoftdpsynthetic,googledpsynthetic} and aligning models with human values~\citep{stiennon2020learning}, reflecting real-world use cases better than web-scraped text.  However, direct training on sensitive user data is risky due to memorization and regurgitation~\citep{carlini2021extracting,carlini22quantifying,ippolito2022preventing,lukas2023analyzing,biderman2024emergent,prashanth2024recite}. This tension—the need for user data versus protecting user privacy—is addressed by differential privacy (DP)~\citep{dwork2006calibrating}.

While DP offers a principled solution to the tension between data utility and privacy in LLM training, applying it in practice, especially to large-scale models, presents significant challenges.  DP mechanisms like \dpsgd \cite{abadi2016deep} and its variants introduce computational overhead, implementation complexity~\citep{subramani2021enabling}, and utility degradation~\citep{bassily2014private}.  
While it is generally well-known that \dpsgd benefits substantially from training with very large batch sizes \cite{anil2021large,de2022unlocking,ponomareva2023dp}, little work has been done to understand the conditions under which this holds in compute-constrained settings, i.e., when an increase in batch size must be coupled with a decrease in model size or the number of iterations.  In part due to this reliance on large batch sizes, the largest models trained with DP today have hundreds of millions, rather than billions, of parameters \cite{anil2021large,li2022largelanguagemodelsstrong,DBLP:journals/corr/abs-2308-10888,DBLP:journals/corr/abs-2302-13861,charles2024fine}.

To train large models with DP, it is crucial to spend both the compute budget \emph{and} the privacy budget judiciously.  In this work, we pave the way towards training at the billion-parameter scale by initiating a study on the \emph{scaling laws of DP training}. To that end, we extend traditional scaling laws to consider a \tradeoff, accounting for intricacies and additional variables introduced by DP training.
Through a rigorous set of experiments, we empirically model this trade-off, and provide a thorough analysis of these experimental results to answer a number of scaling law-style questions, finding (among other things) that:

\begin{itemize}[leftmargin=*, nosep, itemsep=0.5pt]
    \item The compute budget allocation predicted by non-private scaling laws is far from optimal under DP, even for huge privacy budgets, confirming the need for our study.  
    \item However, we can accurately predict the optimal breakdown of the compute budget into model size, batch size, and iterations for virtually any privacy budget and dataset size. These compute-efficient training configurations save $5\times$ to $100\times$ compute compared to baseline configurations, while retaining comparable privacy and utility. 
    \item The optimal model size is typically at least an order of magnitude smaller with DP than without. This provides insight into the challenges of training large billion-parameter or larger language models with DP.
    \item In the DP setting, increasing the compute budget can sometimes yield little to no reduction in the loss unless accompanied by a corresponding increase in the privacy budget or dataset size.
\end{itemize}

\section{Preliminaries and Problem Setup}\label{sec:setup}

\setlength\intextsep{0pt}
\begin{wrapfigure}{r}{0.4321\linewidth} 
\footnotesize
\hspace{-1.5em}
\renewcommand{\arraystretch}{0.9}
\begin{tabular}{c|l}
    \toprule
        \textbf{Key} & \textbf{Definition} \\
    \midrule
        \privacybudget & Privacy budget \\
        \numsamples & Data budget \\ 
        C & Compute budget \\ 
        \batchsize & Batch size \\ 
        \iters & Iterations \\
        \slen & Sequence length \\
        \modelsize & Model parameters \\
        $\bar{\sigma}$ & \nbr \\
    \bottomrule
\end{tabular}
\end{wrapfigure}
Our dataset $\mathcal{D}$ consists of text sequences, where each individual contributes a single sequence $\seq = (x_1, \ldots, x_S)$ of $S$ tokens, and each token is drawn from a predefined vocabulary $\mathcal{V}$. We let $\numsamples$ denote the total number of individuals contributing to the dataset.


\paragraph{Masked Language Modeling.}

In this work we focus on the masked language modeling task \cite{devlin2018bert}, where each sequence has a chosen fraction \pmask of tokens masked out, \ie replaced with a special masking token \mask, uniformly at random. The goal is to predict the original token for each masked token using the entire context (bidirectionally). Let $\xmask$ represent the original sequence of tokens but masked using the above procedure and $\masked$ the ids of the masked tokens in $\xmask$. For a given parameter vector $\theta \in \mathbb{R}^{\modelsize}$, the language model defines a conditional probability $p_{\theta}(x_{j}~\mid~\xmask)$ for each $j \in \masked$, and the goal is to find $\theta$ to maximize the likelihood of all masked training tokens.

\paragraph{Differential Privacy.}

A randomized mechanism $\mathcal{A}$ satisfies \emph{$(\epsilon, \delta)$-DP}~\citep{dwork2006calibrating} if, for any two datasets $\mathcal{D}$, $\mathcal{D}'$ that differ by a single individual, all subsets $\mathcal{O}$ of possible outputs of $\mathcal{A}$ and $\epsilon >0, 0 \leq \delta < 1$:
\begin{equation}
    \Pr[\mathcal{A}(\mathcal{D}) \in \mathcal{O}] \leq e^{\epsilon} \Pr[\mathcal{A}(\mathcal{D}') \in \mathcal{O}] + \delta.
\end{equation}
\newpage

\paragraph{DP-SGD.} \dpsgd is a widely used algorithm to train neural networks with DP. It attains provable DP guarantees through limiting the contribution (\textit{sensitivity}) of each example by clipping its gradient to some $\ell_2$-norm (wlog, $1$), and then adding isotropic Gaussian noise to the averaged clipped gradients; see \cref{alg:user_dp_sgd} for pseudo-code. Our algorithm is a slight generalization of the original \dpsgd~\citep{abadi2016deep}: to enable adaptive optimizers, which are often crucial for training transformer models, the subroutine \optupdate can be any first-order optimizer.
Throughout this work, we set \optupdate to be Adam~\citep{kingma2015adam}, which we denote \dpadam. \cref{alg:user_dp_sgd} satisfies a formal DP guarantee that can readily be computed as a function of $\bar{\sigma}$, $\batchsize$, $\numsamples$, and $\iters$ using a suitable privacy accountant.  The \texttt{dp\_accounting} library provides functions that can efficiently and tightly compute the minimum value of $\bar{\sigma}$ as a function of $\epsilon$, $\delta$, $N$, and $B$ \cite{dp_accounting}. 

\textbf{\NBR.} Note that we parameterize \cref{alg:user_dp_sgd} in terms of the \emph{\nbr}  $\bar{\sigma}$, which is the standard deviation of noise added to the mean minibatch gradient, instead of the usual noise multiplier which typically added to the summed minibatch gradient.  While the noise multiplier is typically governs the privacy properties of the mechanism, the \nbr is a better proxy for the downstream learning performance. Specifically, there are two sources of variance in the stochastic gradient estimate $\tilde{g}$: (1) the minibatch estimate of the true population gradient and (2) the Gaussian noise added to ensure DP. Prior work has shown that the latter dominates the variance in most practical regimes~\citep{ponomareva2023dp}. 

\begin{algorithm}[t]
\caption{(Informal) Generalized \dpsgd.
\\
\cref{sec:dpsgd_discussion} discusses the informalities.}
\label{alg:user_dp_sgd}
\KwIn{Dataset $\mathcal{D}$, \nbr $\bar{\sigma}$, (expected) batch size $\batchsize$, iterations $\iters$}
\KwOut{Model parameters $\theta$.}
Initialize model parameters $\theta_0 \in \mathbb{R}^\modelsize$ \\
\For{$t = 1$ \KwTo $\iters$}{
    Select a (possibly random) size ${\approx}B$ minibatch $\mathcal{B}_t {\subset} \mathcal{D}$ \\
    $\bar{g}= \frac{1}{B} \sum_{\mathbf{x} \in \mathcal{B}_t} \text{clip}( \nabla \ell(\theta_{t-1}; \mathbf{x}))$ \\
    $\tilde{g} = g + \bar{\sigma} \mathcal{N}(0, 1)^\modelsize$ \\
    $\theta_t = \optupdate(\theta_{t-1}, \tilde{g})$
}
\textbf{return} $\theta_{\iters}$
\end{algorithm}

\subsection{Compute-Optimal DP Training}\label{sec:prelim-scaling-laws}
We are interested in empirically modeling how the \tradeoff changes as a function of the problem parameters.
We follow ideas used to model the compute-utility trade-off in the non-private setting \cite{kaplan2020scaling,hoffmann2022training}, but extend them to study the private setting by additionally considering the \emph{privacy budget} and \emph{data budget}. The key concepts are: 

\newpage
\begin{itemize}[leftmargin=*, nosep, itemsep=0.5pt]
  \item \textbf{Compute Budget} ($\computebudget$)  refers to the total floating point operations (FLOPs) required to train the model. We use the standard approximation of~\citet{kaplan2020scaling}: $6 \mul \modelsize \mul \batchsize \mul \slen \mul \iters$ to measure this, which is proportional to the model size (\modelsize) and the total number of training tokens ($\batchsize \mul \slen \mul \iters$). Note that unlike the non-private scaling laws, we use $\batchsize$ to represent the number of examples in a batch (not tokens) because this quantity is what matters for privacy calculations.
  This approximation provides a platform-independent estimate of compute requirements, and is justified further in \cref{sec:flops}.
  
  \item \textbf{Privacy Budget} ($\epsilon$) refers to the value of $\privacybudget$ at fixed $\delta$ in ($\privacybudget$,$\delta$)-DP. We fix $\delta = 10^{-8}=\Theta(1/\numsamples)$ unless otherwise mentioned, which is a common choice in the literature \cite{abadi2016deep,de2022unlocking}.
  \item \textbf{Data Budget} $(N)$ refers to the number of individuals in the training dataset, $|\mathcal{D}|$, which can be different than the number of examples processed by \dpsgd under multiple passes.
  {Note that our analysis and insights also hold in the more general setting where individuals can contribute multiple examples, although the data budget must still be interpreted as the number of individuals rather than the number of examples (see \cref{sec:user_level})}.
\end{itemize}

The privacy and data budgets are absent in most non-private scaling laws because they often assume that an infinite stream of data is available and no privacy protections are needed. In the private setting, model training is often constrained by both a fixed data budget (\ie a limited set of examples) and a fixed privacy budget (\ie $\privacybudget$ in DP). Both of these impact model training; thus, it is crucial to determine the optimal compute usage given the constraints on privacy and data, by fitting a scaling law accounting for this.


\subsection{Private Scaling Law Challenges}

\textbf{Additional Scaling Factors.} 
As mentioned above, our private scaling laws account for the additional data and privacy considerations not present in the non-private scaling law studies. 
These add complexity because DP adds noise beyond what is introduced through the stochasticity of training. Without DP, training with a batch size of $B$ for $T$ iterations is roughly equivalent to training with a batch size of $1$ for $B \mul T$ iterations, as long as $B$ is below the so-called ``critical batch size''~\citep{mccandlish2018empirical,shallue2019measuring,zhang2024does}. However, this relationship does not hold in DP settings, and further, DP training requires 
larger batch sizes to mitigate the impact of the added noise~\citep{anil2021large,de2022unlocking}. 

\textbf{Compute Requirements.} 
Even without DP, exhaustive hyperparameter tuning is infeasible for large models. DP training introduces further complexity with additional hyperparameters and the need to adapt standard defaults (e.g.,\ learning rate) to new regimes, necessitating careful protocol design to achieve near-optimal selection within reasonable compute. Further, it is important to consider that collapsing the privacy and data budgets to a single quantity is unlikely to provide generalizable insights.
\section{Private Scaling Law Methodology}
\label{sec:exp_setup}
In this section, we detail our methodology for estimating the validation cross-entropy loss from model size, noise-batch ratio, and training iterations, which in turn lets us estimate the utility under a fixed compute, privacy, and data budget.

\subsection{Decoupling Noise Calibration}
\label{sec:decoupling}

A key part of our methodology is to directly analyze the impact of the \nbr for a fixed but reasonably large \emph{physical batch size}, rather than indirectly through changes to the privacy budget or batch size.  Via \textit{post-hoc} accounting, we will predict what could happen at different \emph{hypothetical batch sizes}, an approach that is justified by the fact that typically the \nbr is the primary source of noise in the minibatch gradients, outweighing the noise due to minibatch sampling \cite{ponomareva2023dp}.

This decoupling enables for a better understanding of the underlying trade-offs. Without this approach, the non-linearities in DP accounting (detailed in \cref{sec:analysis}) make it difficult to assess these. We note that a naive methodology that tries to directly model the scaling law as a function of privacy budget (without going through the \nbr) would either provide less insight (by not generalizing across data budgets), or require much more compute.

After decoupling, the function we want to fit requires three inputs: the model size $\modelsize$, the number of iterations $T$, and the \nbr\footnote{The learning rate is a hyperparameter that is optimized over and not modeled directly.}. We require the function to be well-defined for a broad range of possible inputs that could be encountered in practical settings. We also need it to cover extreme points that may not be likely to be useful in practice, but may provide additional scientific insight. The methodology described below attempts to balance this need with the goal of using compute responsibly.

\subsection{Detailed Experimental Setup}

{\textbf{Models and Datasets.}} We train \bert models ranging in scale from Tiny (4M parameters) to Mega (778M parameters), summarized in \cref{table:models}.
We focus on the default \bert dataset, which includes  approximately 3.3B words \cite{zhu2015aligning, devlin2018bert} before tokenization. 
Each example is truncated or padded as necessary to a sequence of fixed length  $S=512$.\footnote{Future work could fruitfully consider other sequence lengths, as they are likely to showcase interesting trade-offs.}

\begin{table}[htbp]
  \caption{Models used in this study, taken from \citet{devlin2018bert}.}
  \label{table:models}
  \centering
  \small
  \begin{tabular}{lcccrc}
    \toprule
    \textbf{Model} & \textbf{Layers} & \textbf{Heads} & \textbf{Dims} & \textbf{Params} ($\modelsize$) \\
    \midrule
    \berttiny & 2 & 2 & 128 & 4.5M \\
    \bertmini & 4 & 4 & 256 & 11.4M \\
    \bertsmall & 4 & 4 & 512 & 29M \\
    \bertmedium & 8 & 8 & 512 & 41M \\
    \bertbase & 12 & 12 & 768 & 109M \\
    \bertlarge & 24 & 16 & 1024 & 335M \\
    \bertmega & 24 & 24 & 1536 & 778M \\
    \bottomrule
  \end{tabular}
\end{table}

{\textbf{Optimizer.}} We use \dpadam throughout. We use $1000$ steps of learning rate warm-up, followed by exponential learning rate decay, decreasing the learning rate by a factor of $10\times$ over a horizon of 128K iterations.
We use per-example clipping with an $\ell_2$ clip norm of $1.0$ across all experiments. We employ the normalized variant of clipping proposed by~\citet{de2022unlocking}, to help decouple learning rate tuning from clip norm. We verified that this setting effectively clips most per-example gradients, as recommended in prior work \cite{li2022largelanguagemodelsstrong, de2022unlocking}.

{\textbf{Learning Rates.}} We tune the learning rate with per-example gradient clipping but no noise, finding that the optimal learning rate is consistently $2^{-7}$ across all model scales. With noise, we consider three learning rates: $2^{-7}, 2^{-8}, 2^{-9}$. This methodological choice was based on early ablations that showed that when adding noise the optimal learning rate does decrease, but gradually so; see \cref{sec:learning_rates}.

{\textbf{Batch Sizes.}} We use a fixed physical batch size of $1024$ across all experiments. Via \textit{post-hoc} accounting, we will analyze what \emph{could} happen at different hypothetical batch sizes, under the assumption that cross entropy primarily depends on the privacy budget and batch size through the \nbr. We may expect this choice underestimates the benefit of larger batch sizes, a question we study empirically in \cref{sec:batch_size_ablation}.

{\textbf{\NBR.}} We consider $18$ values of \nbr: $\{2^{-k}~|~k=6, \dots, 23\}$, plus a baseline value of $0$ corresponding to non-private training.

{\textbf{Metrics.}} Every 100 training iterations, we record the average training loss over the previous $100$ iterations (or $102,400$ training examples).
Using training loss instead of evaluation loss is standard practice in scaling laws work, and is justified by the fact that we are training for less than a single physical epoch, so training loss is an unbiased estimate of evaluation loss. 

We provide details on the compute platforms and training throughput in \cref{app:training}.

\subsection{Semi-parametric Modeling}
\begin{figure*}[t]
\begin{subfigure}{0.33\linewidth}
\includegraphics[width=\textwidth]{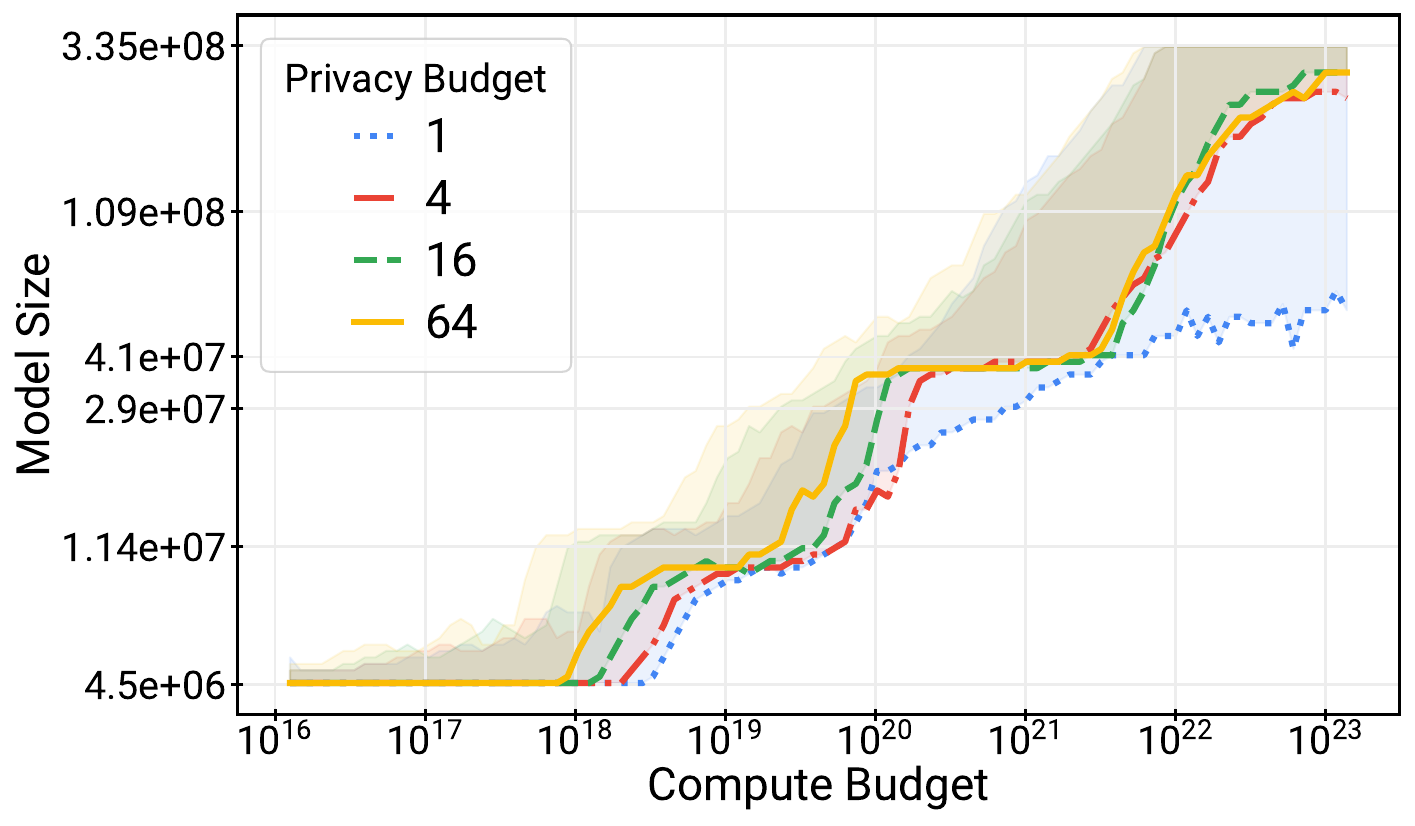}
\caption{Model Size}
\end{subfigure}
\begin{subfigure}{0.32\linewidth}
\includegraphics[width=\textwidth]{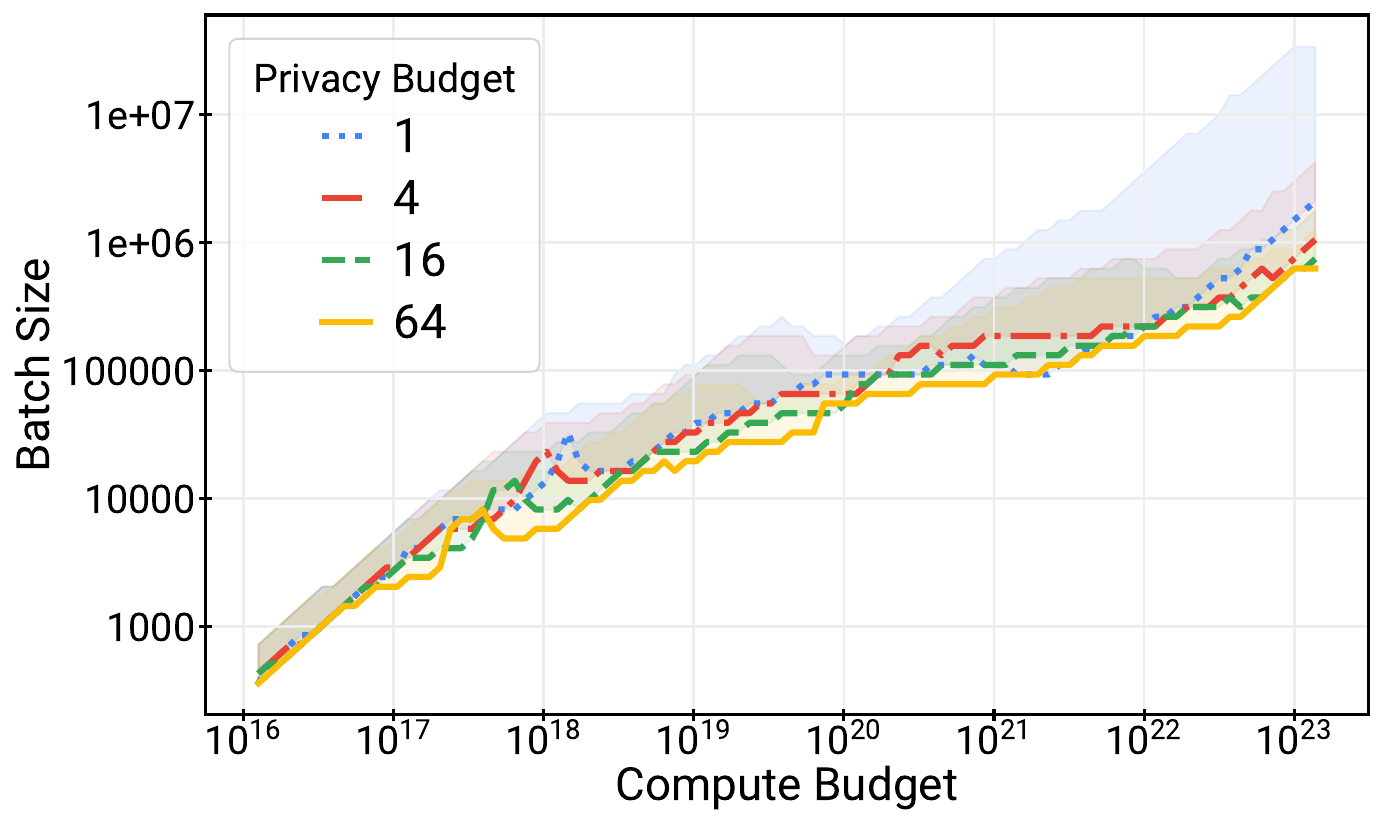}
\caption{Batch Size}
\end{subfigure}
\begin{subfigure}{0.32\linewidth}
\includegraphics[width=\textwidth]{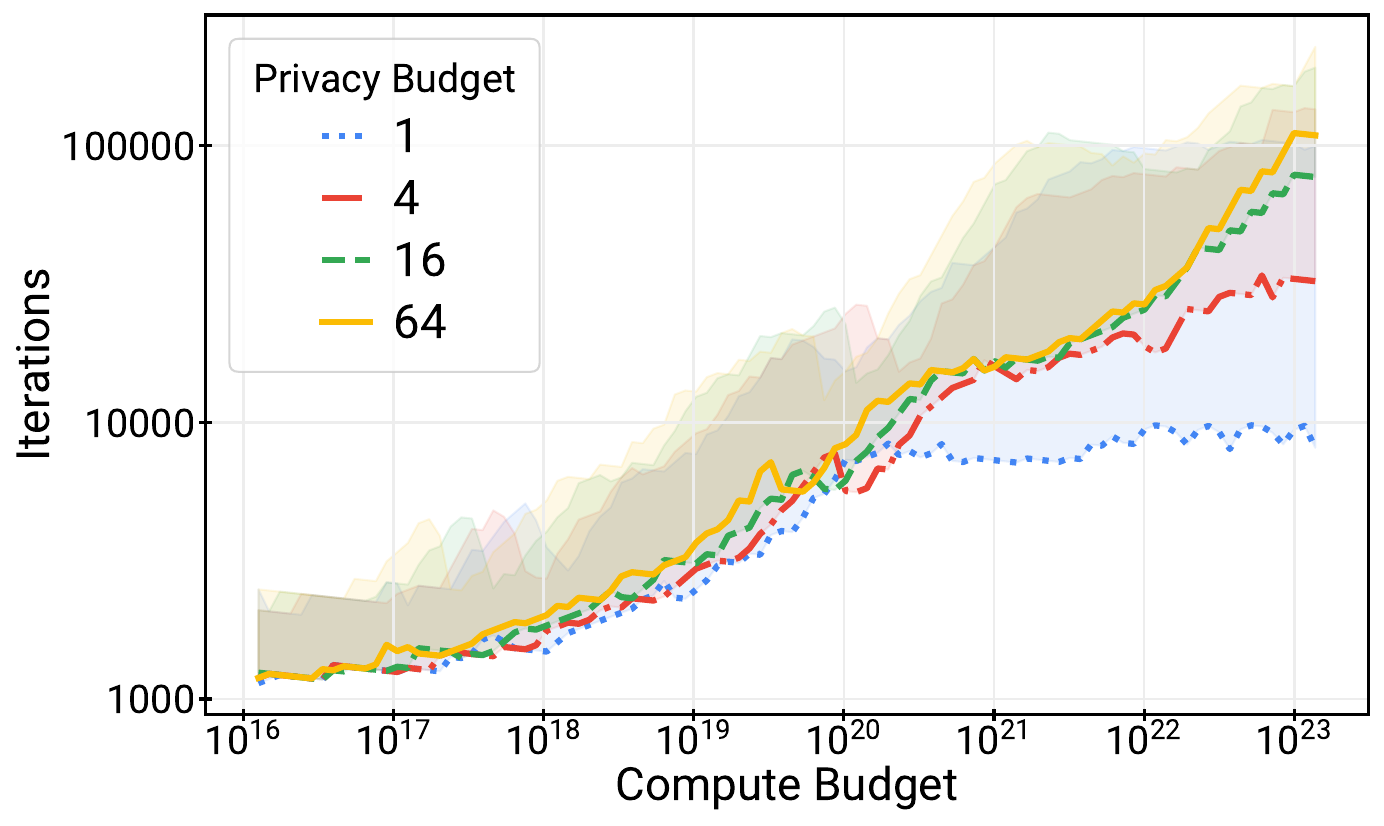}
\caption{Iterations}
\end{subfigure}
\caption{Optimal model size, batch size, and iterations for varying privacy and compute budgets, with a fixed data budget of $10^8$. Lines show minimum values for each hyper-parameter that achieve within 1\% of optimal cross-entropy for constant-compute training. Shaded regions indicate the full range of near-optimal settings.}
\vspace{-1.5em}
\label{fig:compute_allocation_selection}
\end{figure*}

After training the models described above, we obtain a grid of measurements over $6$ unique model sizes, $1280$ unique number of iterations, $18$ unique noise-batch ratios, and three learning rates. While one can directly query this data to answer a variety of interesting questions, we ultimately need to know what might happen in-between (and possibly outside of) the grid points we specifically evaluated. For that, we need to fit a function to the data, for which we follow a semi-parametric approach. See \Cref{sec:parametric-sl} for studies with fully parametric fits.

\paragraph*{Data Cleaning and Smoothing.}

First, we note that loss \emph{should} monotonically increase with increased \nbr, and monotonically decrease with increased iterations (unless training diverges), and we want our fitted function to capture this property. In practice, this invariant only holds approximately due to inherent variance in the training process. To clean the data, we apply the following post-processing steps:

\begin{enumerate}[leftmargin=*, nosep, itemsep=0.5pt]
  \item For each model size and \nbr, we apply a rolling average over the $10$ previous measurements to calculate a smoothed loss value. This corresponds to an average over $10 \mul 100 \mul 1024$ total examples, but does not perfectly preserve the expected invariant.
  \item For each model size and \nbr we apply \emph{isotonic regression} to ensure the $1280$ loss values are monotonically decreasing with respect to the number of iterations. For each model size and number of iterations, we apply isotonic regression again to ensure the $18$ loss values are monotonically increasing with respect to the \nbr. We do not enforce any monotonicity with respect to model size.
\end{enumerate}

We use isotonic regression to enforce desired monotonicity properties, rather than simpler alternatives like taking the cumulative $\min$ across each dimension. The latter approach suffers from a statistical phenomenon known as the \emph{minimum selection bias}, where one outlier sample can compromise the validity of the measurements. We visualize our smoothing process in \cref{sec:smoothing}.

\paragraph*{Training Step Extrapolation.}

Next, we extrapolate our smoothed data with respect to the number of iterations, by fitting a parametric form to the training curve and predicting where the loss would have gone if training continued beyond 128K iterations. We use a simple parametric form inspired by \citet{hoffmann2022training}, namely $ L = E + \frac{A}{T^\alpha} $. We fit this function using \texttt{scipy.optimize.curve\_fit}, which uses the Levenberg--Marquardt algorithm to solve a nonlinear least squares problem \cite{nocedal1999numerical}.  We independently fit a function for each model size and \nbr on data from iterations $16K$ to $128K$.

\paragraph*{Scaling Law Fitting.}
After data cleaning, our goal is to fit a function $L(\modelsize, T, \nbratio)$ that estimates the loss under a $\modelsize$-parameter model training for $\iters$ iterations with a \nbr of $\nbratio$. We fit this function using linear interpolation, and specifically \texttt{scipy.interpolate.RegularGridInterpolator} in Python. Since $\modelsize$, $\iters$, and $\nbratio$ are all naturally varied in log-space, we apply interpolation to the function $F$ such that $F(\log{\modelsize}, \log{T}, \log{\nbratio}) \coloneqq L(\modelsize, T, \nbratio)$ instead. This function is well-defined for any $T$ and any $\modelsize, \nbratio$ within the range of experimental settings considered; that is, $ \modelsize \in [4.5\mathrm{M}, 784\mathrm{M}], \nbratio \in [0.5^{23}, 0.5^{6}]$. Because we use interpolation, our fitted function matches the smoothed data exactly at the evaluation points, and approximates it in between them.  In \cref{sec:parametric-sl} we also fit a parametric form for this function as well, finding that it is largely consistent with the non-parametric fit.

\subsection{Using the Fitted Functions} \label{sec:combining}
\begin{figure}[t]
  \centering
  \includegraphics[width=0.9\linewidth]{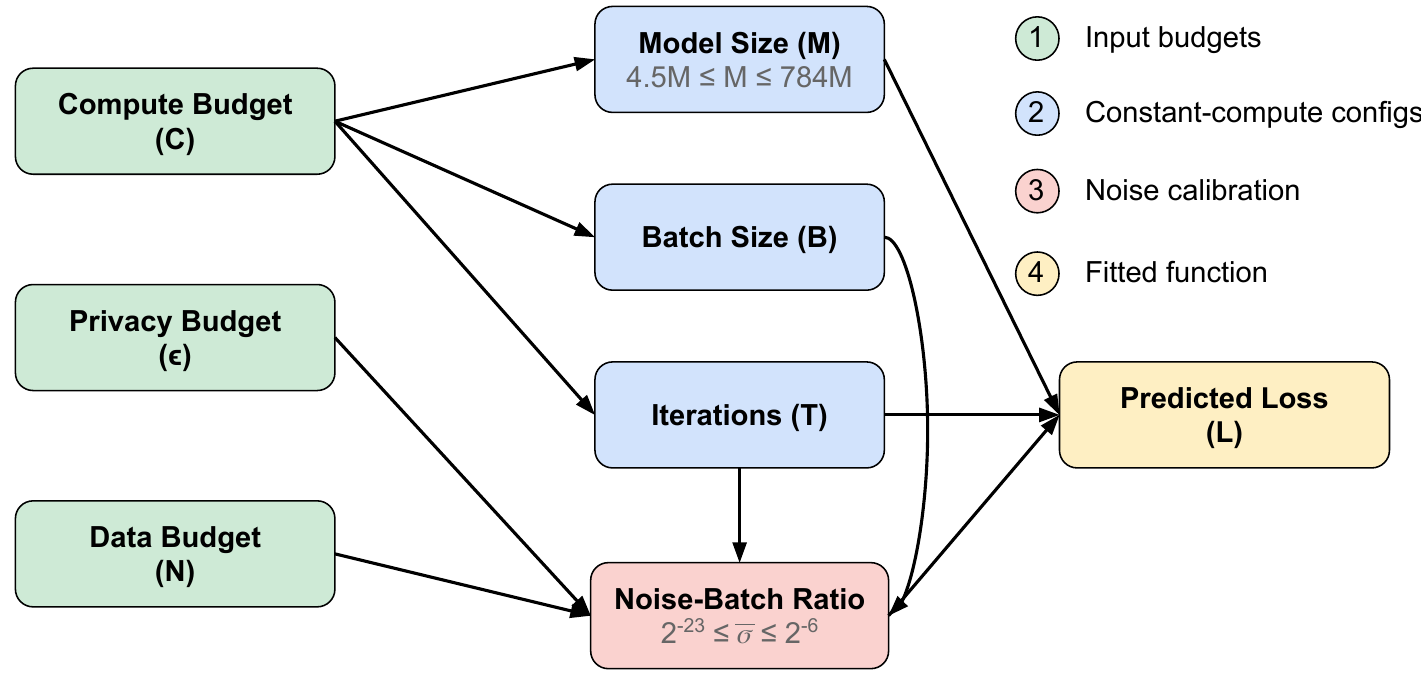}
  \caption{Workflow for estimating cross entropy of different training configurations under given compute, privacy, and data budgets.}
  \label{fig:methodology}
\end{figure}

We are now able to answer DP scaling laws questions. \Cref{fig:methodology} summarizes our approach. We begin with inputs: the compute budget, privacy budget, and data budget. Second, we proceed by enumerating an exhaustive set of constant-compute training configurations; i.e., combinations of model size, batch size, and iterations that require the given compute budget. Using privacy accounting and noise calibration functions from the \dpaccounting library, we compute the \nbr as a function of the privacy budget, data budget, iterations, and (expected) batch size. Finally, we query our fitted function with this \nbr, along with the given model size and number of iterations, giving us a final estimate of the cross entropy of these training configurations. In addition, we can also specify directly the training configurations instead of the compute budget for the purposes of conducting specific ablations or comparisons.

\section{Experimental Findings of Scaling Laws}

\subsection{Optimal Compute Budget Allocation} \label{sec:compute_budget_allocation}
We first determine how to best utilize our compute budget in different situations.  Specifically, for a given compute/privacy/data budget, we aim to understand how to optimally allocate our compute budget among the model size, batch size, and number of iterations. Additionally, we seek to understand how the optimal allocation changes per budget. While this question can be answered for virtually any setting of the budgets with the data we collected, we visualize a few relevant slices of the data in \cref{fig:compute_allocation_selection}. More comprehensive results can be found in \cref{sec:appendix:extra_compute_budget_allocation}. From this visualization, we make the following observations:
\begin{figure*}[htbp]
\begin{subfigure}{0.3\linewidth}
\includegraphics[width=\textwidth]{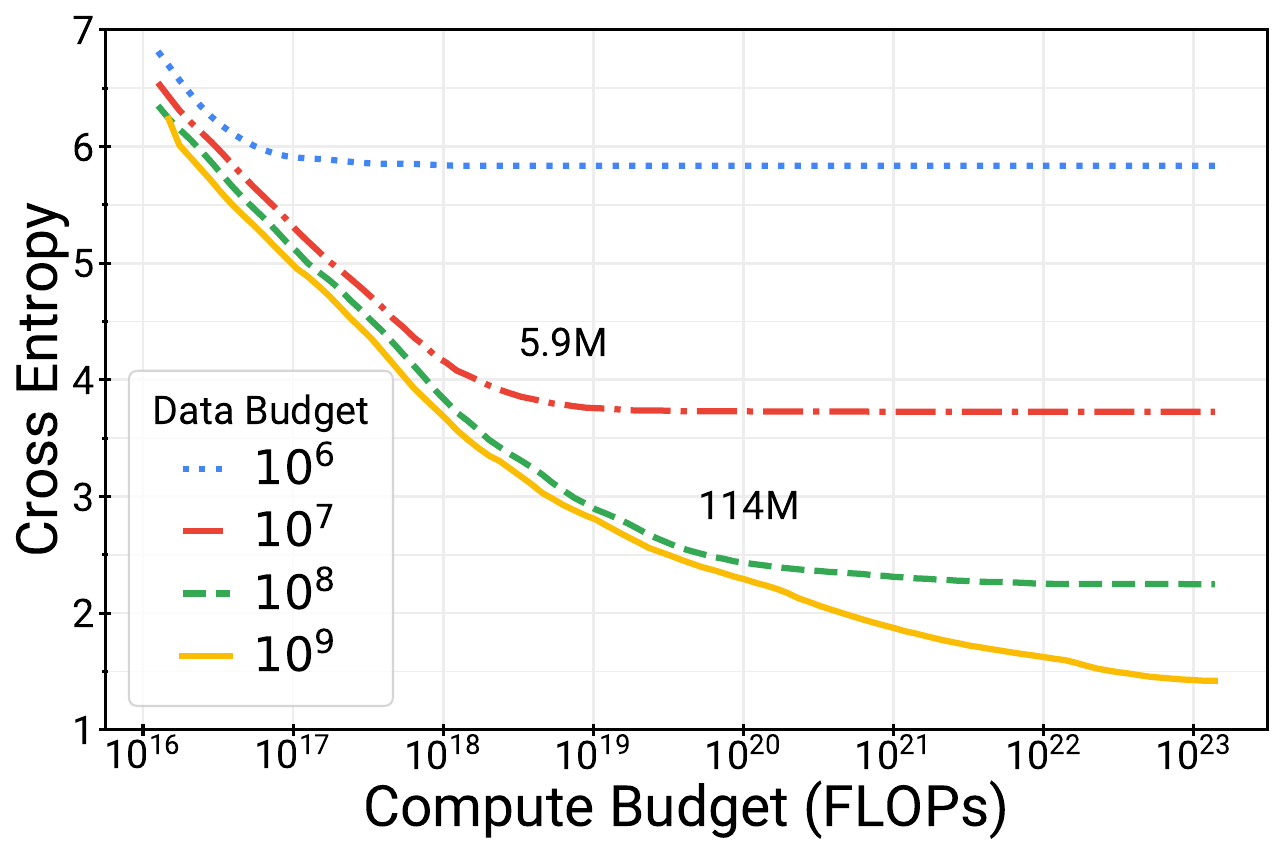}
\caption{Privacy Budget: $\epsilon=1$} \label{fig:vary_compute_and_privacy}
\end{subfigure}
\hfill
\begin{subfigure}{0.285\linewidth}
\includegraphics[width=\textwidth]{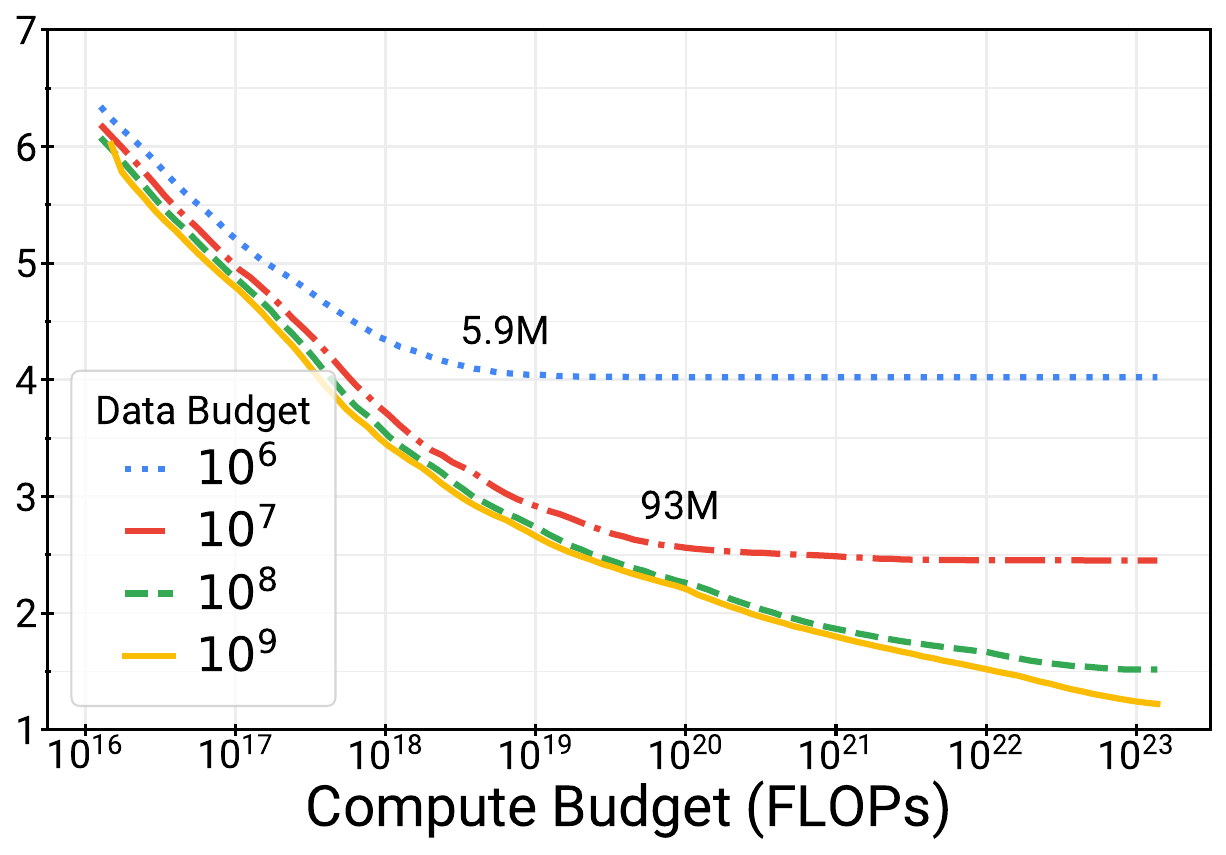}
\caption{Privacy Budget: $\epsilon = 8$} \label{fig:vary_compute_and_privacy2}
\end{subfigure}
\hfill
\begin{subfigure}{0.32\linewidth}
\includegraphics[width=\textwidth]{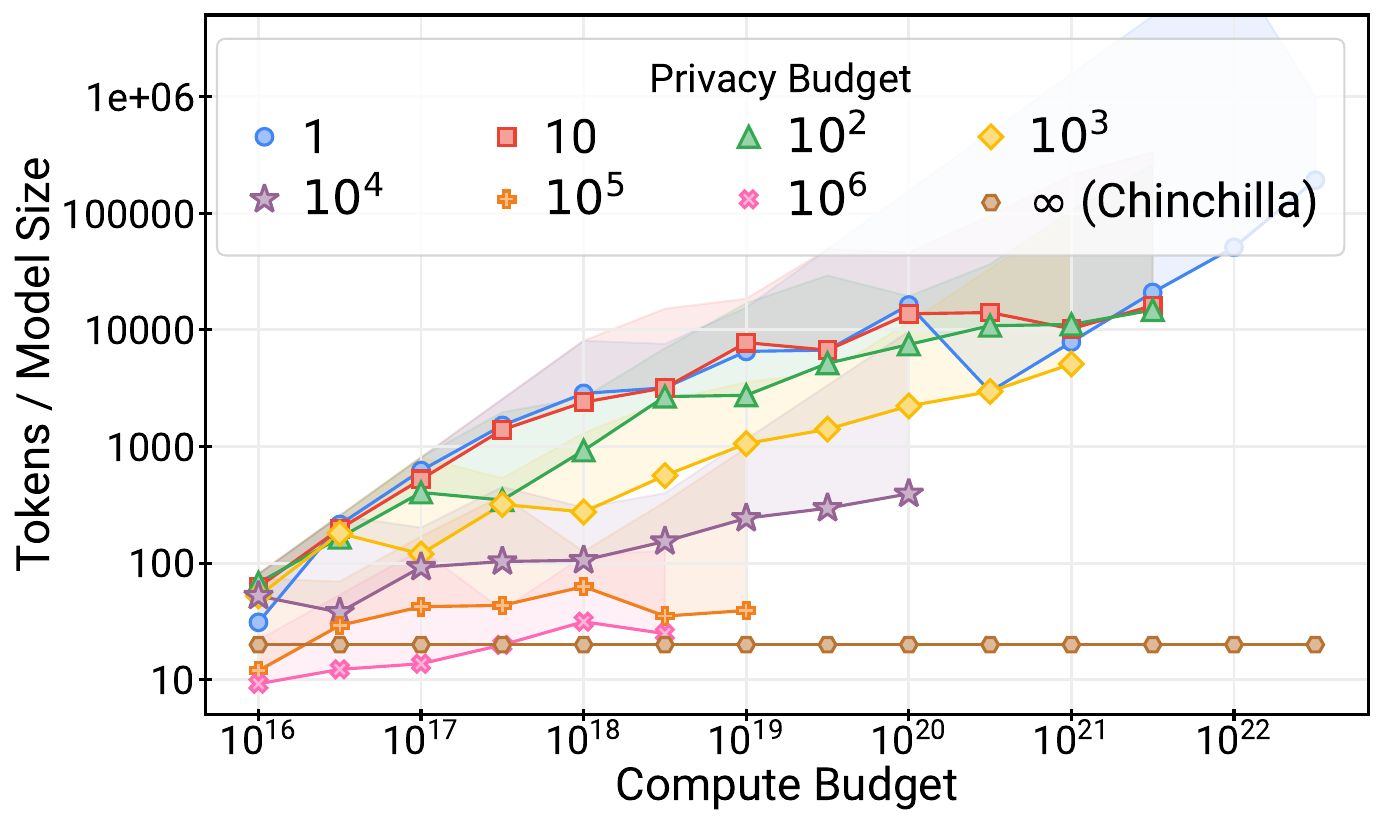}
\caption{Token-to-Model Ratio} \label{fig:token_model_ratios}
\end{subfigure}
\caption{(a-b) Best cross-entropy loss achieved for varying compute budgets, four data budgets, and two different privacy budgets. Each figure is annotated with the optimal model size at the inflection point for two of the curves. (c) Number of training tokens $S \mul B \mul T$ divided by number of model parameters for the compute-optimal training configuration, fixing the data budget to $N=10^7$.}
\end{figure*}

\begin{itemize}[leftmargin=*, nosep, itemsep=0.5pt]
  \item For small compute budgets, the optimal allocation of compute budget does not exhibit a strong dependence on $\epsilon$. However, there is a small but consistent trend that with larger privacy budgets, one should train a larger model with a smaller batch size and for more iterations than one would train with a smaller privacy budget. This finding is somewhat surprising, since as the privacy budget gets larger, the point at which increasing batch size leads to diminishing returns in terms of \nbr increases roughly according to $\approx N \sqrt{\nicefrac{\epsilon}{T}}$ \cite{ponomareva2023dp}.
  \item There are many settings of model size, batch size, and number of iterations that achieve near-optimal loss, as indicated by the large shaded regions. This suggests some amount of robustness for compute-optimal training hyperparameters. All else being equal, training smaller models on more tokens should generally be preferred due to their inference-time efficiency advantages.
  \item Optimal model sizes are much smaller than predicted by non-private scaling laws. For instance, at $10^{22}$ FLOPs, $\sim 10^8$ parameters are compute-optimal, compared to $\sim 10^{10}$ non-privately.
\end{itemize}

\subsection{Benefits of Increased Compute}

We now aim to understand and measure how much benefit increased compute budgets can provide and \emph{when} it can provide it. In \cref{fig:vary_compute_and_privacy}, we look at how the optimal achievable cross entropy depends on the compute budget for different settings of data/privacy budget. Our main observations are:
\begin{itemize}[leftmargin=*, nosep, itemsep=0.5pt]
  \item Increasing the compute budget can be a very effective strategy for reducing cross entropy under a fixed privacy/data budget up to a limit, but there is an inflection point where increasing the compute budget beyond this point provides little to no benefit. The ``critical compute budget'' where this inflection point occurs increases with both privacy budget and data budget. For example, with a data budget of $10^8$ and a privacy budget of $1$, the best cross entropy is achieved with a compute budget $\gtrsim 10^{20}$ and corresponds to a model with $114M$ parameters. This is a qualitatively different behavior than non-private scaling laws, where increasing the compute budget continues to provide benefits even at the extreme scales. 
\end{itemize}

More comprehensive analysis of the saturating compute budget for a representative set of data and privacy budgets can be found in \cref{sec:saturating_compute}.

\begin{figure*}[htbp]
\centering
    \begin{subfigure}{0.28\linewidth}
    \includegraphics[width=\textwidth]{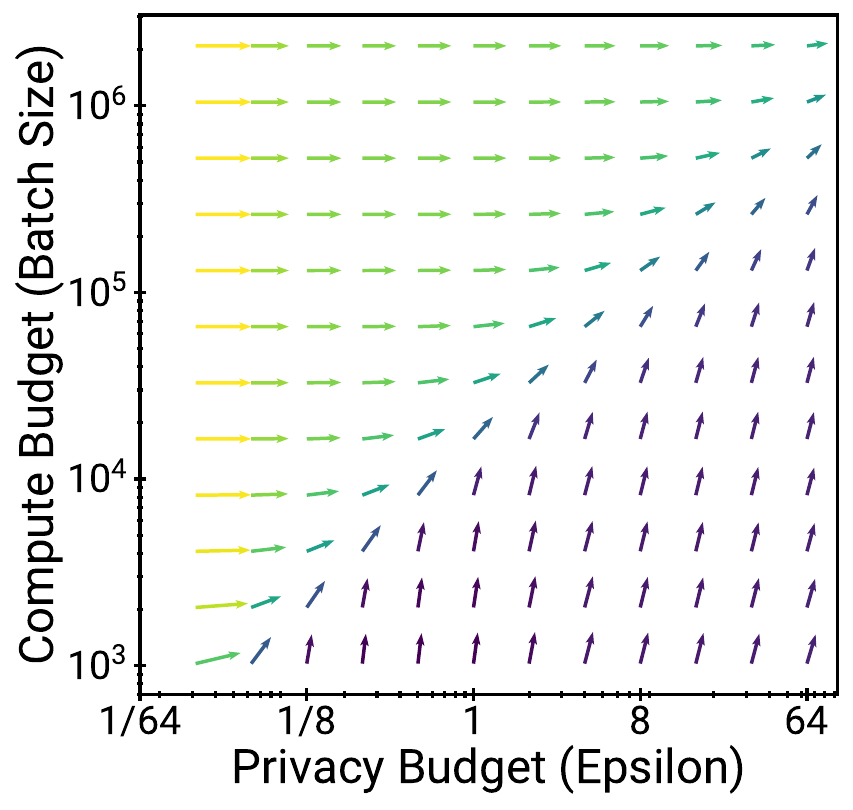}
    \caption{Data Budget: $N = 2^{24}$} \label{fig:accounting_privacy_vs_compute}
    \end{subfigure}
    \hspace{1em}
    \begin{subfigure}{0.28\linewidth}
    \includegraphics[width=\textwidth]{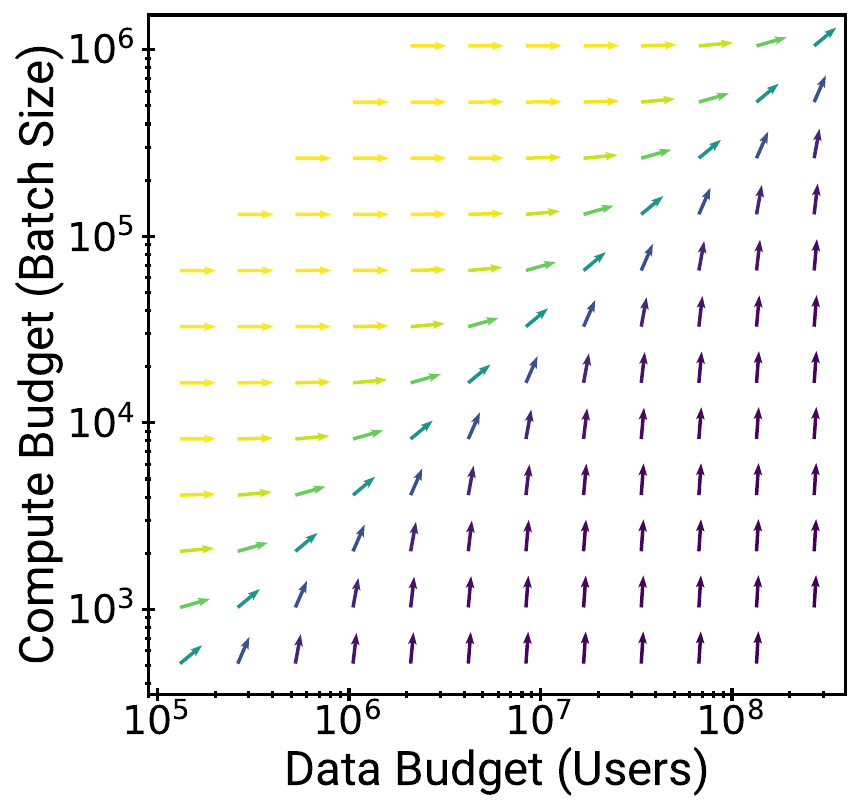}
    \caption{Privacy Budget: $\epsilon = 4$} \label{fig:accounting_data_vs_compute}
    \end{subfigure}
    \hspace{1em}
    \begin{subfigure}{0.28\linewidth}
    \includegraphics[width=\textwidth]{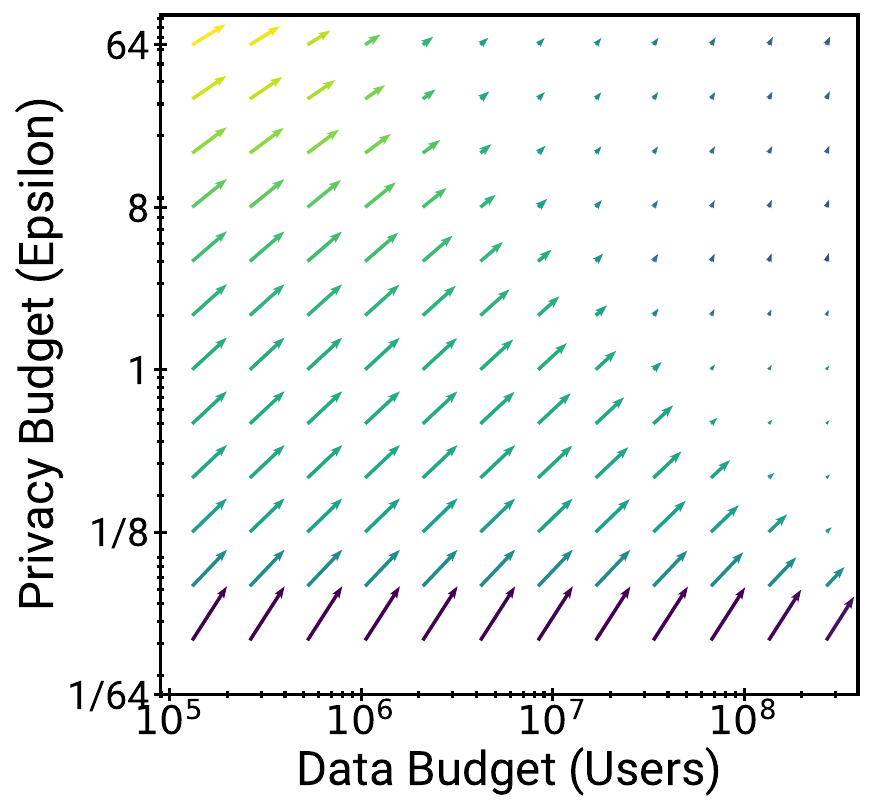}
    \caption{Batch Size: $B = 65536$} \label{fig:accounting_data_vs_privacy}
    \end{subfigure}
\caption{Marginal benefits of increasing the privacy budget ($\epsilon$), compute budget ($B$), and data budget ($N$) on the \nbr.}
\label{fig:vector_fields}
\end{figure*}

\subsection{Token-to-Model Ratio}

We now aim to understand more about compute-optimal training configurations, specifically the ratio of the number of training tokens (as measured by $S \mul \batch \mul T$) to model size and privacy budget. In other words, we study a form of sample complexity. In the absence of DP, a constant token-to-model ratio of $20\times$ is the recommended best practice~\citep{hoffmann2022training}. As we see in \cref{fig:token_model_ratios}, the behavior under DP is not as simple:

\begin{itemize}[leftmargin=*, nosep, itemsep=0.5pt]
  \item The token-to-model ratio increases with compute budget, especially for smaller privacy budgets. As the privacy budget increases, the slope decreases, and for a sufficiently large privacy budget becomes nearly flat as predicted by the prior work. However, the privacy budget required to exhibit behavior similar to prior work is \emph{extremely large}. Note that a privacy budget of $\epsilon = 1000$ provides no meaningful formal membership inference protection.\footnote{However, values even larger than this have been shown to be effective against reconstruction attacks in prior works \citep{balle2022reconstructing, kaissis2023bounding, ziller2024reconciling}.} 
  Nonetheless, the noise added still has a significant impact on training: its behavior in \cref{fig:token_model_ratios} is more similar to a privacy budget of $1$ than non-private training ($\epsilon = \infty$). 
  \item For moderate privacy budgets in the range of $[1, 10]$, a good token-to-model ratio is typically between $1000$ and $100000$, although for sufficiently large compute budgets, it can go beyond this point. This connects back to an earlier observation that even with infinite compute, there is eventually no benefit to increasing the model size when using a modest privacy budget. These ratios roughly correspond to training models $10\times$ to $50\times$ smaller than predicted by \citet{hoffmann2022training}.
\end{itemize}

\subsection{Comparison Against Baselines}

\begin{figure}[htbp]
  \centering
  \includegraphics[width=0.8\linewidth]{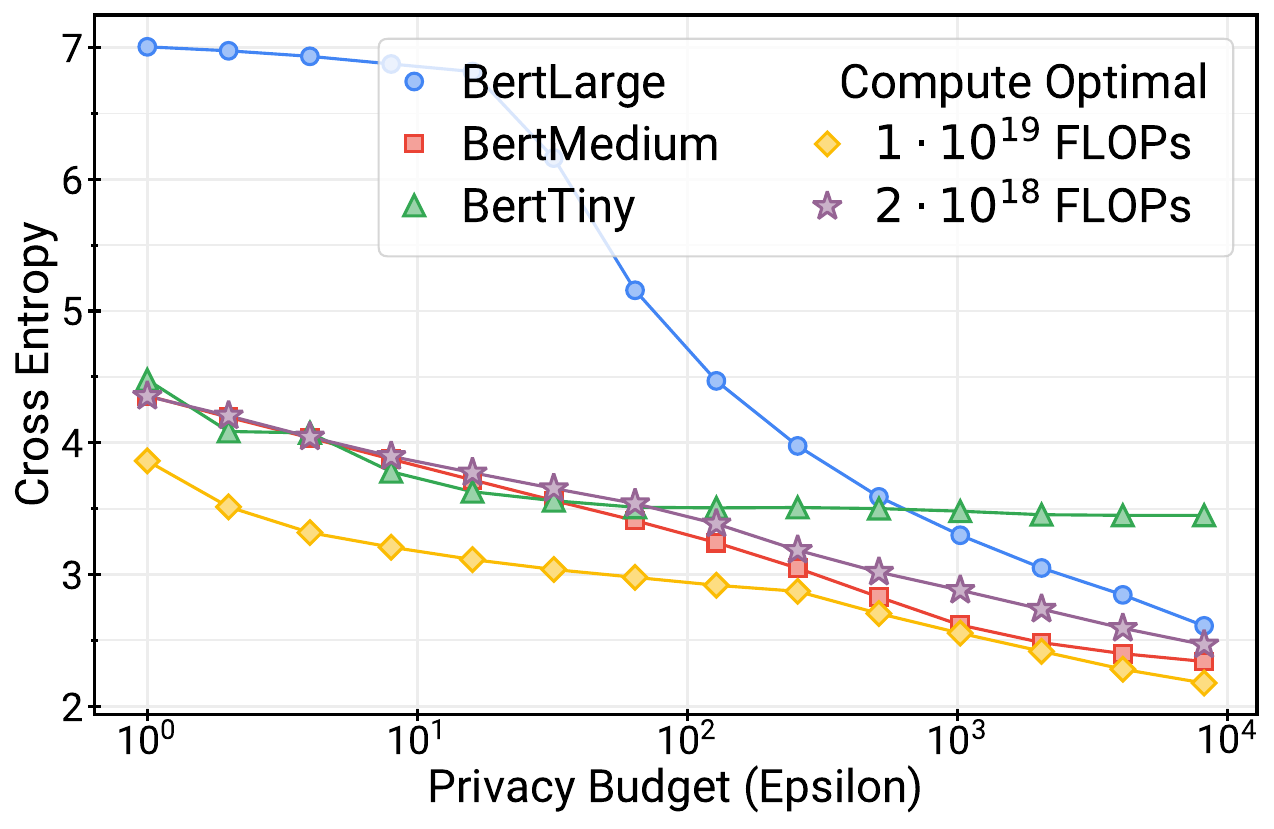}
  \caption{Comparison of a compute-optimal training configuration to some natural baselines as a function of the privacy budget. All models are trained with a compute budget of $10^{19}$ FLOPs and a data budget of $N=10^7$ respectively.}
  \label{fig:baseline_compare}
\end{figure}

We now measure the improvement our compute-optimal training configurations provide over natural baselines. In the DP training literature, it is common to fix the training configuration (model, iterations, batch size), and vary the privacy budget. To that end, we consider 3 baseline training configurations: \bertlarge trained for $7500$ steps with a batch size of $1295$, \bertmedium trained for $5000$ steps with a batch size of $15879$ and \berttiny trained for $2500$ steps with a batch size of $283061$. In all three, we fix the data budget to $\users = 10^7$. Each of these training configurations require $10^{19}$ FLOPs. The first configuration is close to what would be predicted by non-private scaling laws~\citep{hoffmann2022training}, while the last might be selected by an expert in DP who recognizes the importance of large batch sizes. The results are shown in \cref{fig:baseline_compare}, from which we find:

\begin{itemize}[leftmargin=*, nosep, itemsep=0.5pt]
  \item For most privacy budgets, the training configuration predicted by non-private scaling laws (\bertlarge) yields very low utility. While utility improves for sufficiently large privacy budgets, this suggests that private scaling laws are fundamentally distinct from non-private ones.
  \item The optimal training configuration changes with the privacy budget, and naively using a fixed training configuration across all privacy budgets, as is common in the literature, leaves significant utility on the table.
  \item Compute-optimal training can either give better utility, or save compute/privacy budget under fixed utility. Training a compute-optimal model with $2 \times 10^{18}$ FLOPs yields similar utility as the best baseline models with $5\times$ the FLOPs for the reasonable range of privacy budgets. This is just one instructive example. The savings in other settings may change depending on factors like data budget, compute budget, and quality of the baseline training configurations (\eg the compute savings over \bertlarge exceeds $100\times$, although this is not shown).
\end{itemize}

\subsection{Synergy between Privacy/Data/Compute Budgets} 
\label{sec:analysis}

While many of the trade-offs that we explore in this work are data-dependent and require significant empirical investigation, many generalizable scaling insights can be derived purely by exploring privacy accounting. In this section we detail some of these, which corroborate many of our experimental evidence above and require very little compute. These insights are domain-agnostic, and therefore likely to generalize to other machine learning settings beyond language models, while also helping us understand and explain some of the experimental observations presented earlier.

We analyze how the \nbr behaves as a function of privacy budget (as measured by $\epsilon$), compute budget (as measured by $\batch$), and data budget (as measured by \numsamples). 
We fix $T=16000$ training steps here, but our findings hold for any fixed number of steps\footnote{While compute budget could also be varied through $T$, the effect of changing $T$ is data-dependent and the noise batch ratio is not directly comparable across different $T$.}. We compute the \nbr for different settings by using the \dpaccounting library~\cite{dp_accounting}. Although the function that computes the \nbr is generally well-understood in the sense that we know how to compute it tightly given the privacy and training parameters, its precise behavior as a function of the privacy budget, compute budget, and data budget is not common knowledge. 
Indeed, due to lack of clear and simple guidance on how to configure \dpsgd, it is not uncommon to use or compare against sub-optimal configurations of \dpsgd. 

In \cref{fig:vector_fields} we plot three vector fields. Along each axis we vary the privacy budget, compute budget, and data budget. The direction and magnitude of the vectors indicate how much doubling each of these budgets reduces the \nbr. Each budget is varied on a logarithmic scale at different powers of $2$. The length of the $x$ and $y$ components of the vector is determined by ratio of \nbr minus one. For example, a vector of length $1$ along the privacy budget axis means doubling the privacy budget reduces the \nbr by a factor of two.

As there are three budgets that together determine the \nbr and they interact in nuanced ways, we show three plots in \cref{fig:vector_fields}, where we vary two of the budgets at a time while fixing the third. These plots together provide a fairly complete picture of the behavior of the \nbr. Our main observations are enumerated below:

\begin{itemize}[leftmargin=*, nosep, itemsep=0.5pt]
  \item In \cref{fig:accounting_privacy_vs_compute} we see that varying the privacy budget or compute budget alone (while fixing the other) leads to diminishing returns.
  Increasing the privacy and compute budgets in tandem leads to consistent and predictable reductions in the \nbr.
  \item In \cref{fig:accounting_data_vs_compute} we see a similar trend when varying data and compute budgets. At small compute budgets, increasing the data budget provides limited benefit, and vice-versa. Increasing them simultaneously leads to consistent and predictable improvements in the \nbr.
  \item In \cref{fig:accounting_data_vs_privacy} we see that while increasing data and privacy budgets can be helpful, for a fixed compute budget, increasing either provides diminishing and eventually negligible benefits.
\end{itemize}


These observations provide guidance on how to effectively configure \dpsgd and corroborate our scaling laws above.


\section{Related Work}\label{sec:related_work}

\textbf{Scaling Laws of Language Models.} Recent research has explored the scaling laws governing the performance of language models as they increase in size. \citet{kaplan2020scaling} found a power-law relationship between model size, dataset size, and compute budget, with performance on downstream tasks following predictable scaling curves. \citet{hoffmann2022training} extended this to open-ended language models, observing smooth scaling over 7 orders of magnitude. \citet{chowdhery2022palmscalinglanguagemodeling} trained PaLM, a 540 billion parameter model that continued the trends. These results suggest language models may continue improving as they scale, although \citet{Ganguli_2022} note scaling alone may not be sufficient for open-ended intelligence. In the context of training language models with DP, where gradient clipping and noise addition \citep{abadi2016deep} alter training dynamics, the scaling laws have remained largely unexplored until this work.

\textbf{Applying DP in Fine-tuning or Prompting.} Recent studies demonstrate that fine-tuning \citep{bu2023differentially,wang2023can,du2023dp,thaker2023leveraging,zhang2023differentially,tobaben2023efficacy,wu2024privately,zhang2024dpzero,chua2024mind} or prompting \citep{duan2023privacy,duan2024flocks,wu2024privacy,tang2024privacy,hong2024dp,amin2024private} LLMs can achieve strong performance while ensuring downstream data privacy. However, these privacy guarantees are limited to downstream data, leaving the pre-training process exposed. Given that LLMs are pre-trained on extensive Internet data, which is often sourced without explicit user consent \citep{gold2017robots}, this raises ethical and privacy concerns \citep{tramer2022considerations}. Safeguarding privacy during pre-training remains a significant challenge. This study seeks to provide new insights to advance privacy-preserving pre-training of language models.

\textbf{DP Training of Vision Models.} Training DP models from scratch for vision tasks is an active area of research \citep{yu2021large,de2022unlocking,bu2022scalable,kurakin2022toward,sander2024differentially}. The most related work is \citet{sander2023tan}, which investigates the scaling behavior of DP training on vision tasks by varying key hyperparameters. They demonstrate that, under a fixed privacy budget, carefully tuning batch size, training steps, and learning rate is critical for better accuracy. However, \citet{sander2023tan} do not account for a bounded compute budget, a crucial factor in scaling law studies for language models \citep{hoffmann2022training}. Additionally, it remains unclear how their findings translate to language modeling tasks. In this work, we extend scaling law analyses to language models, incorporating both standard optimization hyperparameters and a bounded compute budgets to align more closely with recent LLM scaling research.

\section{Conclusion and Future Directions}

This work establishes a principled methodology for understanding the \tradeoff of language models trained under DP, and it represents an important step towards training larger, more capable models efficiently on sensitive user data. This endeavor will require collecting increasingly larger datasets over larger groups of individuals, while simultaneously scaling up compute.  For example, to train a billion parameter model optimally with DP, one could collect data from one billion individuals, using a generous privacy budget of $\epsilon \approx 10$, and train on large compute clusters for $\approx 10^{23}$ FLOPs. This is in stark contrast to non-private laws, \eg~\citet{anil2023palm} suggests a much larger $\approx 20B$ parameter model could be trained with $\approx 2B$ examples.


This work raises several new questions worth exploring in future work, including how do the scaling laws change when (1) doing finetuning instead of pretraining, (2) using better underlying mechanisms, and (3) when allowed to vary the sequence length.  These questions (along with several others) are discussed in greater detail in \cref{sec:limitations}.
\clearpage
\newpage

\section*{Impact Statement}

This paper presents work whose goal is to advance the field of machine learning, specifically in the area of differentially private (DP) language models. It establishes DP scaling laws that shed light on the trade-offs between compute, privacy, and utility, and can lead to more efficient and effective methods for training LLMs on user data while satisfying DP, a gold standard for bounding the privacy loss. The scaling laws presented can help researchers and practitioners choose model sizes, batch sizes, and training iterations based on available compute, data, and privacy budgets. By developing methods to make DP training more feasible, the paper contributes to the responsible development and deployment of AI technologies. We point out that, when applying DP in practice, the privacy unit has to be chosen carefully; in particular, a user-level guarantee may be needed. Moreover, while a valuable tool, DP may not be sufficient when training on user data; additional mitigations may need to be simultaneously applied depending on the application.

{\small
\bibliography{refs}

\begin{thebibliography}{78}
\providecommand{\natexlab}[1]{#1}
\providecommand{\url}[1]{\texttt{#1}}
\expandafter\ifx\csname urlstyle\endcsname\relax
  \providecommand{\doi}[1]{doi: #1}\else
  \providecommand{\doi}{doi: \begingroup \urlstyle{rm}\Url}\fi

\bibitem[Abadi et~al.(2016)Abadi, Chu, Goodfellow, McMahan, Mironov, Talwar,
  and Zhang]{abadi2016deep}
Abadi, M., Chu, A., Goodfellow, I., McMahan, H.~B., Mironov, I., Talwar, K.,
  and Zhang, L.
\newblock Deep learning with differential privacy.
\newblock In \emph{CCS}, pp.\  308--318, 2016.

\bibitem[Achiam et~al.(2023)Achiam, Adler, Agarwal, Ahmad, Akkaya, Aleman,
  Almeida, Altenschmidt, Altman, Anadkat, et~al.]{achiam2023gpt}
Achiam, J., Adler, S., Agarwal, S., Ahmad, L., Akkaya, I., Aleman, F.~L.,
  Almeida, D., Altenschmidt, J., Altman, S., Anadkat, S., et~al.
\newblock {GPT}-4 technical report.
\newblock \emph{arXiv:2303.08774}, 2023.

\bibitem[Afonja et~al.(2024)Afonja, Sim, Lin, Inan, and
  Yekhanin]{microsoftdpsynthetic}
Afonja, G., Sim, R., Lin, Z., Inan, A., and Yekhanin, S.
\newblock The crossroads of innovation and privacy: Private synthetic data for
  generative {AI}.
\newblock Blog post, 2024.
\newblock URL
  \url{https://www.microsoft.com/en-us/research/blog/the-crossroads-of-innovation-and-privacy-private-synthetic-data-for-generative-ai}.

\bibitem[Amin et~al.(2024)Amin, Bie, Kong, Kurakin, Ponomareva, Syed, Terzis,
  and Vassilvitskii]{amin2024private}
Amin, K., Bie, A., Kong, W., Kurakin, A., Ponomareva, N., Syed, U., Terzis, A.,
  and Vassilvitskii, S.
\newblock Private prediction for large-scale synthetic text generation.
\newblock \emph{arXi:2407.12108}, 2024.

\bibitem[Anil et~al.(2022)Anil, Ghazi, Gupta, Kumar, and
  Manurangsi]{anil2021large}
Anil, R., Ghazi, B., Gupta, V., Kumar, R., and Manurangsi, P.
\newblock Large-scale differentially private {BERT}.
\newblock In \emph{EMNLP (Findings)}, pp.\  6481--6491, 2022.

\bibitem[Anil et~al.(2023)Anil, Dai, Firat, Johnson, Lepikhin, Passos, Shakeri,
  Taropa, Bailey, Chen, et~al.]{anil2023palm}
Anil, R., Dai, A.~M., Firat, O., Johnson, M., Lepikhin, D., Passos, A.,
  Shakeri, S., Taropa, E., Bailey, P., Chen, Z., et~al.
\newblock Palm 2 technical report.
\newblock \emph{arXiv:2305.10403}, 2023.

\bibitem[Balle et~al.(2018)Balle, Barthe, and Gaboardi]{balle2018privacy}
Balle, B., Barthe, G., and Gaboardi, M.
\newblock Privacy amplification by subsampling: Tight analyses via couplings
  and divergences, 2018.

\bibitem[Balle et~al.(2022)Balle, Cherubin, and Hayes]{balle2022reconstructing}
Balle, B., Cherubin, G., and Hayes, J.
\newblock Reconstructing training data with informed adversaries.
\newblock In \emph{S \& P}, pp.\  1138--1156, 2022.

\bibitem[Bassily et~al.(2014)Bassily, Smith, and Thakurta]{bassily2014private}
Bassily, R., Smith, A., and Thakurta, A.
\newblock Private empirical risk minimization: Efficient algorithms and tight
  error bounds.
\newblock In \emph{FOCS}, pp.\  464--473, 2014.

\bibitem[Berrada et~al.(2023)Berrada, De, Shen, Hayes, Stanforth, Stutz, Kohli,
  Smith, and Balle]{DBLP:journals/corr/abs-2308-10888}
Berrada, L., De, S., Shen, J.~H., Hayes, J., Stanforth, R., Stutz, D., Kohli,
  P., Smith, S.~L., and Balle, B.
\newblock Unlocking accuracy and fairness in differentially private image
  classification.
\newblock \emph{arXiv:2308.10888}, 2023.

\bibitem[Biderman et~al.(2023)Biderman, Prashanth, Sutawika, Schoelkopf,
  Anthony, Purohit, and Raff]{biderman2024emergent}
Biderman, S., Prashanth, U.~S., Sutawika, L., Schoelkopf, H., Anthony, Q.,
  Purohit, S., and Raff, E.
\newblock Emergent and predictable memorization in large language models.
\newblock In \emph{NeurIPS}, 2023.

\bibitem[Bu et~al.(2022)Bu, Mao, and Xu]{bu2022scalable}
Bu, Z., Mao, J., and Xu, S.
\newblock Scalable and efficient training of large convolutional neural
  networks with differential privacy.
\newblock In \emph{NeurIPS}, 2022.

\bibitem[Bu et~al.(2023)Bu, Wang, Zha, and Karypis]{bu2023differentially}
Bu, Z., Wang, Y., Zha, S., and Karypis, G.
\newblock Differentially private optimization on large model at small cost.
\newblock In \emph{ICML}, pp.\  3192--3218, 2023.

\bibitem[Carlini et~al.(2021)Carlini, Tramer, Wallace, Jagielski, Herbert-Voss,
  Lee, Roberts, Brown, Song, Erlingsson, et~al.]{carlini2021extracting}
Carlini, N., Tramer, F., Wallace, E., Jagielski, M., Herbert-Voss, A., Lee, K.,
  Roberts, A., Brown, T., Song, D., Erlingsson, U., et~al.
\newblock Extracting training data from large language models.
\newblock In \emph{USENIX Security}, 2021.

\bibitem[Carlini et~al.(2023)Carlini, Ippolito, Jagielski, Lee, Tram{\`{e}}r,
  and Zhang]{carlini22quantifying}
Carlini, N., Ippolito, D., Jagielski, M., Lee, K., Tram{\`{e}}r, F., and Zhang,
  C.
\newblock Quantifying memorization across neural language models.
\newblock In \emph{ICLR}, 2023.

\bibitem[Carlini et~al.(2024)Carlini, Jagielski, Choquette-Choo, Paleka,
  Pearce, Anderson, Terzis, Thomas, and Tram{\`e}r]{carlini2024poisoning}
Carlini, N., Jagielski, M., Choquette-Choo, C.~A., Paleka, D., Pearce, W.,
  Anderson, H., Terzis, A., Thomas, K., and Tram{\`e}r, F.
\newblock Poisoning web-scale training datasets is practical.
\newblock In \emph{S \& P}, pp.\  407--425, 2024.

\bibitem[Charles et~al.(2024)Charles, Ganesh, McKenna, McMahan, Mitchell,
  Pillutla, and Rush]{charles2024fine}
Charles, Z., Ganesh, A., McKenna, R., McMahan, H.~B., Mitchell, N., Pillutla,
  K., and Rush, K.
\newblock Fine-tuning large language models with user-level differential
  privacy.
\newblock \emph{arXiv:2407.07737}, 2024.

\bibitem[Chen et~al.(2023)Chen, Liang, Huang, Real, Wang, Liu, Pham, Dong,
  Luong, Hsieh, Lu, and Le]{chen2023symbolicdiscoveryoptimizationalgorithms}
Chen, X., Liang, C., Huang, D., Real, E., Wang, K., Liu, Y., Pham, H., Dong,
  X., Luong, T., Hsieh, C.-J., Lu, Y., and Le, Q.~V.
\newblock Symbolic discovery of optimization algorithms, 2023.
\newblock URL \url{https://arxiv.org/abs/2302.06675}.

\bibitem[Chowdhery et~al.(2022)Chowdhery, Narang, Devlin, Bosma, Mishra,
  Roberts, Barham, Chung, Sutton, Gehrmann, Schuh, Shi, Tsvyashchenko, Maynez,
  Rao, Barnes, Tay, Shazeer, Prabhakaran, Reif, Du, Hutchinson, Pope, Bradbury,
  Austin, Isard, Gur-Ari, Yin, Duke, Levskaya, Ghemawat, Dev, Michalewski,
  Garcia, Misra, Robinson, Fedus, Zhou, Ippolito, Luan, Lim, Zoph, Spiridonov,
  Sepassi, Dohan, Agrawal, Omernick, Dai, Pillai, Pellat, Lewkowycz, Moreira,
  Child, Polozov, Lee, Zhou, Wang, Saeta, Diaz, Firat, Catasta, Wei,
  Meier-Hellstern, Eck, Dean, Petrov, and
  Fiedel]{chowdhery2022palmscalinglanguagemodeling}
Chowdhery, A., Narang, S., Devlin, J., Bosma, M., Mishra, G., Roberts, A.,
  Barham, P., Chung, H.~W., Sutton, C., Gehrmann, S., Schuh, P., Shi, K.,
  Tsvyashchenko, S., Maynez, J., Rao, A., Barnes, P., Tay, Y., Shazeer, N.,
  Prabhakaran, V., Reif, E., Du, N., Hutchinson, B., Pope, R., Bradbury, J.,
  Austin, J., Isard, M., Gur-Ari, G., Yin, P., Duke, T., Levskaya, A.,
  Ghemawat, S., Dev, S., Michalewski, H., Garcia, X., Misra, V., Robinson, K.,
  Fedus, L., Zhou, D., Ippolito, D., Luan, D., Lim, H., Zoph, B., Spiridonov,
  A., Sepassi, R., Dohan, D., Agrawal, S., Omernick, M., Dai, A.~M., Pillai,
  T.~S., Pellat, M., Lewkowycz, A., Moreira, E., Child, R., Polozov, O., Lee,
  K., Zhou, Z., Wang, X., Saeta, B., Diaz, M., Firat, O., Catasta, M., Wei, J.,
  Meier-Hellstern, K., Eck, D., Dean, J., Petrov, S., and Fiedel, N.
\newblock {PaLM}: Scaling language modeling with pathways, 2022.

\bibitem[Chua et~al.(2024{\natexlab{a}})Chua, Ghazi, Huang, Kamath, Kumar, Liu,
  Manurangsi, Sinha, and Zhang]{chua2024mind}
Chua, L., Ghazi, B., Huang, Y., Kamath, P., Kumar, R., Liu, D., Manurangsi, P.,
  Sinha, A., and Zhang, C.
\newblock Mind the privacy unit! user-level differential privacy for language
  model fine-tuning.
\newblock In \emph{CoLM}, 2024{\natexlab{a}}.

\bibitem[Chua et~al.(2024{\natexlab{b}})Chua, Ghazi, Kamath, Kumar, Manurangsi,
  Sinha, and Zhang]{chua2024scalable}
Chua, L., Ghazi, B., Kamath, P., Kumar, R., Manurangsi, P., Sinha, A., and
  Zhang, C.
\newblock Scalable {DP-SGD}: Shuffling vs. {P}oisson subsampling.
\newblock In \emph{NeurIPS}, 2024{\natexlab{b}}.

\bibitem[De et~al.(2022)De, Berrada, Hayes, Smith, and Balle]{de2022unlocking}
De, S., Berrada, L., Hayes, J., Smith, S.~L., and Balle, B.
\newblock Unlocking high-accuracy differentially private image classification
  through scale.
\newblock \emph{arXiv:2204.13650}, 2022.

\bibitem[Devlin et~al.(2019)Devlin, Chang, Lee, and Toutanova]{devlin2018bert}
Devlin, J., Chang, M.-W., Lee, K., and Toutanova, K.
\newblock {BERT}: Pre-training of deep bidirectional transformers for language
  understanding.
\newblock In \emph{NAACL-HLT}, pp.\  4171--4186, 2019.

\bibitem[Du et~al.(2023)Du, Yue, Chow, Wang, Huang, and Sun]{du2023dp}
Du, M., Yue, X., Chow, S.~S., Wang, T., Huang, C., and Sun, H.
\newblock {DP}-forward: Fine-tuning and inference on language models with
  differential privacy in forward pass.
\newblock In \emph{CCS}, pp.\  2665--2679, 2023.

\bibitem[Duan et~al.(2023{\natexlab{a}})Duan, Dziedzic, Papernot, and
  Boenisch]{duan2024flocks}
Duan, H., Dziedzic, A., Papernot, N., and Boenisch, F.
\newblock Flocks of stochastic parrots: Differentially private prompt learning
  for large language models.
\newblock In \emph{NeurIPS}, 2023{\natexlab{a}}.

\bibitem[Duan et~al.(2023{\natexlab{b}})Duan, Dziedzic, Yaghini, Papernot, and
  Boenisch]{duan2023privacy}
Duan, H., Dziedzic, A., Yaghini, M., Papernot, N., and Boenisch, F.
\newblock On the privacy risk of in-context learning.
\newblock In \emph{ACL}, 2023{\natexlab{b}}.

\bibitem[Dubey et~al.(2024)Dubey, Jauhri, Pandey, Kadian, Al-Dahle, Letman,
  Mathur, Schelten, Yang, Fan, et~al.]{dubey2024llama}
Dubey, A., Jauhri, A., Pandey, A., Kadian, A., Al-Dahle, A., Letman, A.,
  Mathur, A., Schelten, A., Yang, A., Fan, A., et~al.
\newblock The {Llama} 3 herd of models.
\newblock \emph{arXiv:2407.21783}, 2024.

\bibitem[Dwork et~al.(2006)Dwork, McSherry, Nissim, and
  Smith]{dwork2006calibrating}
Dwork, C., McSherry, F., Nissim, K., and Smith, A.
\newblock Calibrating noise to sensitivity in private data analysis.
\newblock In \emph{TCC}, pp.\  265--284, 2006.

\bibitem[Gadre et~al.(2024)Gadre, Smyrnis, Shankar, Gururangan, Wortsman, Shao,
  Mercat, Fang, Li, Keh, et~al.]{gadre2024language}
Gadre, S.~Y., Smyrnis, G., Shankar, V., Gururangan, S., Wortsman, M., Shao, R.,
  Mercat, J., Fang, A., Li, J., Keh, S., et~al.
\newblock Language models scale reliably with over-training and on downstream
  tasks.
\newblock \emph{arXiv:2403.08540}, 2024.

\bibitem[Ganguli et~al.(2022)Ganguli, Hernandez, Lovitt, Askell, Bai, Chen,
  Conerly, Dassarma, Drain, Elhage, El~Showk, Fort, Hatfield-Dodds, Henighan,
  Johnston, Jones, Joseph, Kernian, Kravec, Mann, Nanda, Ndousse, Olsson,
  Amodei, Brown, Kaplan, McCandlish, Olah, Amodei, and Clark]{Ganguli_2022}
Ganguli, D., Hernandez, D., Lovitt, L., Askell, A., Bai, Y., Chen, A., Conerly,
  T., Dassarma, N., Drain, D., Elhage, N., El~Showk, S., Fort, S.,
  Hatfield-Dodds, Z., Henighan, T., Johnston, S., Jones, A., Joseph, N.,
  Kernian, J., Kravec, S., Mann, B., Nanda, N., Ndousse, K., Olsson, C.,
  Amodei, D., Brown, T., Kaplan, J., McCandlish, S., Olah, C., Amodei, D., and
  Clark, J.
\newblock Predictability and surprise in large generative models.
\newblock In \emph{FAccT}, 2022.

\bibitem[{Gemini Team}(2023)]{team2023gemini}
{Gemini Team}.
\newblock Gemini: a family of highly capable multimodal models.
\newblock \emph{arXiv:2312.11805}, 2023.

\bibitem[{Gemma Team} et~al.(2024{\natexlab{a}}){Gemma Team}, Mesnard, Hardin,
  Dadashi, Bhupatiraju, Pathak, Sifre, Rivi{\`e}re, Kale, Love,
  et~al.]{team2024gemma}
{Gemma Team}, Mesnard, T., Hardin, C., Dadashi, R., Bhupatiraju, S., Pathak,
  S., Sifre, L., Rivi{\`e}re, M., Kale, M.~S., Love, J., et~al.
\newblock Gemma: Open models based on gemini research and technology.
\newblock \emph{arXiv:2403.08295}, 2024{\natexlab{a}}.

\bibitem[{Gemma Team} et~al.(2024{\natexlab{b}}){Gemma Team}, Riviere, Pathak,
  Sessa, Hardin, Bhupatiraju, Hussenot, Mesnard, Shahriari, Ram{\'e},
  et~al.]{team2024gemma2}
{Gemma Team}, Riviere, M., Pathak, S., Sessa, P.~G., Hardin, C., Bhupatiraju,
  S., Hussenot, L., Mesnard, T., Shahriari, B., Ram{\'e}, A., et~al.
\newblock Gemma 2: Improving open language models at a practical size.
\newblock \emph{arXiv:2408.00118}, 2024{\natexlab{b}}.

\bibitem[Ghalebikesabi et~al.(2023)Ghalebikesabi, Berrada, Gowal, Ktena,
  Stanforth, Hayes, De, Smith, Wiles, and
  Balle]{DBLP:journals/corr/abs-2302-13861}
Ghalebikesabi, S., Berrada, L., Gowal, S., Ktena, I., Stanforth, R., Hayes, J.,
  De, S., Smith, S.~L., Wiles, O., and Balle, B.
\newblock Differentially private diffusion models generate useful synthetic
  images.
\newblock \emph{arXiv:2302.13861}, 2023.

\bibitem[Gold \& Latonero(2017)Gold and Latonero]{gold2017robots}
Gold, Z. and Latonero, M.
\newblock Robots welcome: Ethical and legal considerations for web crawling and
  scraping.
\newblock \emph{Wash. JL Tech. \& Arts}, 2017.

\bibitem[{Google DP Team}(2022)]{dp_accounting}
{Google DP Team}.
\newblock Google's differential privacy libraries., 2022.
\newblock \url{https://github.com/google/differential-privacy}.

\bibitem[Hoffmann et~al.(2022)Hoffmann, Borgeaud, Mensch, Buchatskaya, Cai,
  Rutherford, Casas, Hendricks, Welbl, Clark, et~al.]{hoffmann2022training}
Hoffmann, J., Borgeaud, S., Mensch, A., Buchatskaya, E., Cai, T., Rutherford,
  E., Casas, D. d.~L., Hendricks, L.~A., Welbl, J., Clark, A., et~al.
\newblock Training compute-optimal large language models.
\newblock \emph{arXiv:2203.15556}, 2022.

\bibitem[Hong et~al.(2024)Hong, Wang, Zhang, Li, Li, and Wang]{hong2024dp}
Hong, J., Wang, J.~T., Zhang, C., Li, Z., Li, B., and Wang, Z.
\newblock {DP-OPT}: Make large language model your privacy-preserving prompt
  engineer.
\newblock In \emph{ICLR}, 2024.

\bibitem[Huber(1992)]{huber1992robust}
Huber, P.~J.
\newblock Robust estimation of a location parameter.
\newblock In \emph{Breakthroughs in statistics: Methodology and distribution},
  pp.\  492--518. Springer, 1992.

\bibitem[Ippolito et~al.(2022)Ippolito, Tram{\`e}r, Nasr, Zhang, Jagielski,
  Lee, Choquette-Choo, and Carlini]{ippolito2022preventing}
Ippolito, D., Tram{\`e}r, F., Nasr, M., Zhang, C., Jagielski, M., Lee, K.,
  Choquette-Choo, C.~A., and Carlini, N.
\newblock Preventing verbatim memorization in language models gives a false
  sense of privacy.
\newblock \emph{arXiv:2210.17546}, 2022.

\bibitem[Kaissis et~al.(2023)Kaissis, Hayes, Ziller, and
  Rueckert]{kaissis2023bounding}
Kaissis, G., Hayes, J., Ziller, A., and Rueckert, D.
\newblock Bounding data reconstruction attacks with the hypothesis testing
  interpretation of differential privacy.
\newblock \emph{arXiv:2307.03928}, 2023.

\bibitem[Kaissis et~al.(2024)Kaissis, Kolek, Balle, Hayes, and
  Rueckert]{kaissis2024beyond}
Kaissis, G., Kolek, S., Balle, B., Hayes, J., and Rueckert, D.
\newblock Beyond the calibration point: Mechanism comparison in differential
  privacy.
\newblock In \emph{ICML}, pp.\  22840--22860, 2024.

\bibitem[Kaplan et~al.(2020)Kaplan, McCandlish, Henighan, Brown, Chess, Child,
  Gray, Radford, Wu, and Amodei]{kaplan2020scaling}
Kaplan, J., McCandlish, S., Henighan, T., Brown, T.~B., Chess, B., Child, R.,
  Gray, S., Radford, A., Wu, J., and Amodei, D.
\newblock Scaling laws for neural language models.
\newblock \emph{arXiv:2001.08361}, 2020.

\bibitem[Kingma \& Ba(2015)Kingma and Ba]{kingma2015adam}
Kingma, D.~P. and Ba, J.
\newblock Adam: {A} method for stochastic optimization.
\newblock In \emph{ICLR}, 2015.

\bibitem[Kurakin \& Ponomareva(2024)Kurakin and Ponomareva]{googledpsynthetic}
Kurakin, A. and Ponomareva, N.
\newblock Protecting users with differentially private synthetic training data.
\newblock Blog post, 2024.
\newblock URL
  \url{https://research.google/blog/protecting-users-with-differentially-private-synthetic-training-data/}.

\bibitem[Kurakin et~al.(2022)Kurakin, Song, Chien, Geambasu, Terzis, and
  Thakurta]{kurakin2022toward}
Kurakin, A., Song, S., Chien, S., Geambasu, R., Terzis, A., and Thakurta, A.
\newblock Toward training at {ImageNet} scale with differential privacy.
\newblock \emph{arXiv:2201.12328}, 2022.

\bibitem[Li et~al.(2022)Li, Tram{\`{e}}r, Liang, and
  Hashimoto]{li2022largelanguagemodelsstrong}
Li, X., Tram{\`{e}}r, F., Liang, P., and Hashimoto, T.
\newblock Large language models can be strong differentially private learners.
\newblock In \emph{ICLR}, 2022.

\bibitem[Liu et~al.(2024)Liu, Novak, Lee, Wortsman, Xiao, Everett, Alemi,
  Kurzeja, Marcenac, Gur, Kornblith, Xu, Elsayed, Fischer, Pennington, Adlam,
  and Dickstein]{nanodo}
Liu, P.~J., Novak, R., Lee, J., Wortsman, M., Xiao, L., Everett, K., Alemi,
  A.~A., Kurzeja, M., Marcenac, P., Gur, I., Kornblith, S., Xu, K., Elsayed,
  G., Fischer, I., Pennington, J., Adlam, B., and Dickstein, J.-S.
\newblock {NanoDO}: A minimal transformer decoder-only language model
  implementation in {JAX}., 2024.
\newblock URL \url{http://github.com/google-deepmind/nanodo}.

\bibitem[Loshchilov \& Hutter(2019)Loshchilov and
  Hutter]{loshchilov2019decoupledweightdecayregularization}
Loshchilov, I. and Hutter, F.
\newblock Decoupled weight decay regularization, 2019.
\newblock URL \url{https://arxiv.org/abs/1711.05101}.

\bibitem[Lukas et~al.(2023)Lukas, Salem, Sim, Tople, Wutschitz, and
  Zanella-B{\'e}guelin]{lukas2023analyzing}
Lukas, N., Salem, A., Sim, R., Tople, S., Wutschitz, L., and
  Zanella-B{\'e}guelin, S.
\newblock Analyzing leakage of personally identifiable information in language
  models.
\newblock In \emph{S \& P}, 2023.

\bibitem[McCandlish et~al.(2018)McCandlish, Kaplan, Amodei, and
  Team]{mccandlish2018empirical}
McCandlish, S., Kaplan, J., Amodei, D., and Team, O.~D.
\newblock An empirical model of large-batch training.
\newblock \emph{arXiv:1812.06162}, 2018.

\bibitem[Nocedal(1980)]{nocedal1980updating}
Nocedal, J.
\newblock Updating quasi-{N}ewton matrices with limited storage.
\newblock \emph{Mathematics of Computation}, 35\penalty0 (151):\penalty0
  773--782, 1980.

\bibitem[Nocedal \& Wright(1999)Nocedal and Wright]{nocedal1999numerical}
Nocedal, J. and Wright, S.~J.
\newblock \emph{Numerical optimization}.
\newblock Springer, 1999.

\bibitem[Ponomareva et~al.(2023)Ponomareva, Hazimeh, Kurakin, Xu, Denison,
  McMahan, Vassilvitskii, Chien, and Thakurta]{ponomareva2023dp}
Ponomareva, N., Hazimeh, H., Kurakin, A., Xu, Z., Denison, C., McMahan, H.~B.,
  Vassilvitskii, S., Chien, S., and Thakurta, A.~G.
\newblock How to {DP}-fy {ML}: A practical guide to machine learning with
  differential privacy.
\newblock \emph{JAIR}, 2023.

\bibitem[Prashanth et~al.(2024)Prashanth, Deng, O'Brien, SV, Khan, Borkar,
  Choquette-Choo, Fuehne, Biderman, Ke, et~al.]{prashanth2024recite}
Prashanth, U.~S., Deng, A., O'Brien, K., SV, J., Khan, M.~A., Borkar, J.,
  Choquette-Choo, C.~A., Fuehne, J.~R., Biderman, S., Ke, T., et~al.
\newblock Recite, reconstruct, recollect: Memorization in {LMs} as a
  multifaceted phenomenon.
\newblock \emph{arXiv:2406.17746}, 2024.

\bibitem[Rush et~al.(2024)Rush, Charles, Garrett, Augenstein, and
  Mitchell]{rush2024drjax}
Rush, J.~K., Charles, Z., Garrett, Z., Augenstein, S., and Mitchell, N.~E.
\newblock {DrJAX}: Scalable and differentiable mapreduce primitives in {JAX}.
\newblock In \emph{WANT@ ICML}, 2024.

\bibitem[Sander et~al.(2023)Sander, Stock, and Sablayrolles]{sander2023tan}
Sander, T., Stock, P., and Sablayrolles, A.
\newblock {TAN} without a burn: Scaling laws of {DP-SGD}.
\newblock In \emph{ICML}, pp.\  29937--29949, 2023.

\bibitem[Sander et~al.(2024)Sander, Yu, Sanjabi, Durmus, Ma, Chaudhuri, and
  Guo]{sander2024differentially}
Sander, T., Yu, Y., Sanjabi, M., Durmus, A., Ma, Y., Chaudhuri, K., and Guo, C.
\newblock Differentially private representation learning via image captioning.
\newblock In \emph{ICML}, 2024.

\bibitem[Shallue et~al.(2019)Shallue, Lee, Antognini, Sohl-Dickstein, Frostig,
  and Dahl]{shallue2019measuring}
Shallue, C.~J., Lee, J., Antognini, J., Sohl-Dickstein, J., Frostig, R., and
  Dahl, G.~E.
\newblock Measuring the effects of data parallelism on neural network training.
\newblock \emph{JMLR}, 2019.

\bibitem[Stiennon et~al.(2020)Stiennon, Ouyang, Wu, Ziegler, Lowe, Voss,
  Radford, Amodei, and Christiano]{stiennon2020learning}
Stiennon, N., Ouyang, L., Wu, J., Ziegler, D.~M., Lowe, R., Voss, C., Radford,
  A., Amodei, D., and Christiano, P.~F.
\newblock Learning to summarize with human feedback.
\newblock In \emph{NeurIPS}, 2020.

\bibitem[Subramani et~al.(2021)Subramani, Vadivelu, and
  Kamath]{subramani2021enabling}
Subramani, P., Vadivelu, N., and Kamath, G.
\newblock Enabling fast differentially private {SGD} via just-in-time
  compilation and vectorization.
\newblock In \emph{NeurIPS}, pp.\  26409--26421, 2021.

\bibitem[Tang et~al.(2024)Tang, Shin, Inan, Manoel, Mireshghallah, Lin, Gopi,
  Kulkarni, and Sim]{tang2024privacy}
Tang, X., Shin, R., Inan, H.~A., Manoel, A., Mireshghallah, F., Lin, Z., Gopi,
  S., Kulkarni, J., and Sim, R.
\newblock Privacy-preserving in-context learning with differentially private
  few-shot generation.
\newblock \emph{ICLR}, 2024.

\bibitem[Thaker et~al.(2023)Thaker, Setlur, Wu, and
  Smith]{thaker2023leveraging}
Thaker, P., Setlur, A., Wu, Z.~S., and Smith, V.
\newblock Leveraging public representations for private transfer learning.
\newblock \emph{arXiv:2312.15551}, 2023.

\bibitem[Tobaben et~al.(2023)Tobaben, Shysheya, Bronskill, Paverd, Tople,
  Zanella-Beguelin, Turner, and Honkela]{tobaben2023efficacy}
Tobaben, M., Shysheya, A., Bronskill, J., Paverd, A., Tople, S.,
  Zanella-Beguelin, S., Turner, R.~E., and Honkela, A.
\newblock On the efficacy of differentially private few-shot image
  classification.
\newblock \emph{TMLR}, 2023.

\bibitem[Tram{\`e}r et~al.(2022)Tram{\`e}r, Kamath, and
  Carlini]{tramer2022considerations}
Tram{\`e}r, F., Kamath, G., and Carlini, N.
\newblock Considerations for differentially private learning with large-scale
  public pretraining.
\newblock \emph{arXiv:2212.06470}, 2022.

\bibitem[Wang et~al.(2024)Wang, Zhang, Cao, Li, McMahan, Oh, Xu, and
  Zaheer]{wang2023can}
Wang, B., Zhang, Y., Cao, Y., Li, B., McMahan, H., Oh, S., Xu, Z., and Zaheer,
  M.
\newblock Can public large language models help private cross-device federated
  learning?
\newblock In \emph{NAACL (Findings)}, pp.\  934--949, 2024.

\bibitem[Wu et~al.(2024{\natexlab{a}})Wu, Inan, Backurs, Chandrasekaran,
  Kulkarni, and Sim]{wu2024privately}
Wu, F., Inan, H.~A., Backurs, A., Chandrasekaran, V., Kulkarni, J., and Sim, R.
\newblock Privately aligning language models with reinforcement learning.
\newblock \emph{ICLR}, 2024{\natexlab{a}}.

\bibitem[Wu et~al.(2024{\natexlab{b}})Wu, Panda, Wang, and
  Mittal]{wu2024privacy}
Wu, T., Panda, A., Wang, J.~T., and Mittal, P.
\newblock Privacy-preserving in-context learning for large language models.
\newblock In \emph{ICLR}, 2024{\natexlab{b}}.

\bibitem[Xu et~al.(2021)Xu, Lee, Chen, Hechtman, Huang, Joshi, Krikun,
  Lepikhin, Ly, Maggioni, et~al.]{xu2021gspmd}
Xu, Y., Lee, H., Chen, D., Hechtman, B., Huang, Y., Joshi, R., Krikun, M.,
  Lepikhin, D., Ly, A., Maggioni, M., et~al.
\newblock {GSPMD}: general and scalable parallelization for {ML} computation
  graphs.
\newblock \emph{arXiv:2105.04663}, 2021.

\bibitem[Yeom et~al.(2018)Yeom, Giacomelli, Fredrikson, and
  Jha]{yeom2018privacy}
Yeom, S., Giacomelli, I., Fredrikson, M., and Jha, S.
\newblock Privacy risk in machine learning: Analyzing the connection to
  overfitting.
\newblock In \emph{CSF}, pp.\  268--282, 2018.

\bibitem[You et~al.(2020)You, Li, Reddi, Hseu, Kumar, Bhojanapalli, Song,
  Demmel, Keutzer, and Hsieh]{you2020largebatchoptimizationdeep}
You, Y., Li, J., Reddi, S., Hseu, J., Kumar, S., Bhojanapalli, S., Song, X.,
  Demmel, J., Keutzer, K., and Hsieh, C.-J.
\newblock Large batch optimization for deep learning: Training bert in 76
  minutes, 2020.
\newblock URL \url{https://arxiv.org/abs/1904.00962}.

\bibitem[Yu et~al.(2021)Yu, Zhang, Chen, Yin, and Liu]{yu2021large}
Yu, D., Zhang, H., Chen, W., Yin, J., and Liu, T.-Y.
\newblock Large scale private learning via low-rank reparametrization.
\newblock In \emph{ICML}, 2021.

\bibitem[Yu et~al.(2022)Yu, Naik, Backurs, Gopi, Inan, Kamath, Kulkarni, Lee,
  Manoel, Wutschitz, Yekhanin, and Zhang]{yu2021differentially}
Yu, D., Naik, S., Backurs, A., Gopi, S., Inan, H.~A., Kamath, G., Kulkarni, J.,
  Lee, Y.~T., Manoel, A., Wutschitz, L., Yekhanin, S., and Zhang, H.
\newblock Differentially private fine-tuning of language models.
\newblock In \emph{ICLR}, 2022.

\bibitem[Zhang et~al.(2024{\natexlab{a}})Zhang, Morwani, Vyas, Wu, Zou, Ghai,
  Foster, and Kakade]{zhang2024does}
Zhang, H., Morwani, D., Vyas, N., Wu, J., Zou, D., Ghai, U., Foster, D., and
  Kakade, S.
\newblock How does critical batch size scale in pre-training?
\newblock \emph{arXiv:2410.21676}, 2024{\natexlab{a}}.

\bibitem[Zhang et~al.(2024{\natexlab{b}})Zhang, Li, Thekumparampil, Oh, and
  He]{zhang2024dpzero}
Zhang, L., Li, B., Thekumparampil, K.~K., Oh, S., and He, N.
\newblock {DPZero}: Private fine-tuning of language models without
  backpropagation.
\newblock In \emph{ICML}, 2024{\natexlab{b}}.

\bibitem[Zhang et~al.(2023)Zhang, Bu, Wu, and Hong]{zhang2023differentially}
Zhang, X., Bu, Z., Wu, Z.~S., and Hong, M.
\newblock Differentially private {SGD} without clipping bias: An error-feedback
  approach.
\newblock \emph{arXiv:2311.14632}, 2023.

\bibitem[Zhu et~al.(2015)Zhu, Kiros, Zemel, Salakhutdinov, Urtasun, Torralba,
  and Fidler]{zhu2015aligning}
Zhu, Y., Kiros, R., Zemel, R.~S., Salakhutdinov, R., Urtasun, R., Torralba, A.,
  and Fidler, S.
\newblock Aligning books and movies: Towards story-like visual explanations by
  watching movies and reading books.
\newblock In \emph{ICCV}, pp.\  19--27, 2015.

\bibitem[Ziller et~al.(2024)Ziller, Mueller, Stieger, Feiner, Brandt, Braren,
  Rueckert, and Kaissis]{ziller2024reconciling}
Ziller, A., Mueller, T.~T., Stieger, S., Feiner, L.~F., Brandt, J., Braren, R.,
  Rueckert, D., and Kaissis, G.
\newblock Reconciling privacy and accuracy in ai for medical imaging.
\newblock \emph{Nature Machine Intelligence}, 6\penalty0 (7):\penalty0
  764--774, 2024.

\end{thebibliography}
\bibliographystyle{icml2025}
}

\newpage
\appendix
\onecolumn
\appendix

\section{Limitations and Open Questions} \label{sec:limitations}

While our methodology revealed a number of interesting findings about the behavior of scaling laws under DP, there are some limitations of our approach and questions that remain unanswered that we enumerate below.

\paragraph{Fixed Physical Batch Size.} Our methodology relies crucially on the assumption that the Gaussian noise introduced to preserve privacy far outweighs the randomness introduced from minibatch sampling, and thus it would be sufficient vary the \nbr while keeping the physical batch size fixed to a large constant value of $1024$. \cref{sec:batch_size_ablation} reveals that this assumption may not be fully true, and that the physical batch size has a more nuanced effect that we cannot fully explain.

\paragraph{Robustness to other training setups.} Our methodology focuses on a single class of \bert models, with a fixed dataset and DP mechanism, which allowed us to do deeper experimentation on other relevant variables. Our general methodology holds for different models, datasets, and mechanisms, but the exact quantitative findings may differ under different training setups. As the field continues to make advancements on training transformers with DP, it would be interesting and informative to rerun our experiments with better base mechanisms.

\paragraph{Pretraining vs. Finetuning.} As an important first step, we focused on the pretraining regime in this work, where we start with a completely random model which we train from scratch. Finetuning a pretrained model with DP is often a preferable approach in practice to get the best privacy/utility trade-offs \cite{yu2021differentially,li2022largelanguagemodelsstrong}. There are a number of challenges to overcome to quantify the scaling laws in this regime, but it remains an interesting question for future work.

\paragraph{Sequence Length.} Our experiments focus on a fixed sequence length of $512$ tokens, which was the default value in the experiment we branched.  However, the sequence length is yet another important knob that can be tuned alongside the batch size, model size, and number of iterations in language modeling tasks. There are likely interesting trade-offs to explore here: with smaller sequence lengths, less context is available to predict the next / missing tokens, but the saved computation can be used to increase the batch size, model size, or number of iterations. Whether the trade-off is worth it likely depends on the exact setting as well as the distributional properties of the training data.

\paragraph{Over-Training and Inference-Time Compute.}

While this work focuses on the FLOPs required to pre-train a model to a given loss threshold, in practice language models are often over-trained in order to account for inference-time costs~\citep{gadre2024language}. If a model is going to be deployed, it may make sense to over-train a smaller model (which is cheaper to serve) than to train a larger model for a compute-optimal FLOPs budget. While we do not study over-training in our work, we note that such a study is particularly fruitful in the case of DP training; the privacy costs already often favor smaller models (when compared to non-private scaling laws). Investigating this confluence would likely yield valuable insights into DP scaling laws.

\paragraph{Larger Model Sizes.} The accuracy of any given scaling law is predicated to some degree on the range of model sizes trained on. For example, \citet{hoffmann2022training} train model of up to 16 billion parameters. Due to the necessity of using very large batch sizes (for privacy reasons) training models of such scale requires a significant amount of compute. We leave the task of training on model of larger scale to future work, along with analysis of how much this affects the derived scaling law.

\paragraph{Efficient implementations of per-example gradient clipping}

When considering to use a significant compute budget to train a large language model with DP, it is important that that model training code is carefully optimized to minimize the overheads of DP training. Using efficient vectorized per-example clipping implementations in JAX have been shown to work perform well with a reasonable overhead compared to non-private training \cite{subramani2021enabling}, although this focused on single-machine training scenarios, and more careful study is needed in this area when doing multi-machine training, especially when moving beyond pure data-parallelism which we focused on in this paper.

\paragraph{The choice of optimizer}

Our analysis relies on current optimization techniques which may not be optimal for privacy-preserving training. Several potential optimizer improvements could affect our findings. A uniformly better optimizer would likely preserve the observed scaling relationships while the actual optimal operating points might shift. In previous scaling law studies we do see the better optimizer can somehow smooth out the discontinuities in scaling behavior~\citep{chen2023symbolicdiscoveryoptimizationalgorithms,loshchilov2019decoupledweightdecayregularization} or even enable new scaling regimes sometimes (e.g., LAMB~\citep{you2020largebatchoptimizationdeep} for large batch size pre-training shows a very different scaling behavior). The optimizers specifically designed for privacy-preserving training might recommend a new set of parameters to enable better absolute performance. 

\section{Additional Details}

\subsection{Notes on \cref{alg:user_dp_sgd} (Generalized \dpsgd)} \label{sec:dpsgd_discussion}

\begin{algorithm}[t]
\caption{Generalized \dpsgd.}
\KwIn{Dataset $\mathcal{D}$, \nbr $\bar{\sigma}$, (expected) batch size $\batchsize$, iterations $\iters$}
\KwOut{Model parameters $\theta$.}
Initialize model parameters $\theta_0 \in \mathbb{R}^\modelsize$ \\
\For{$t = 1$ \KwTo $\iters$}{
    Select a (possibly random) size ${\approx}B$ minibatch $\mathcal{B}_t {\subset} \mathcal{D}$  \\
    $\bar{g}= \frac{1}{B} \sum_{\mathbf{x} \in \mathcal{B}_t} \text{clip}( \nabla \ell(\theta_{t-1}; \mathbf{x}))$ \\
    $\tilde{g} = g + \bar{\sigma} \mathcal{N}(0, 1)^\modelsize$ \\
    $\theta_t = \optupdate(\theta_{t-1}, \tilde{g})$
}
\textbf{return} $\theta_{\iters}$
\end{algorithm}

\textbf{Minibatch Selection} We were vague in our description of the minibatch selection step. In most descriptions of \dpsgd, the minibatch is formed by Poisson subsampling with a fixed probability. Sampling with or without replacement, as well as deterministic batching are also possible \cite{balle2018privacy}. In our paper, we calibrated noise under both the Poisson sampling assumption and the deterministic batching strategy, picking the lower noise multiplier. When doing Poisson sampling, we use the sampling probability $\nicefrac{B}{N}$ and noise multiplier $B \mul \bar{\sigma}$.

\textbf{Known Quantities} If doing Poisson sampling, we typically are operating under the add/remove adjacency definition. Under this definition, $\numsamples$ is considered a sensitive quantity which we do not have access to directly, hence we cannot technically define the sampling probability as $\nicefrac{\batchsize}{\numsamples}$ without violating DP. We also rely on $\numsamples$ later on, discussing its importance as it is interpreted as the data budget. If necessary, one can approximate $N$ quite accurately with DP since it is a simple count.

Alternatively, one can simply use the ``zero-out'' adjacency notion \cite{chua2024scalable}, where $\numsamples$ is known but Poisson sampling still enjoys the same privacy analysis.

\textbf{Clipping Function} We omit a clipping norm parameter in the definition of ``clip''.  This can be any function that maps an arbitrary real-valued vector to one with $\ell_2$ norm at most one.  One standard choice is to clip the norm to $C$, and then divide by $C$ \cite{de2022unlocking}.

\subsection{Unit of Privacy and Multiple Participations} \label{sec:user_level}

In traditional scaling laws work, it is common to assume access to an endless stream of data that does not require privacy protections. Therefore, every training example is only seen once, which simplifies the analysis of the scaling laws. In our case, we trained our models for $128K$ iterations with a physical batch size of $1024$, which is slightly less than a single pass over our entire dataset, satisfying the typical assumption. However, in our data analysis, we estimate what would happen with significantly larger batch sizes than we ran with, and in some cases this would involve multiple passes over the actual private dataset, something we did not account for directly in our analysis. Therefore, the actually setting that is best represented by our experimental methodology is not actually example-level DP, but rather user-level DP. There, we may assume that we have a finite number of users $N$ (which we should now interpret as the data budget), but we have an endless stream of data for each user. This circumvents the main concern, while allowing for users to participate multiple times during training which is typically very useful under DP. Alternatively, one can still consider the example-level DP setting, where each base example has multiple augmentations (e.g., rewritten text sequences that are semantically similar) that we can train on. All of our findings should hold, and be more reliable in this setting based on our methodology. 

\subsection{FLOPs estimation under DP} \label{sec:flops}
As defined in \cref{sec:prelim-scaling-laws}, we approximate the compute cost $\computebudget$ as $6 \mul \modelsize \mul \batchsize \mul \slen \mul \iters$ based on the non-private scaling laws~\citep{kaplan2020scaling,hoffmann2022training} except that $\batchsize$ represents the number of examples (not tokens) in a batch, as this determines the privacy budget. 
This cost model is useful because we can directly compare to the non-private scaling laws. Further, this cost model is also accurate because the extra overhead of \dpsgd in \cref{alg:user_dp_sgd} compared to Adam can be directly amortized: compiler-based systems like GSPMD~\citep{xu2021gspmd} and parallel machine learning libraries~\citep{rush2024drjax} let us parallelize the per-example gradient computations without a linear (in $\batchsize$) increase in memory usage. 
The total clipping costs are only a small linear cost (comprising of only element-wise operations and no matrix multiplications) in $\modelsize$, $\iters$, and $\batchsize$ (and are independent of sequence length $\slen$); the total noising costs are independent of $\batchsize$ and is linear in only $\modelsize$ and $\iters$. Thus, the overall compute in \dpsgd is dominated by the non-private approximation above.

\section{Additional Experiments}

\subsection{Saturating Compute Budget} \label{sec:saturating_compute}

Building on our findings above, it is natural to ask where the saturation point occurs for different privacy budget and data budgets. This can be helpful to determine how much compute is needed to get the most utility under a fixed data and privacy budget, as well as how to spend that compute optimally. These results are shown in \cref{table:inflection_points}.  

\begin{itemize}[leftmargin=*, nosep, itemsep=0.5pt]
  \item With a higher data and privacy budget, we benefit substantially from larger compute budgets.
  \item With DP, the compute-optimal training configurations requires training significantly smaller models over significantly more tokens than without DP. For these training configurations, the ratio of training tokens to model parameters varies in different settings, but in all settings it is significantly larger than it would be without DP, where prior work found $20\times$ to be a good rule of thumb \cite{hoffmann2022training}.
\end{itemize}

\begin{table}[htbp]
\caption{Saturating compute budgets, as well as optimal training configurations for those compute budgets across a representative set of data and privacy budgets.}
\label{table:inflection_points}
\centering
\setlength{\tabcolsep}{5pt}
\resizebox{0.618\linewidth}{!}{
\begin{tabular}{c|ccccccc}
\toprule
\textbf{Data} & \textbf{Privacy} & \textbf{Compute} & \textbf{Cross} & \textbf{Model} & \textbf{Iterations} & \textbf{Batch} & \textbf{Token / Model} \\
\textbf{Budget} & \textbf{Budget} & \textbf{Budget} & \textbf{Entropy} & \textbf{Size} & & \textbf{Size} & \textbf{Ratio} \\
\midrule
$1.0 \times 10^5$ & 1 & $1.3 \times 10^{16}$ & 7.28 & $4.6 \times 10^6$ & $1.8 \times 10^3$ & $5.1 \times 10^2$ & $1.0 \times 10^2$ \\
 & 4 & $1.1 \times 10^{17}$ & 6.65 & $4.6 \times 10^6$ & $1.8 \times 10^3$ & $4.1 \times 10^3$ & $8.5 \times 10^2$ \\
 & 16 & $2.0 \times 10^{18}$ & 5.60 & $1.7 \times 10^7$ & $2.7 \times 10^3$ & $1.4 \times 10^4$ & $1.1 \times 10^3$ \\
 & 64 & $7.5 \times 10^{18}$ & 4.63 & $2.0 \times 10^7$ & $6.3 \times 10^3$ & $1.9 \times 10^4$ & $3.2 \times 10^3$ \\
\midrule
$1.0 \times 10^6$ & 1 & $2.8 \times 10^{17}$ & 5.89 & $4.6 \times 10^6$ & $2.5 \times 10^3$ & $8.2 \times 10^3$ & $2.3 \times 10^3$ \\
 & 4 & $8.8 \times 10^{18}$ & 4.62 & $1.9 \times 10^7$ & $6.5 \times 10^3$ & $2.3 \times 10^4$ & $4.1 \times 10^3$ \\
 & 16 & $3.3 \times 10^{19}$ & 3.61 & $1.7 \times 10^7$ & $1.4 \times 10^4$ & $4.6 \times 10^4$ & $1.9 \times 10^4$ \\
 & 64 & $3.2 \times 10^{20}$ & 2.82 & $4.9 \times 10^7$ & $1.2 \times 10^4$ & $1.9 \times 10^5$ & $2.2 \times 10^4$ \\
\midrule
$1.0 \times 10^7$ & 1 & $3.8 \times 10^{19}$ & 3.73 & $1.7 \times 10^7$ & $9.6 \times 10^3$ & $7.8 \times 10^4$ & $2.3 \times 10^4$ \\
 & 4 & $3.8 \times 10^{20}$ & 2.81 & $4.9 \times 10^7$ & $1.1 \times 10^4$ & $2.2 \times 10^5$ & $2.6 \times 10^4$ \\
 & 16 & $2.0 \times 10^{21}$ & 2.15 & $7.0 \times 10^7$ & $1.2 \times 10^4$ & $7.4 \times 10^5$ & $6.7 \times 10^4$ \\
 & 64 & $4.4 \times 10^{22}$ & 1.66 & $3.3 \times 10^8$ & $4.9 \times 10^4$ & $8.8 \times 10^5$ & $6.7 \times 10^4$ \\
\midrule
$1.0 \times 10^8$ & 1 & $5.2 \times 10^{21}$ & 2.26 & $1.3 \times 10^8$ & $5.8 \times 10^4$ & $2.2 \times 10^5$ & $4.9 \times 10^4$ \\
 & 4 & $4.4 \times 10^{22}$ & 1.66 & $3.3 \times 10^8$ & $4.9 \times 10^4$ & $8.8 \times 10^5$ & $6.7 \times 10^4$ \\
 & 16 & $1.0 \times 10^{23}$ & 1.32 & $3.3 \times 10^8$ & $9.3 \times 10^4$ & $1.0 \times 10^6$ & $1.5 \times 10^5$ \\
 & 64 & $1.0 \times 10^{23}$ & 1.23 & $3.3 \times 10^8$ & $1.1 \times 10^5$ & $8.8 \times 10^5$ & $1.5 \times 10^5$ \\
\midrule
$1.0 \times 10^9$ & 1 & $8.5 \times 10^{22}$ & 1.36 & $3.3 \times 10^8$ & $9.4 \times 10^4$ & $8.8 \times 10^5$ & $1.3 \times 10^5$ \\
 & 4 & $1.0 \times 10^{23}$ & 1.23 & $3.3 \times 10^8$ & $1.1 \times 10^5$ & $8.8 \times 10^5$ & $1.5 \times 10^5$ \\
 & 16 & $1.0 \times 10^{23}$ & 1.22 & $3.3 \times 10^8$ & $1.1 \times 10^5$ & $8.8 \times 10^5$ & $1.5 \times 10^5$ \\
 & 64 & $1.2 \times 10^{23}$ & 1.20 & $3.3 \times 10^8$ & $1.1 \times 10^5$ & $1.1 \times 10^6$ & $1.8 \times 10^5$ \\
\bottomrule
\end{tabular}
}
\end{table}

\subsection{Full Experiment Grid}

\begin{figure*}[t!]
\includegraphics[width=\textwidth]{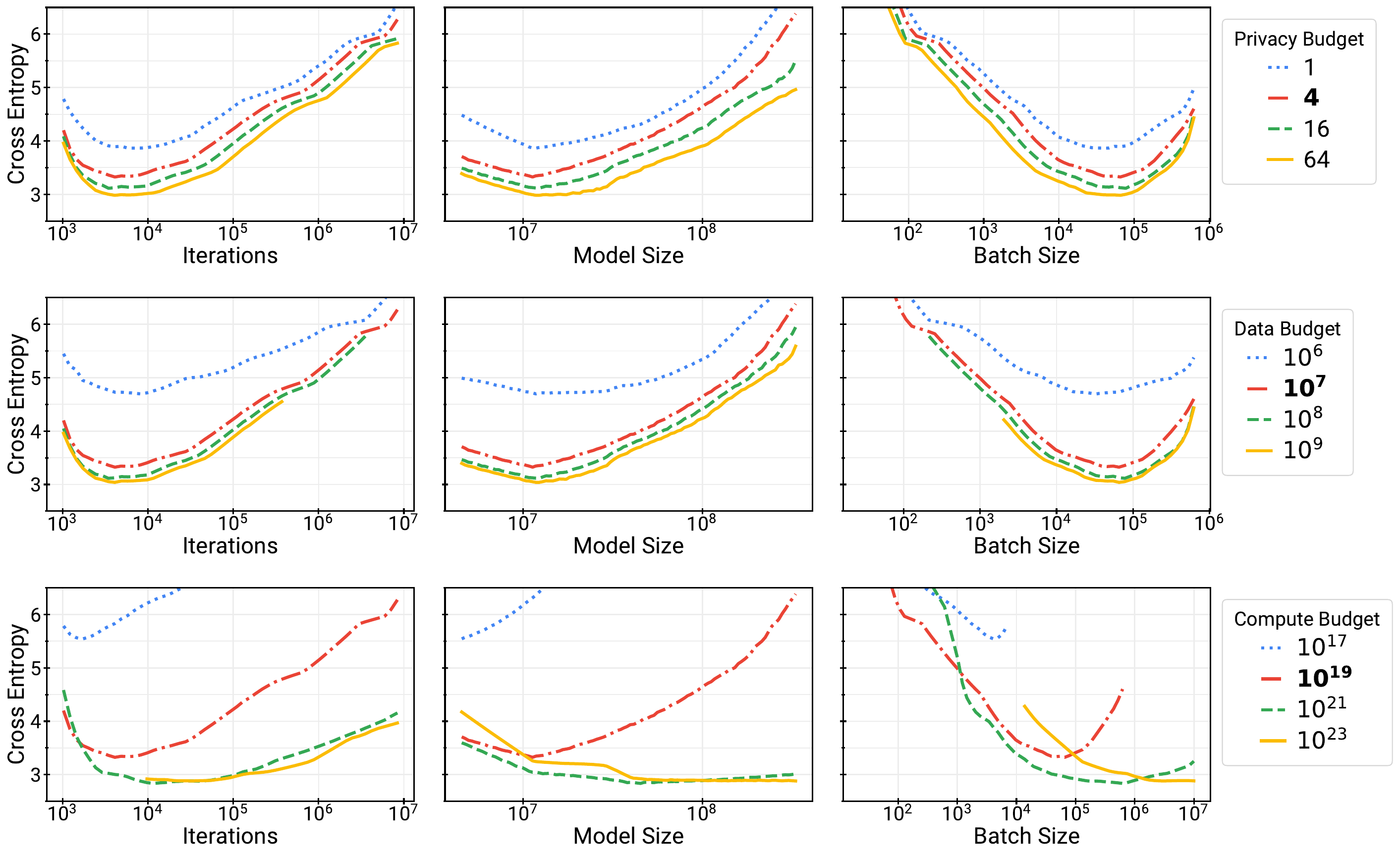}
\caption{Cross entropy of best models trained in each setting. From top to bottom , we vary the Privacy Budget, Data Budget, and Compute Budget, keeping the other two budgets fixed to default values (bolded). From left to right, we vary the number of Iterations, the Model Size, and the Batch Size, and treat the other two as nuisance parameters which we minimize over.} \label{fig:full_grid}
\end{figure*}

In \cref{fig:full_grid}, we plot the cross entropy loss for different privacy budgets, data budgets, and compute budgets under varying numbers of iterations, model sizes, and batch sizes. Much can be learned from these plots, including:

\begin{itemize}
  \item The optimal number of iterations typically falls around $\iters \approx 10K$, and the optimal batch size often falls in the range $\batchsize \approx 10 - 100K$, although neither of these are universally true and as expected it depends on the values of the privacy, data, and compute budgets. Batch size seems to be the most important parameter, as indicated by the steep slope of those lines. 
\end{itemize}

\subsection{Physical Batch Size Ablation} \label{sec:batch_size_ablation}

\begin{figure*}[h!]
\begin{subfigure}{0.31\linewidth}
\includegraphics[width=\textwidth]{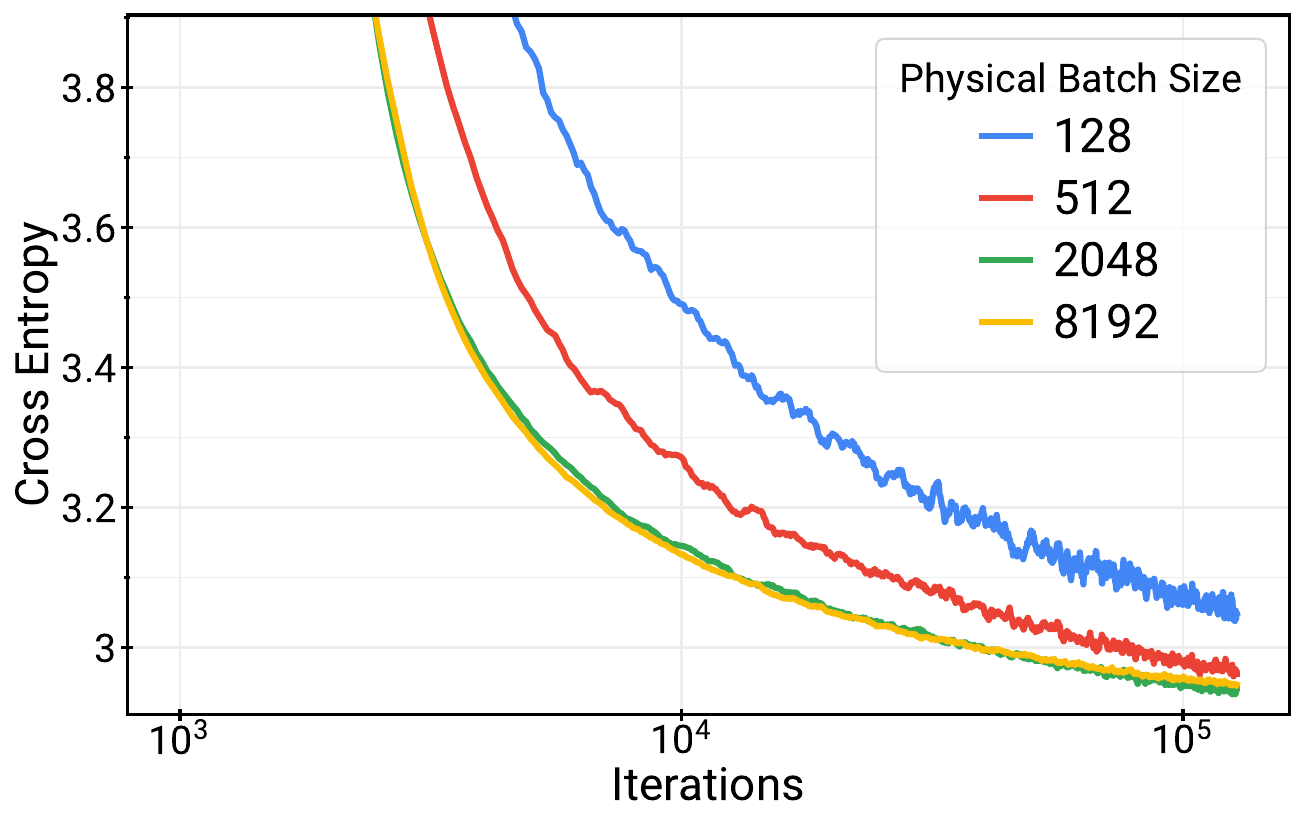}
\caption{\nbr $= 0.5^{20}$} \label{fig:physical_batch_size_small}
\end{subfigure}
\begin{subfigure}{0.3\linewidth}
\includegraphics[width=\textwidth]{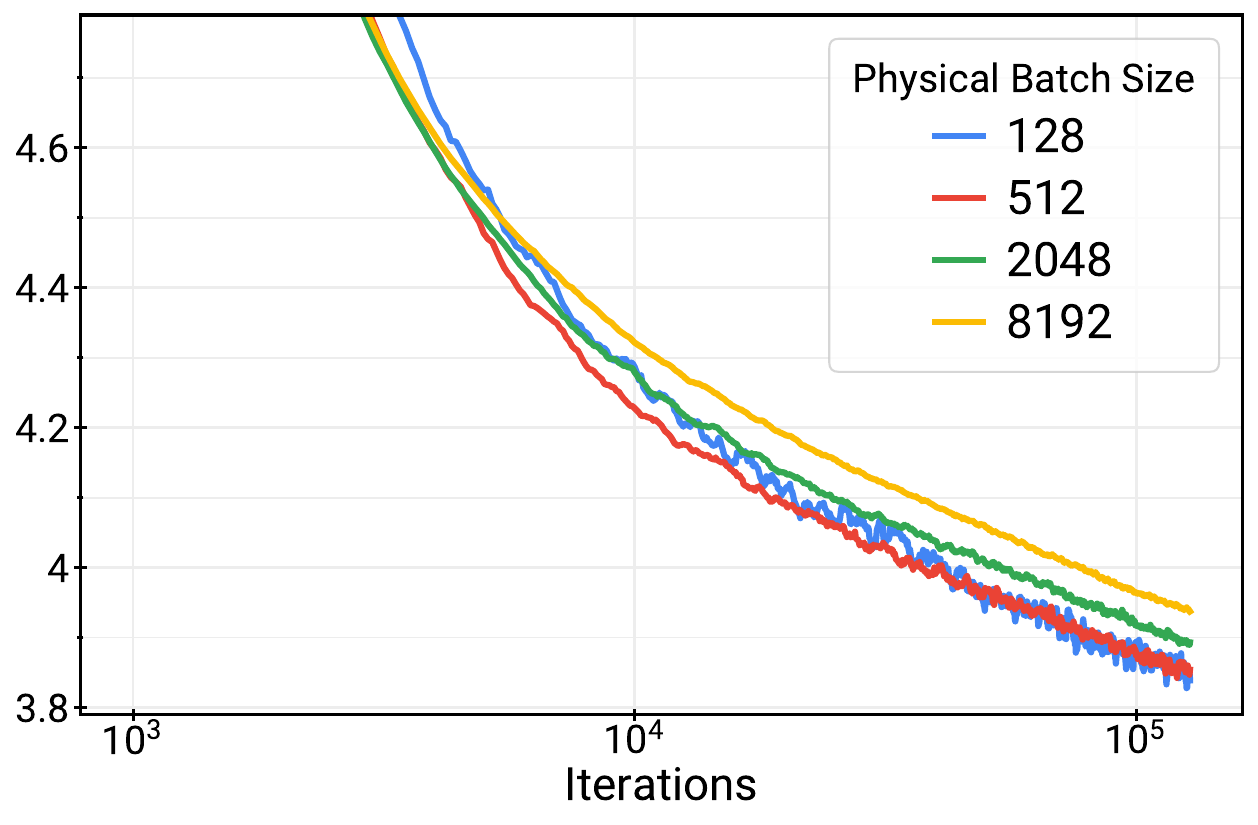}
\caption{\nbr $= 0.5^{15}$} \label{fig:physical_batch_size_medium}
\end{subfigure}
\begin{subfigure}{0.3\linewidth}
\includegraphics[width=\textwidth]{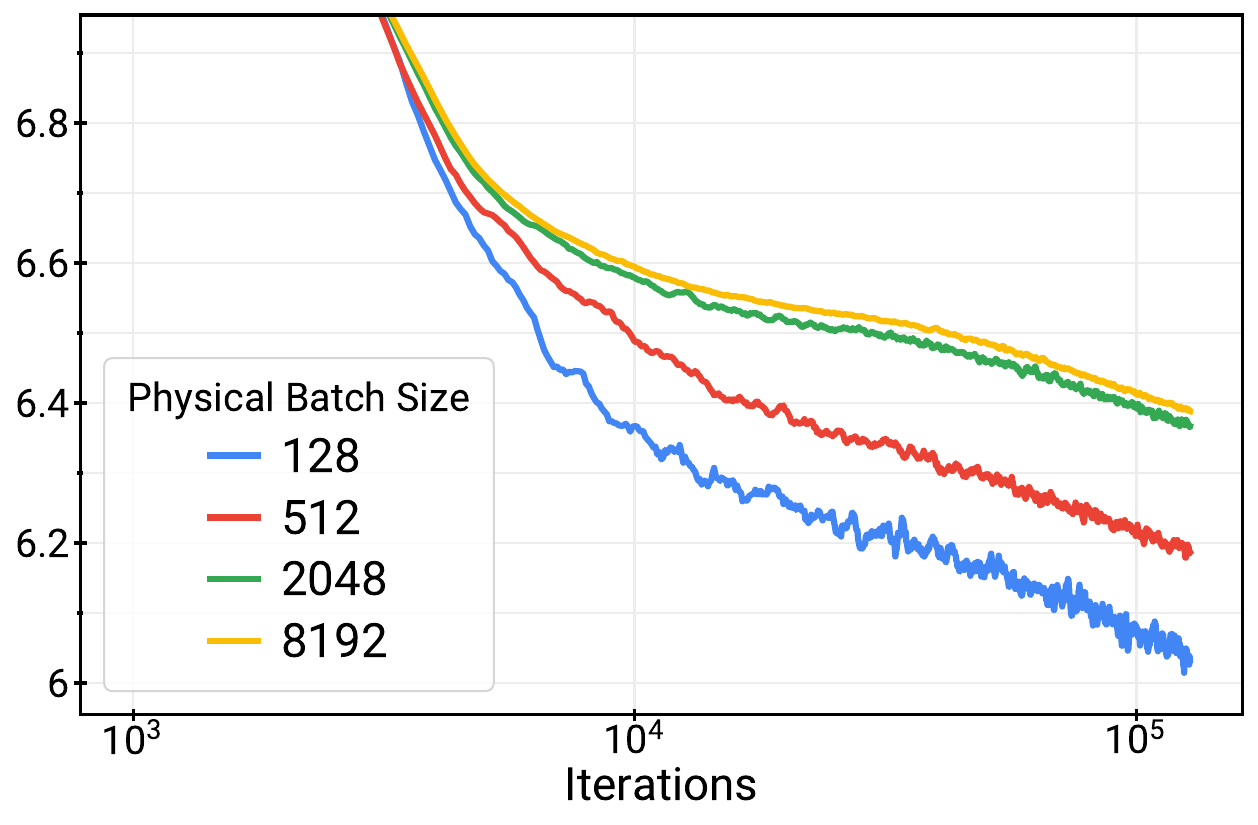}
\caption{\nbr $= 0.5^{10}$} \label{fig:physical_batch_size_large}
\end{subfigure}
\caption{Cross Entropy Loss of \berttiny averaged over $3$ trials for different physical batch sizes and \nbr values.} \label{fig:physical_batch_size_ablation}
\end{figure*}

Central to our methodology is an assumption that for a fixed \nbr, the training curves should be similar for different physical batch sizes. In this section, we conduct ablations to test this hypothesis, and quantify the impact of varying physical batch size under a fixed \nbr. We consider 3 values for \nbr: $0.5^{20}$, $0.5^{15}$, and $0.5^{10}$, and physical batch sizes of $128$, $512$, $2048$, and $8192$. For this ablation we focus on the \berttiny model, which we train for $128$K iterations. We average the losses across three random trials.

The results of this experiment are shown in \cref{fig:physical_batch_size_ablation}. Our primary findings are:

\begin{itemize}[leftmargin=16pt, nosep, itemsep=0.5pt]
  \item At the smallest \nbr in \cref{fig:physical_batch_size_small}, results are as expected. Specifically, larger batch sizes lead to better model performance, but there are diminishing returns. Physical Batch Sizes of $2048$ and $8192$ have nearly identical training curves. 
  \item At the medium and larger \nbr values shown in \cref{fig:physical_batch_size_medium,fig:physical_batch_size_large}, we observe a surprising phenomenon: smaller physical batch sizes lead to models with lower loss. The effect is most prominent in \cref{fig:physical_batch_size_large}. We do not have a good explanation for this behavior, but we did additional experiments to rule out some plausible explanations in \cref{sec:batch_size_ablation_appendix}. Large physical batch sizes ($B=2048$ and $B=8192$) still have very similar learning curves.
\end{itemize}

While the results of this experiment did not fully match expectations, a similar behavior was observed in prior work \cite{sander2023tan} (Figure 4b). Moreover, for sufficiently large batch sizes the training curves are very similar across all \nbr values tested. Thus, we believe that the physical batch size of $1024$ that we use in our main experiments is a reasonable (although not perfect) indicator of what would happen with much larger batch sizes that would be needed to get favorable privacy/utility trade-offs in real-world settings. Understanding when and why thsi behavior manifests is a very interesting direction for future work.

\subsection{Physical Batch Size Ablation - Extended} \label{sec:batch_size_ablation_appendix}

In \cref{sec:batch_size_ablation} we observed a surprising phenomenon where for some values of \nbr, smaller physical batch sizes perform better than larger physical batch sizes. This is in contrast to our initial hypothesis, and our experimental results for very small values of \nbr that larger physical batch sizes should be on par with or better than smaller physical batch sizes for the same \nbr. 

While we do not have a great explanation for the observed phenomenon, we have ruled out several possible explanations, which we discuss below:

\begin{enumerate}
  \item \textbf{Learning Rate Tuning}. While our main experiment used a fixed learning rate of $0.5^8$ across all values of \nbr, we ran further experiments for a \nbr of $0.5^{15}$ with four different learning rates ($0.5^6, 0.5^7, 0.5^8, 0.5^9$), and report the best cross entropy across all learning rates on a per-iteration basis. Even with learning rate tuning, the conclusion is the same: smaller physical batch sizes achieve lower loss than larger ones (see \cref{fig:physical-batch-size-sweep-loss}).
  
  \begin{figure}[ht]
    \centering
    \includegraphics[width=0.5\linewidth]{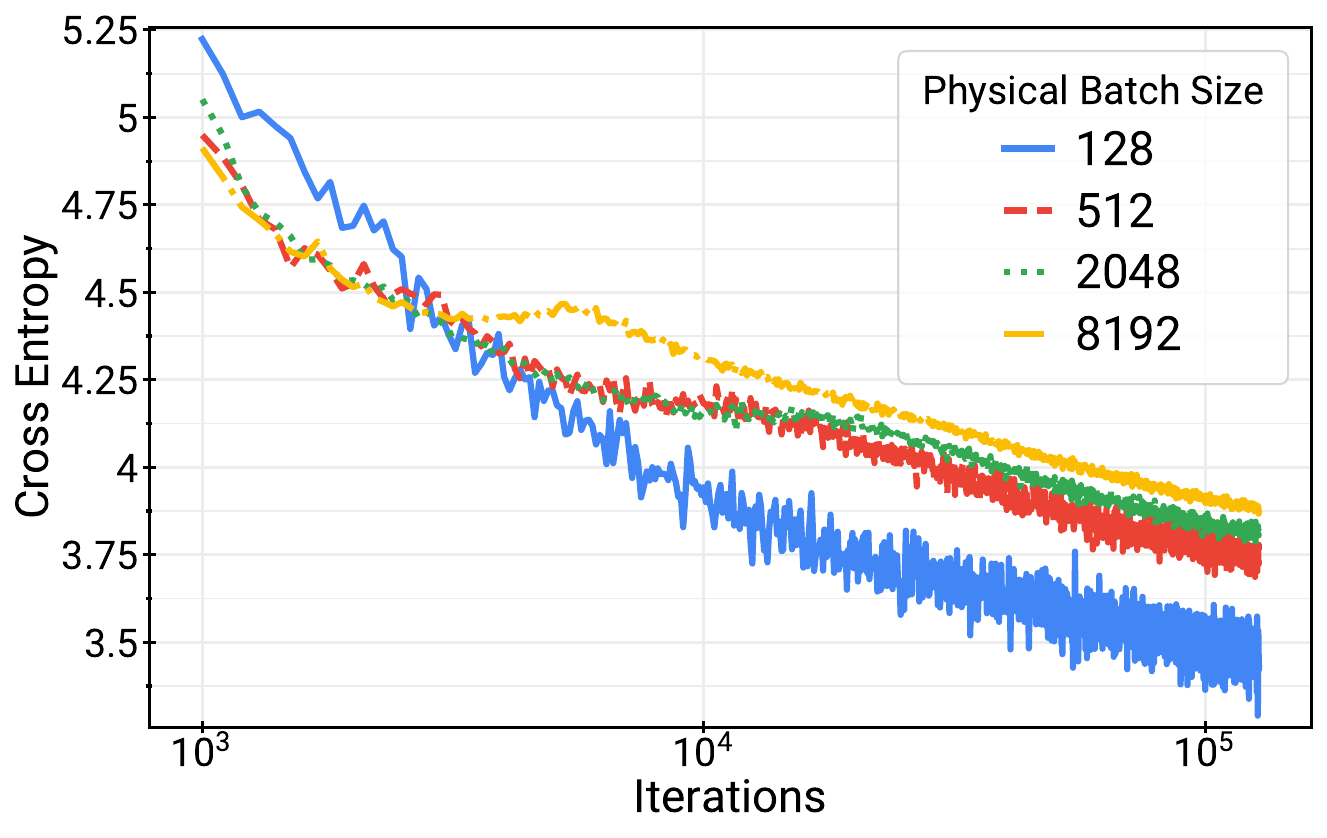}
    \caption{Smaller physical batch sizes achieve lower loss than larger ones.}
    \label{fig:physical-batch-size-sweep-loss}
  \end{figure}
  
  \item \textbf{Differences in Train / Eval Loss}. Our main experiment measures the training loss, but since the loss is computed before incorporating the gradient into the model, and because we train for less than one pass over the entire dataset, this is an unbiased estimate of the evaluation loss. It is natural to ask which models have lower final loss on the training set (after incorporating those examples into the model). To test whether lower physical batch sizes somehow generalize better, or whether they also do better on the training loss, we measured the loss of the final trained model on $1$ million examples from the training set. We focus on the \nbr of $0.5^{15}$ in this test. The table below shows that smaller physical batch sizes also have better performance on the already-seen training examples, ruling out this explanation (see \cref{tab:physical-batch-size-sweep-loss-entire-train-set}).
  
  \begin{table}[ht]
  \centering
  \begin{tabular}{c|cc}
  \toprule
    & \multicolumn{2}{c}{Training Set} \\
    Batch Size & Cross Entropy & Accuracy \\\hline
    128 & 3.586 & 43.59\% \\
    512 & 3.971 & 37.27\% \\
    2048 & 4.01 & 37.55\% \\
    8192 & 4.057 & 36.73\% \\
    \bottomrule
  \end{tabular}
  \caption{Loss over the entire training set is also better for lower physical batch sizes.}
  \label{tab:physical-batch-size-sweep-loss-entire-train-set}
  \end{table}
  \FloatBarrier
  
  \item \textbf{Model Size}. The main experiment uses \berttiny, which is a relative small model. It is natural to ask whether the same behavior would be observed for a larger model like \bertbase. The figure below shows that the same phenomenon happens for \bertlarge, but only for the largest \nbr. The other two values of \nbr do not exhibit this behavior, although at the middle \nbr, the trend line suggests there may be a crossover point beyond the limits of the x axis. Thus, increasing model size seem to influence and mitigate this behavior, but not eliminate it completely. See \cref{fig:physical-batch-size-sweep-model-size}.
  
  \begin{figure*}[h!]
  \begin{subfigure}{0.31\linewidth}
  \includegraphics[width=\textwidth]{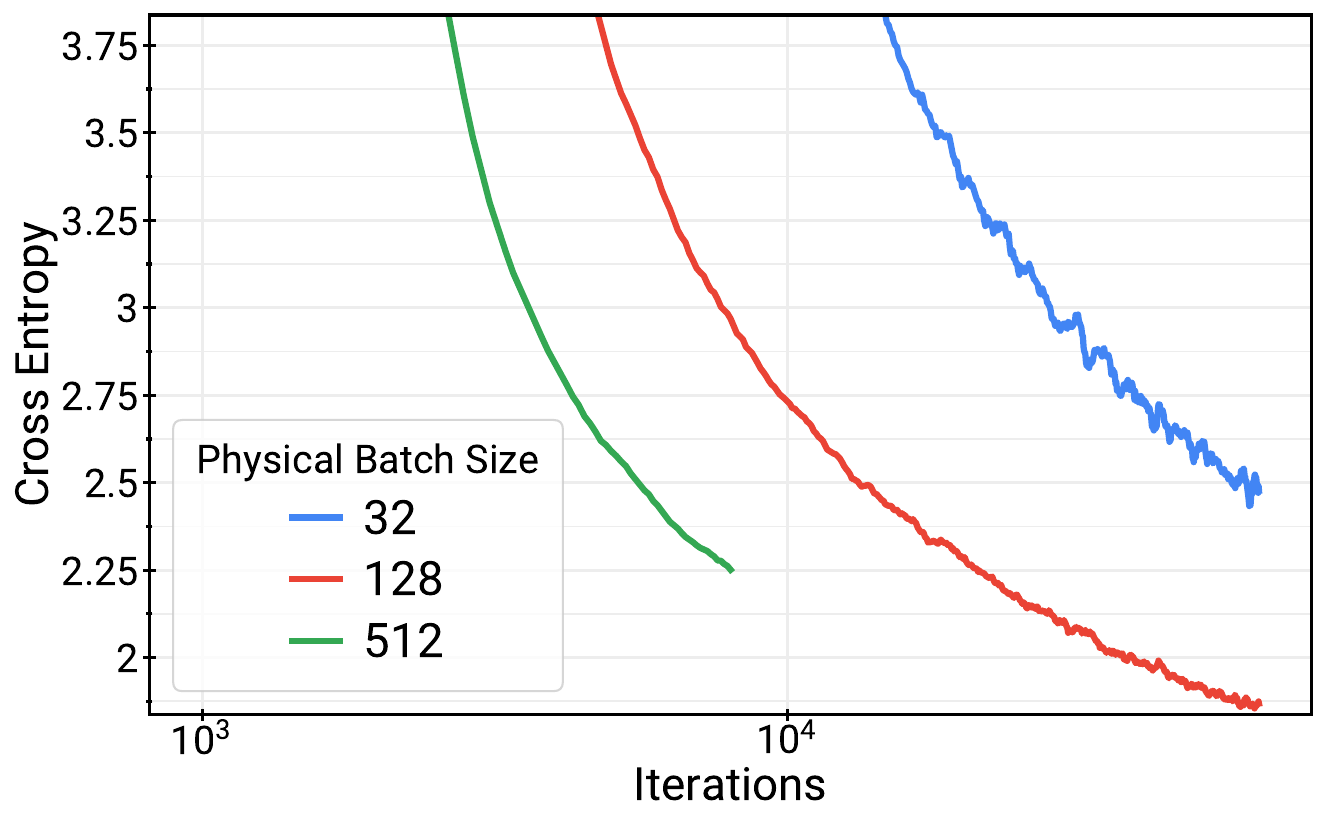}
  \caption{\nbr $= 0.5^{20}$}
  \end{subfigure}
  \begin{subfigure}{0.3\linewidth}
  \includegraphics[width=\textwidth]{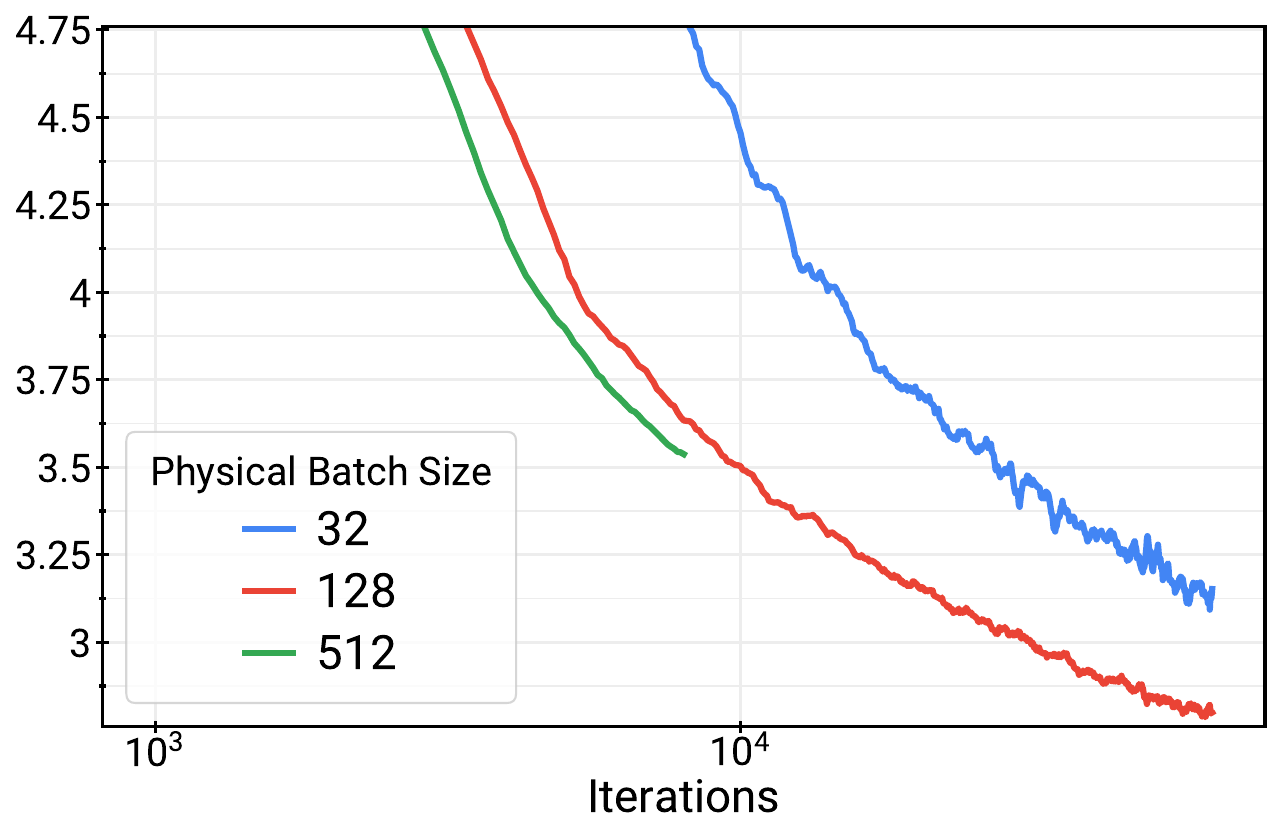}
  \caption{\nbr $= 0.5^{15}$}
  \end{subfigure}
  \begin{subfigure}{0.3\linewidth}
  \includegraphics[width=\textwidth]{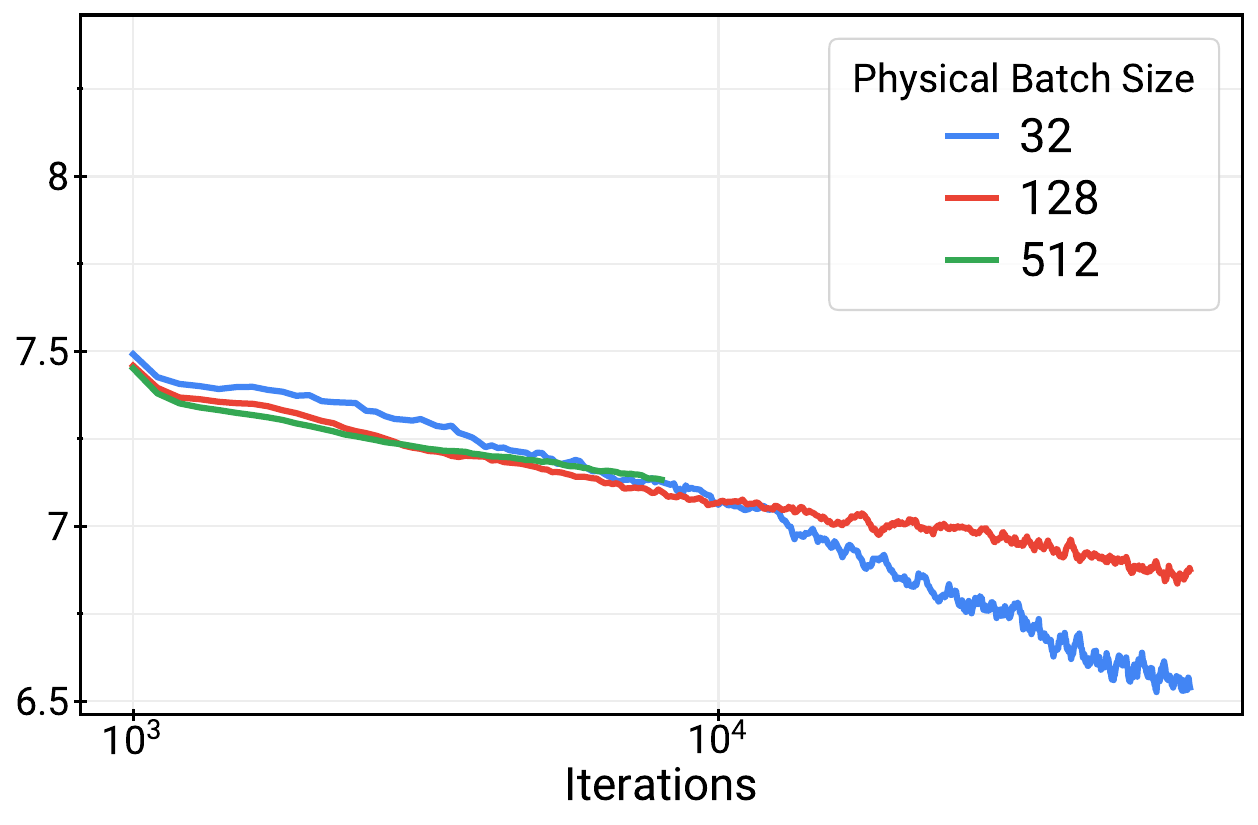}
  \caption{\nbr $= 0.5^{10}$}
  \end{subfigure}
  \caption{Cross Entropy Loss of \bertlarge averaged over $3$ trials for different physical batch sizes and \nbr values.}
  \label{fig:physical-batch-size-sweep-model-size}
  \end{figure*}
  \FloatBarrier
  
  \item \textbf{Training Pipelines}. It is natural to question whether this behavior is explained by some bug in the training pipeline. We carefully reviewed the implementation and did not find any bugs that could explain this behavior, and also did additional experiments on a totally separate training pipeline based on \texttt{NanoDO}~\citep{nanodo}, where we observed the same qualitative behavior when training a $30$ million parameter decoder-only transformer model with \dpadam for 32K iterations. The figures below show the smoothed cross entropy averaged over 3 random trials. 
  
  \begin{figure*}[h!]
  \includegraphics[width=\textwidth]{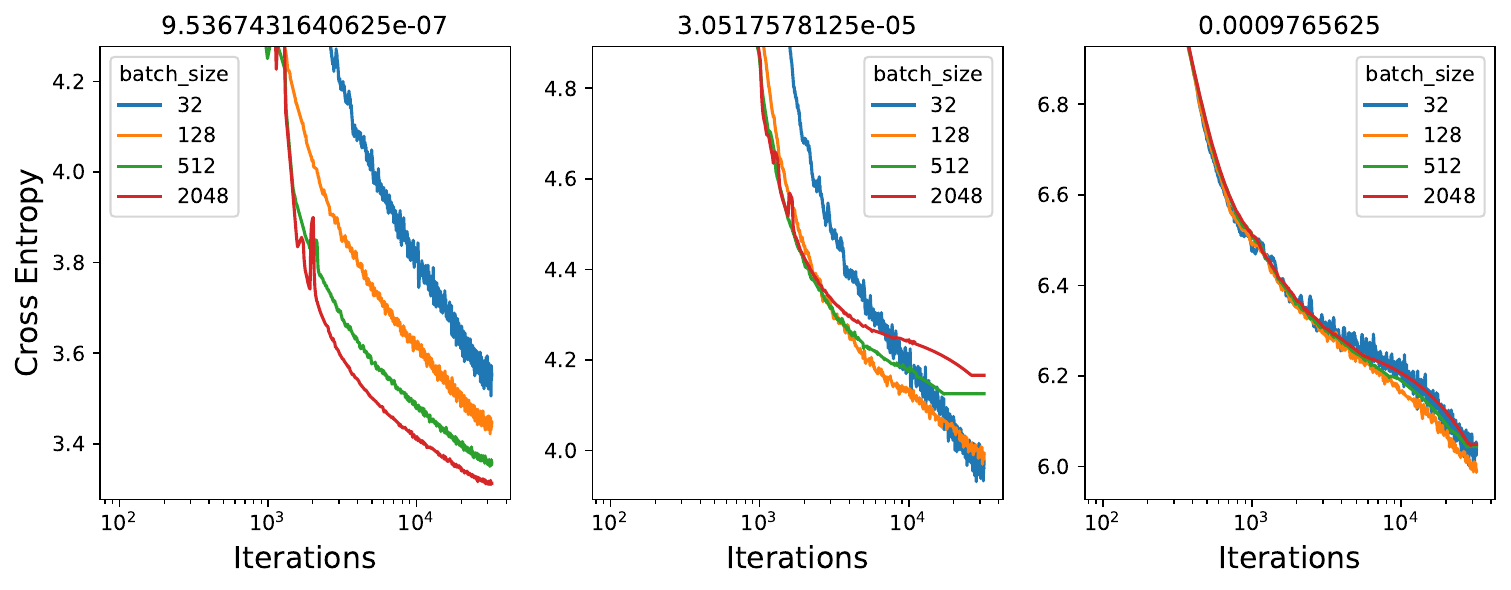}
  \caption{Loss on NanoDO~\citep{nanodo}.}
  \label{fig:physical-batch-size-sweep-nanodo}
  \end{figure*}
  \FloatBarrier
  
\end{enumerate}

\subsection{Training Throughput}
\label{app:training}

By looking at intermediates, using a single physical batch size, and separating the accounting from the experimentation we greatly reduce the number of experiments to run. However, the set of experiments we outline above is still very compute-intensive. We utilize TPUv3 pods to run all experiments, and configured the models to use pure data parallelism, using more cores for larger models so that each experiment finishes within four to ten hours. \berttiny was trained on $16$ TPUv3 cores, while \bertlarge was trained on 128. \cref{tab:throughput} provides the training throughputs for all models in our experiments.

\begin{table}[h!]
\centering
\begin{tabular}{@{}lcccc@{}}
\toprule
\textbf{Model} & \textbf{Params} & \textbf{Steps/sec} & \textbf{Per Core Batch Size} & \textbf{Records / Sec} \\ 
\midrule
\berttiny & 4.52M & 8.959 & 64 & 573 \\ 
\bertmini & 11.4M & 5.494 & 64 & 352 \\ 
\bertsmall & 29.0M & 6.602 & 32 & 211 \\ 
\bertmedium & 41.6M & 4.196 & 32 & 134 \\ 
\bertbase & 110M & 3.621 & 16 & 54 \\ 
\bertlarge & 335M & 2.225 & 8 & 17.8 \\ 
\bertmega & 729M & 1.536 & 4 & 6.1 \\ 
\bottomrule
\end{tabular}
\caption{Training throughput for various \bert models}
\label{tab:throughput}
\end{table}
\FloatBarrier

\subsection{Reproducing non-private scaling laws results}

We now confirm that the experimental data we collected matches the expected behavior of \citet{hoffmann2022training}, specifically that in the absence of noise, the optimal model size and tokens should grow in roughly equal proportion with increasing compute budget. This is true despite our several methodological differences, including: (1) doing per-example gradient clipping, (2) using a different optimizer and not retraining for each number of iterations, (3) using a large physical batch size, etc. The exact Token / Model ratio predicted here is larger than prior work, but that is well explained by the fact that a batch size of $1024$ examples is well beyond the critical batch size of compute-efficient training \cite{mccandlish2018empirical}.

\begin{figure}[h!]
  \centering 
  \includegraphics[width=\textwidth]{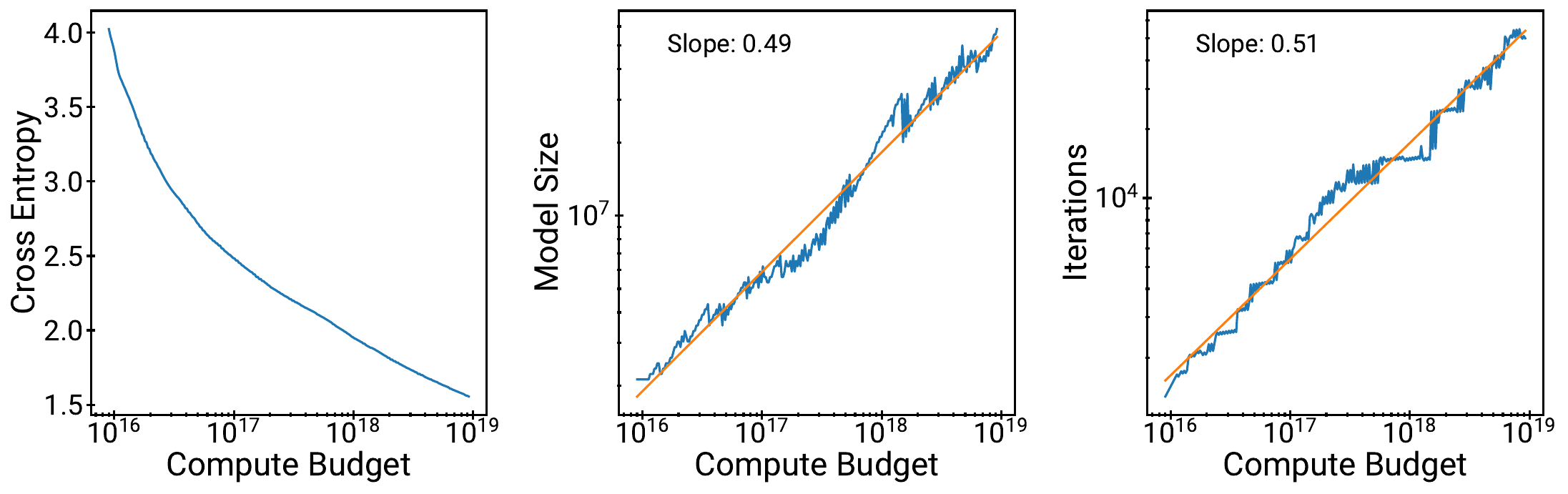}
  \caption{Compute-optimal cross entropy, model size, and number of iterations when running DP-Adam with $\sigma=0$.}
  \label{fig:my_label}
\end{figure}

\subsection{Optimal Learning Rates} \label{sec:learning_rates}

\begin{figure*}[h!]
\begin{subfigure}{0.3\linewidth}
\includegraphics[width=\textwidth]{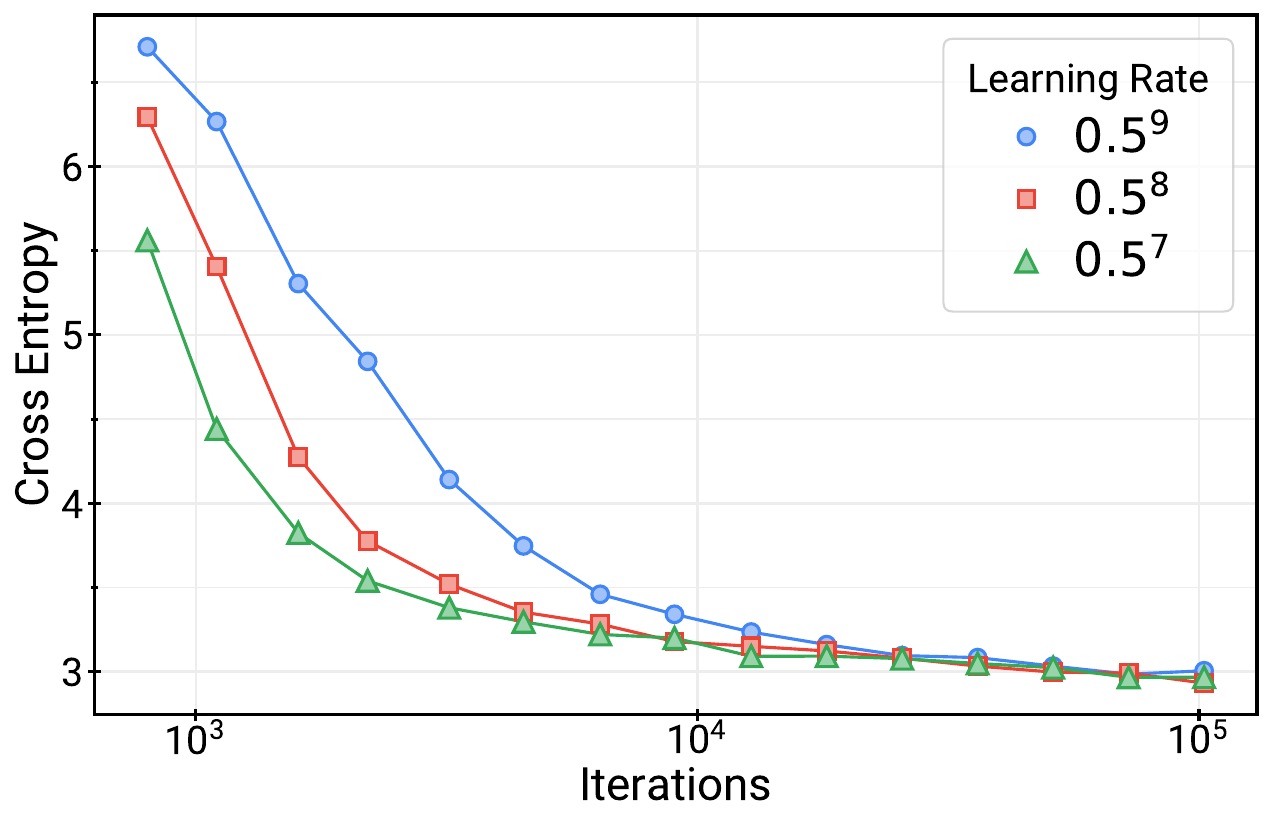}
\includegraphics[width=\textwidth]{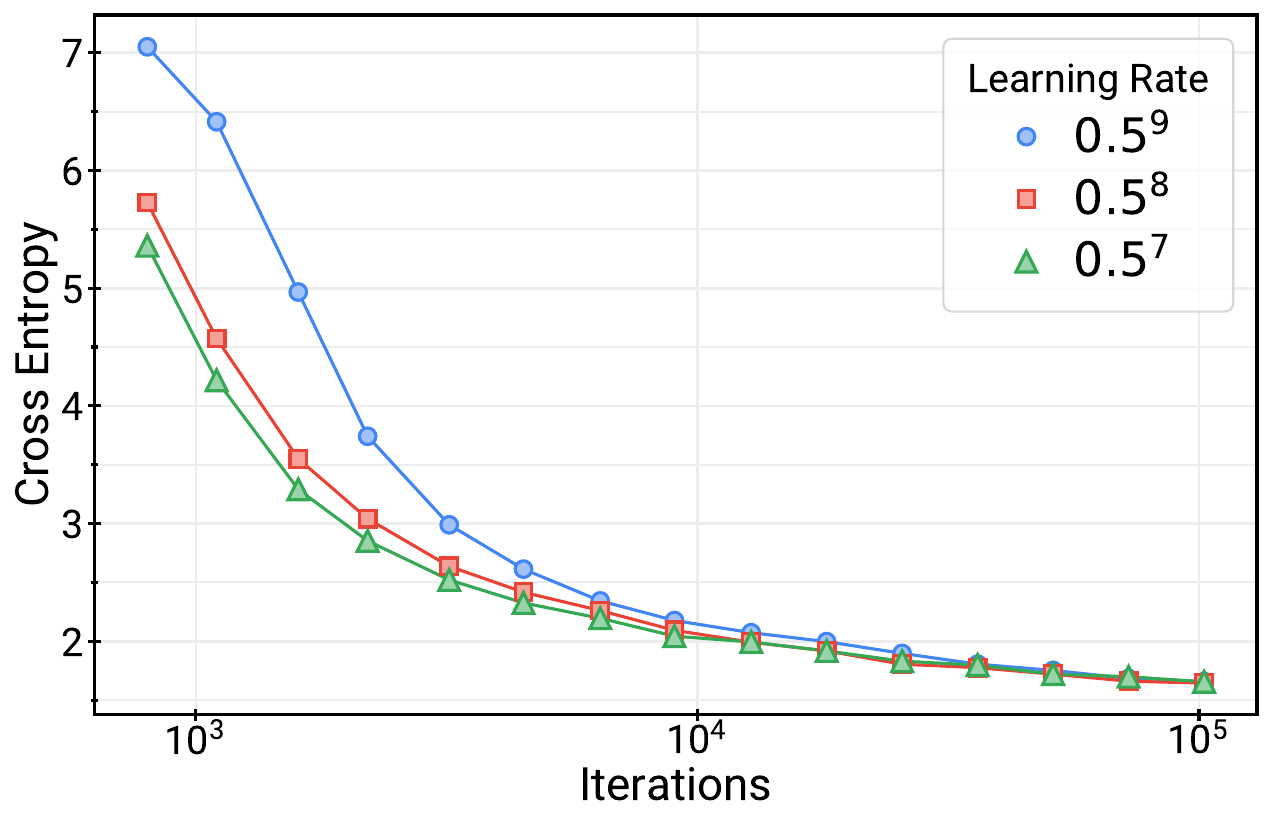}
\caption{\nbr = $0.5^{20}$}
\end{subfigure}
\begin{subfigure}{0.3\linewidth}
\includegraphics[width=\textwidth]{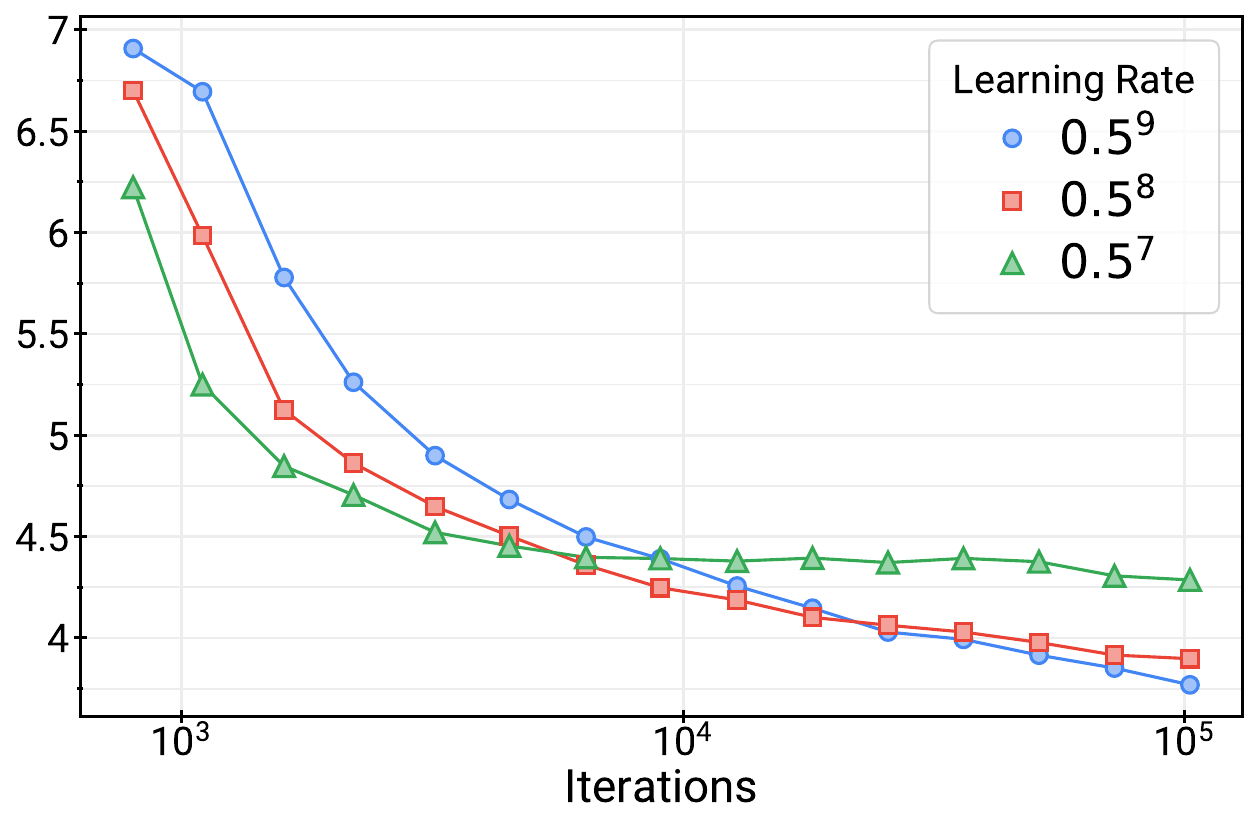}
\includegraphics[width=\textwidth]{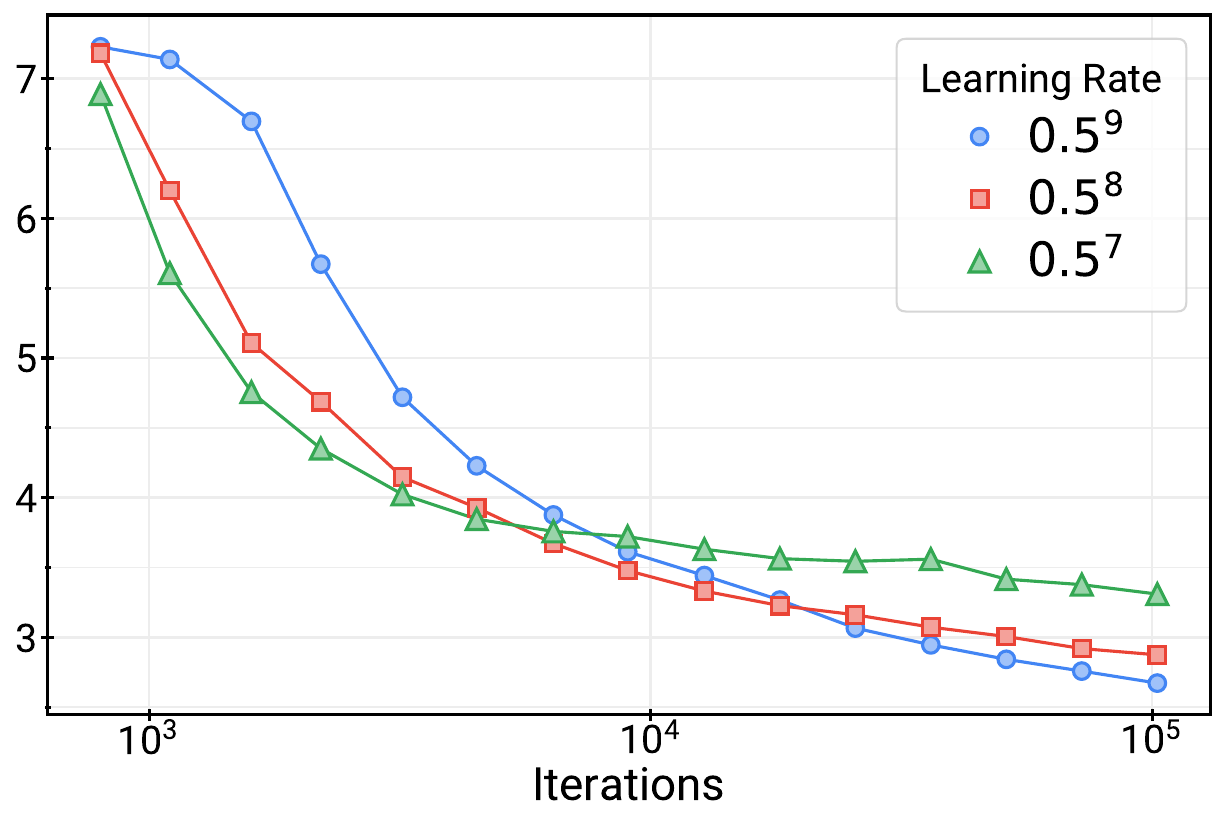}
\caption{\nbr = $0.5^{15}$}
\end{subfigure}
\begin{subfigure}{0.3\linewidth}
\includegraphics[width=\textwidth]{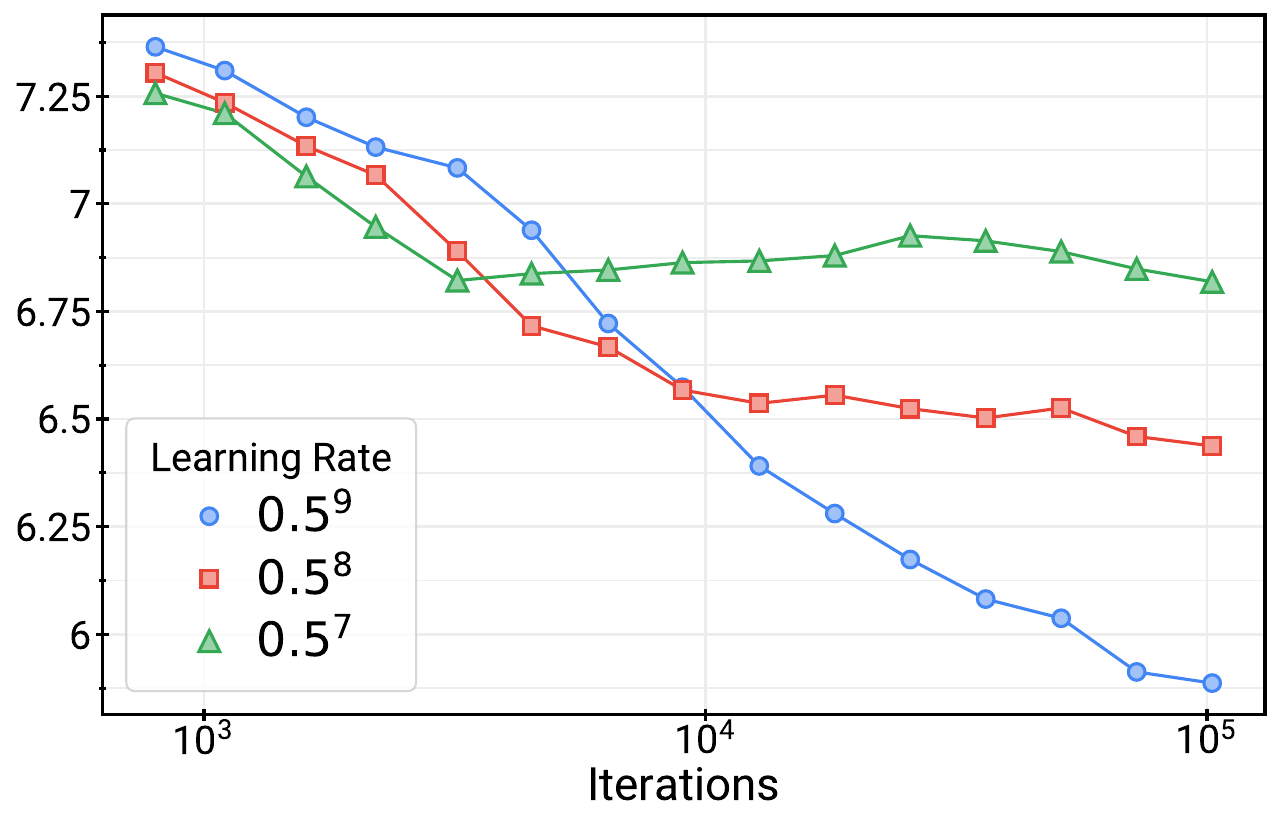}
\includegraphics[width=\textwidth]{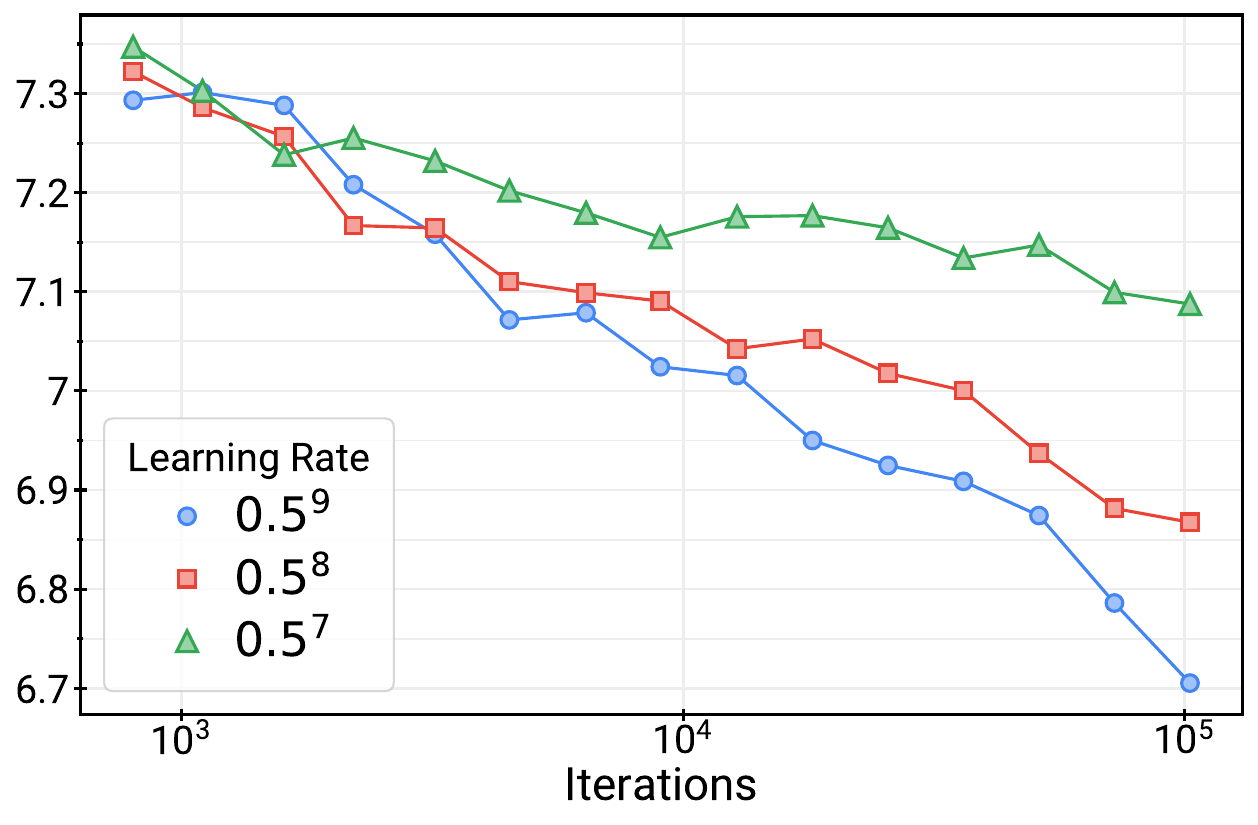}
\caption{\nbr = $0.5^{10}$}
\end{subfigure}
\caption{Training curves for \berttiny (top) and \bertmedium (bottom) with varying learning rates at different \nbr values.} \label{fig:training}
\end{figure*}

We now look at the training curves for different learning rates and different \nbr values. These results generally match expectations and demonstrate that the learning rates we chose were selected from the correct regime.

\FloatBarrier

\subsection{Optimal Compute Budget Allocation} \label{sec:appendix:extra_compute_budget_allocation}

\begin{figure*}[h!]
\begin{subfigure}{0.33\linewidth}
\includegraphics[width=\textwidth]{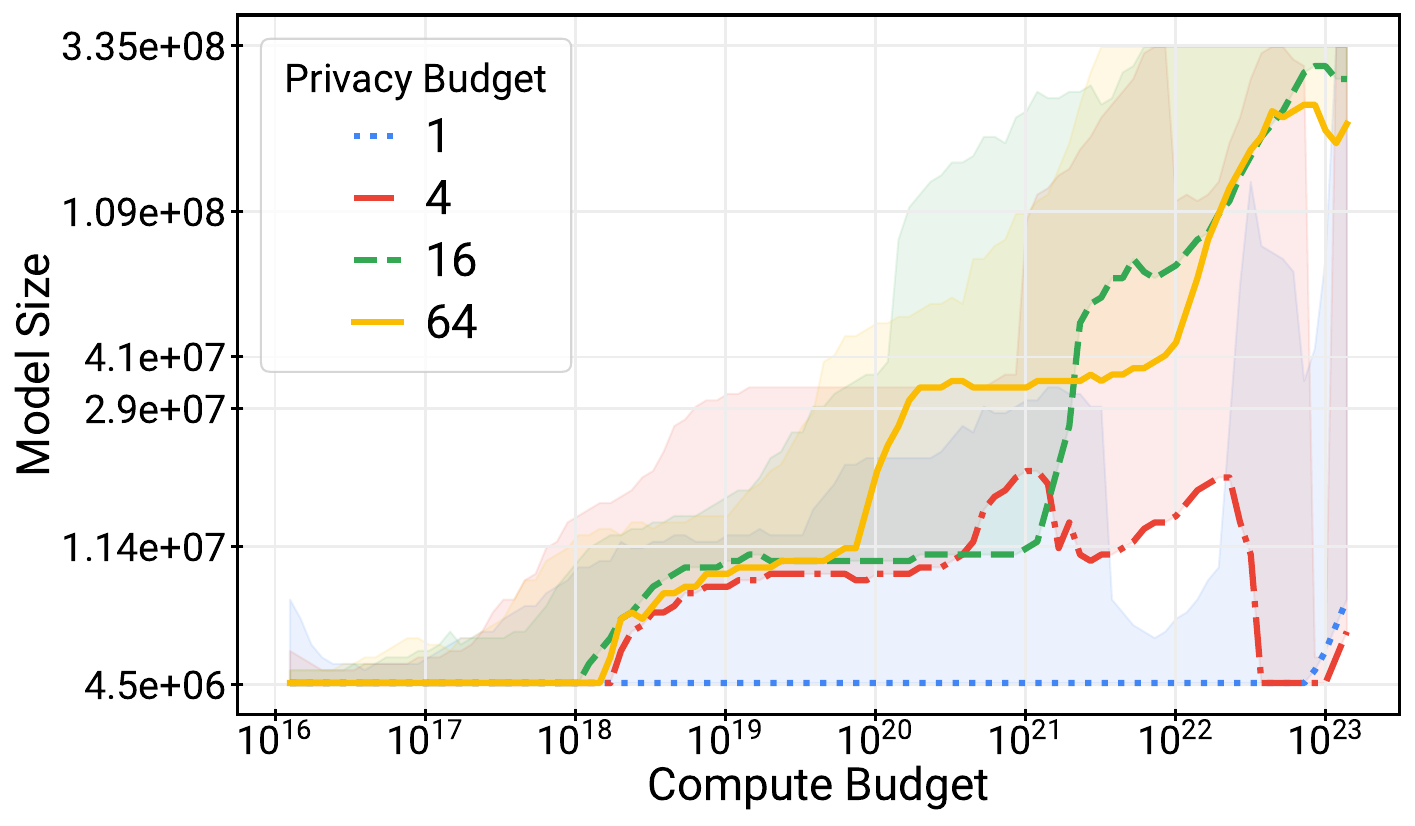}
\includegraphics[width=\textwidth]{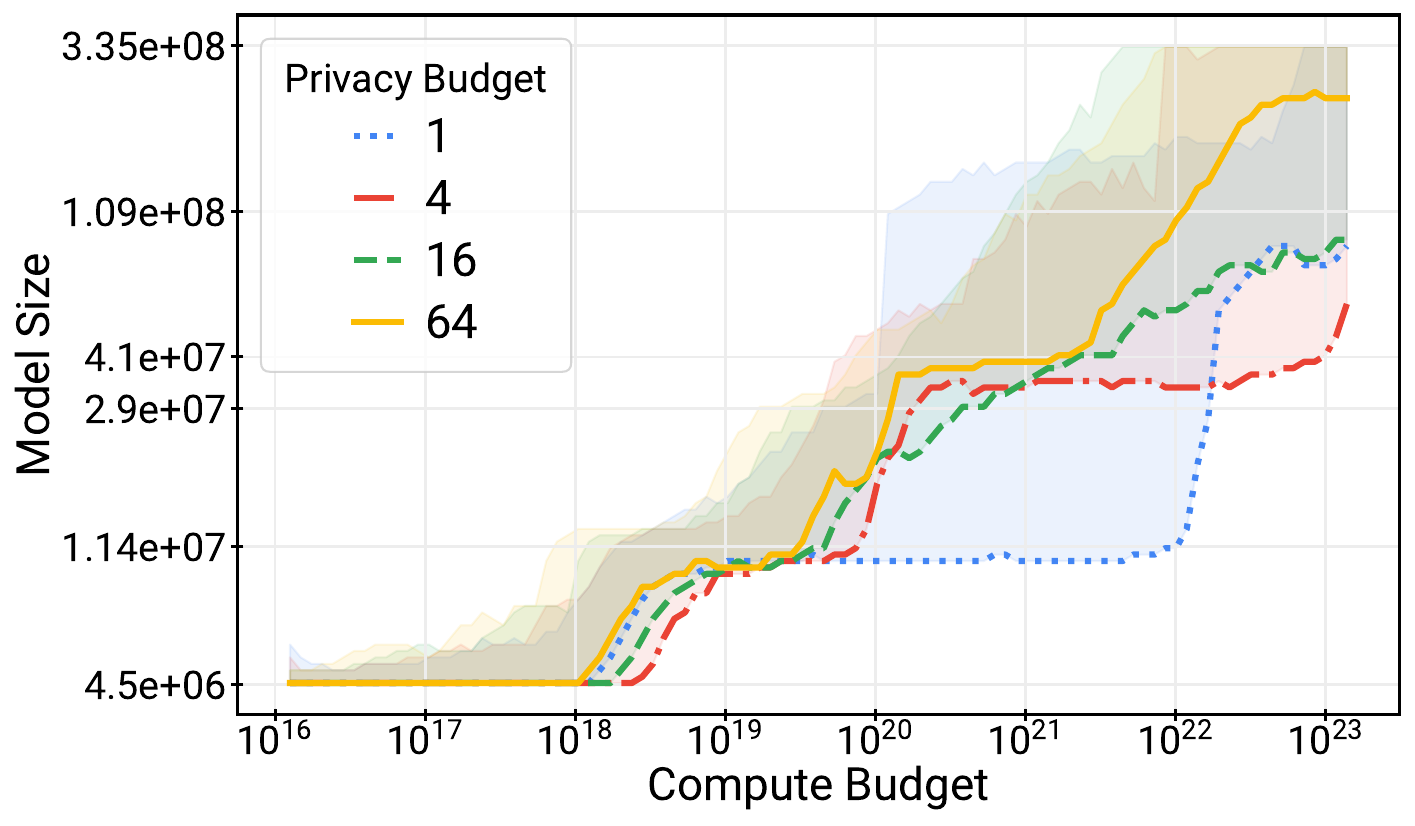}
\includegraphics[width=\textwidth]{plots/optimal_Model_Size_100000000users.pdf}
\includegraphics[width=\textwidth]{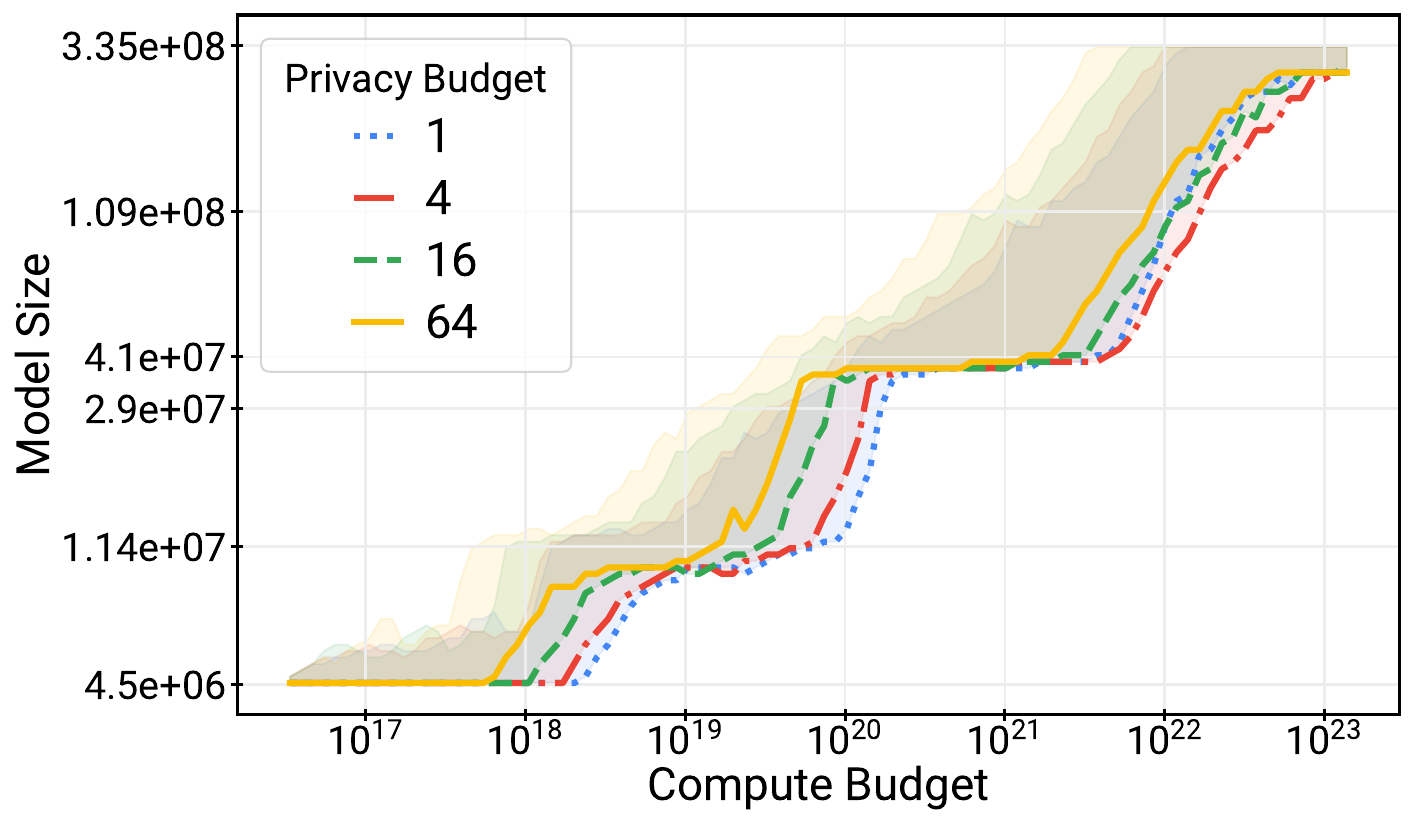}
\caption{Model Sizes}
\end{subfigure}
\begin{subfigure}{0.32\linewidth}
\includegraphics[width=\textwidth]{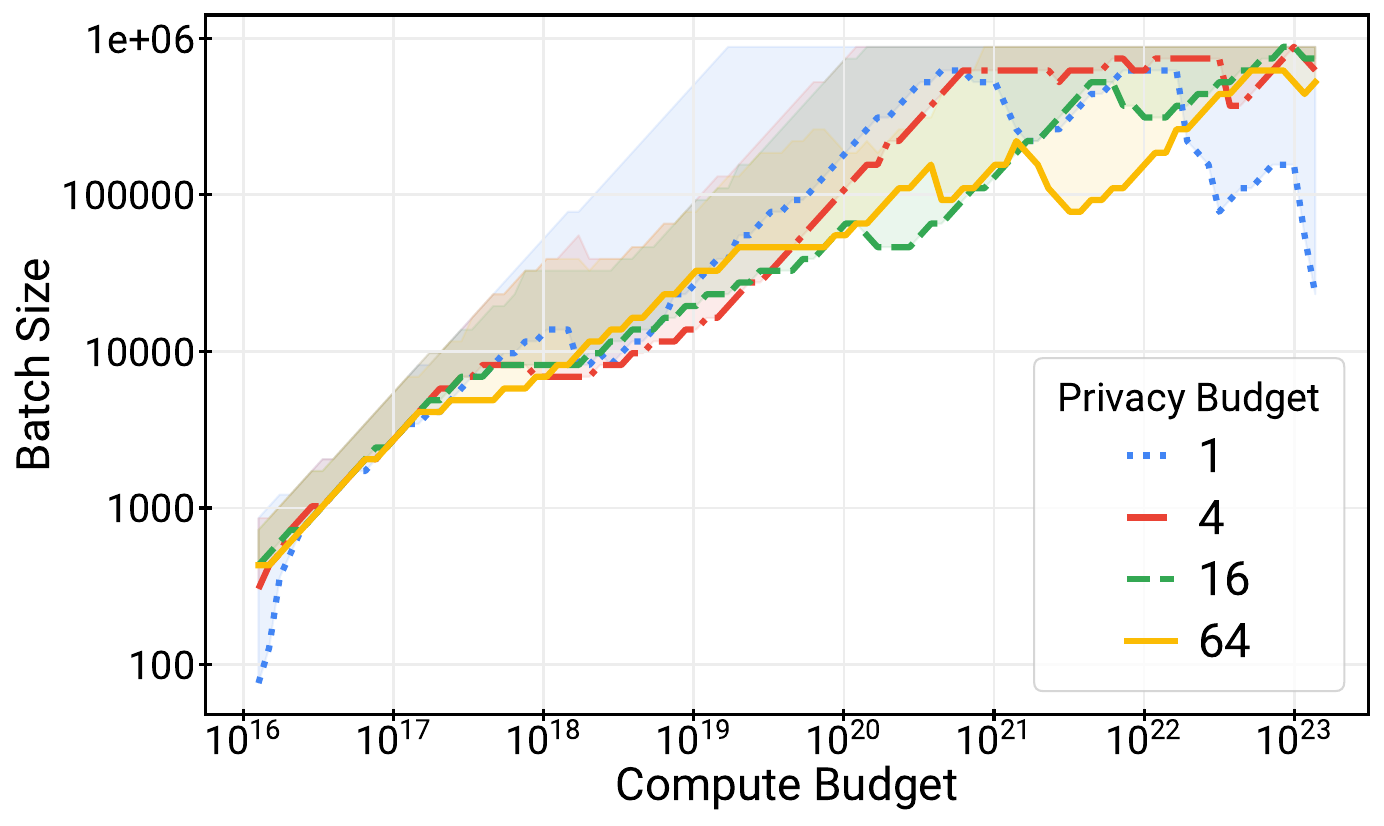}
\includegraphics[width=\textwidth]{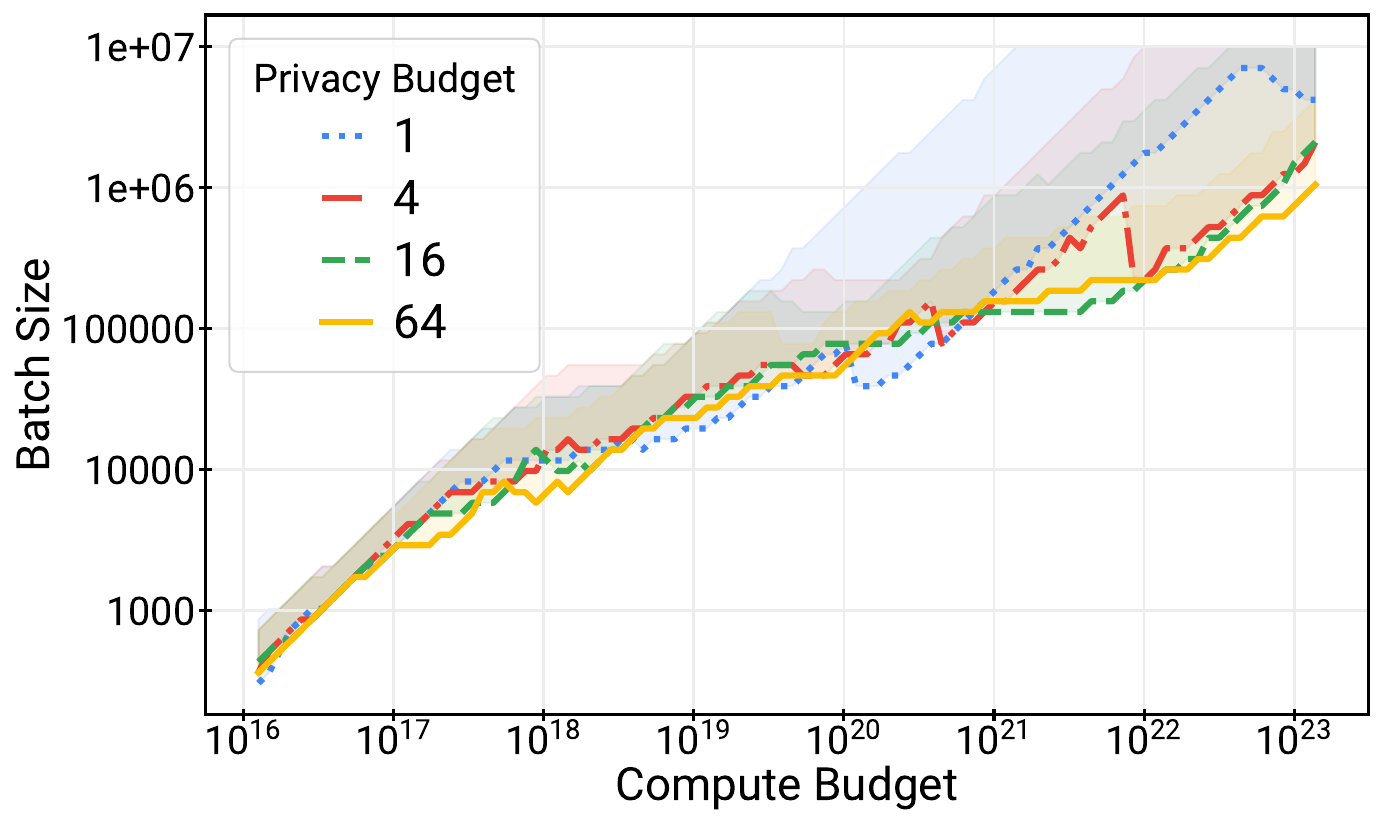}
\includegraphics[width=\textwidth]{plots/optimal_Batch_Size_100000000users.pdf}
\includegraphics[width=\textwidth]{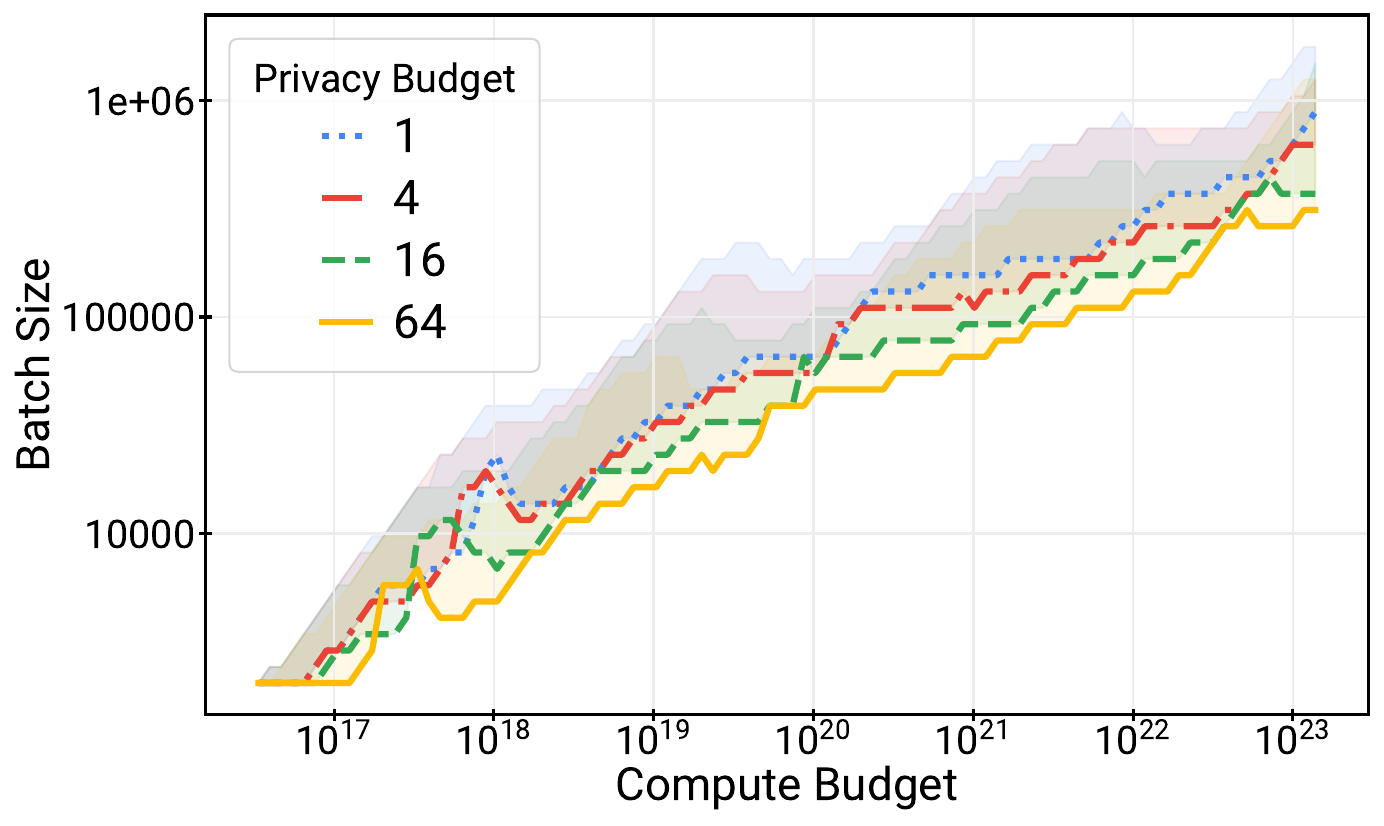}
\caption{Batch Size}
\end{subfigure}
\begin{subfigure}{0.32\linewidth}
\includegraphics[width=\textwidth]{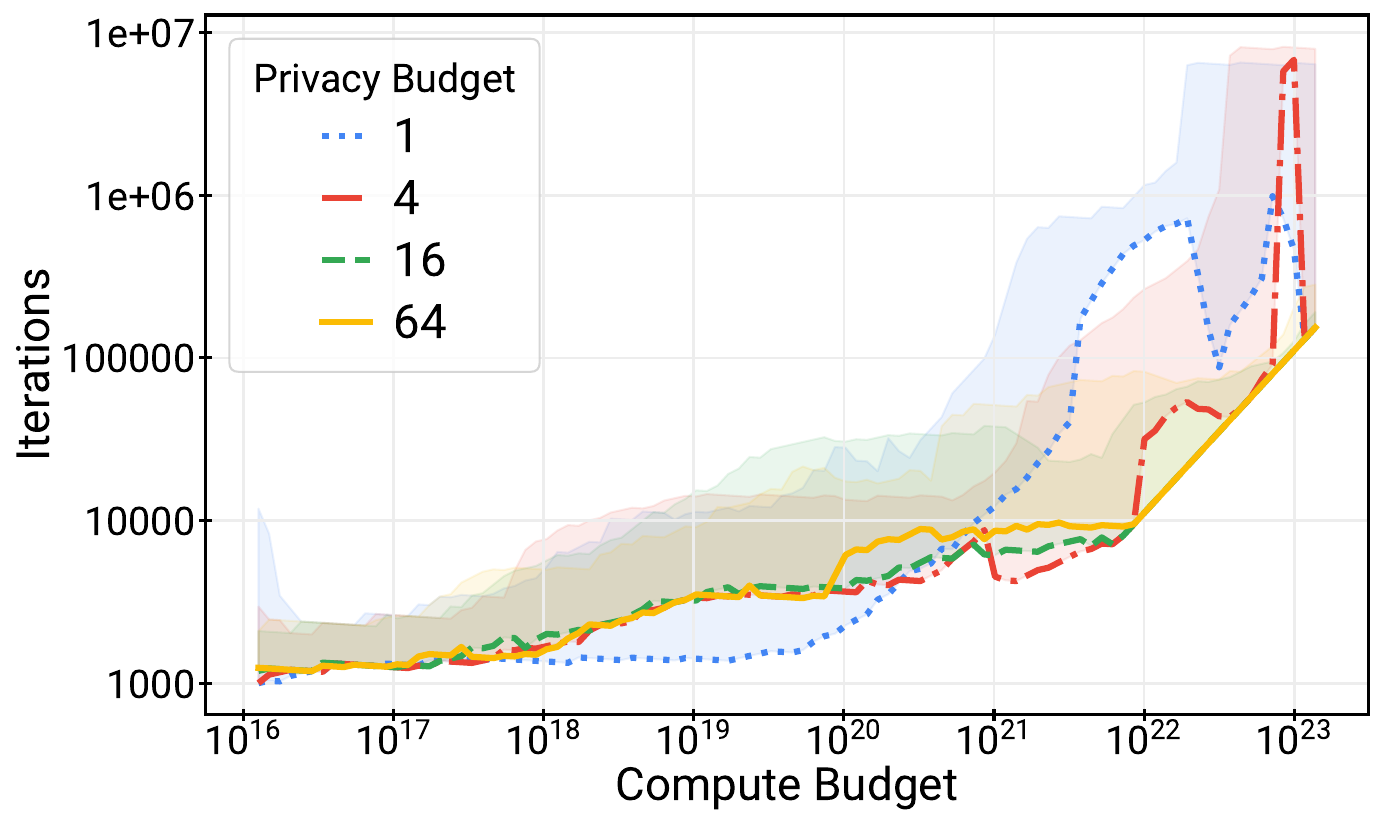}
\includegraphics[width=\textwidth]{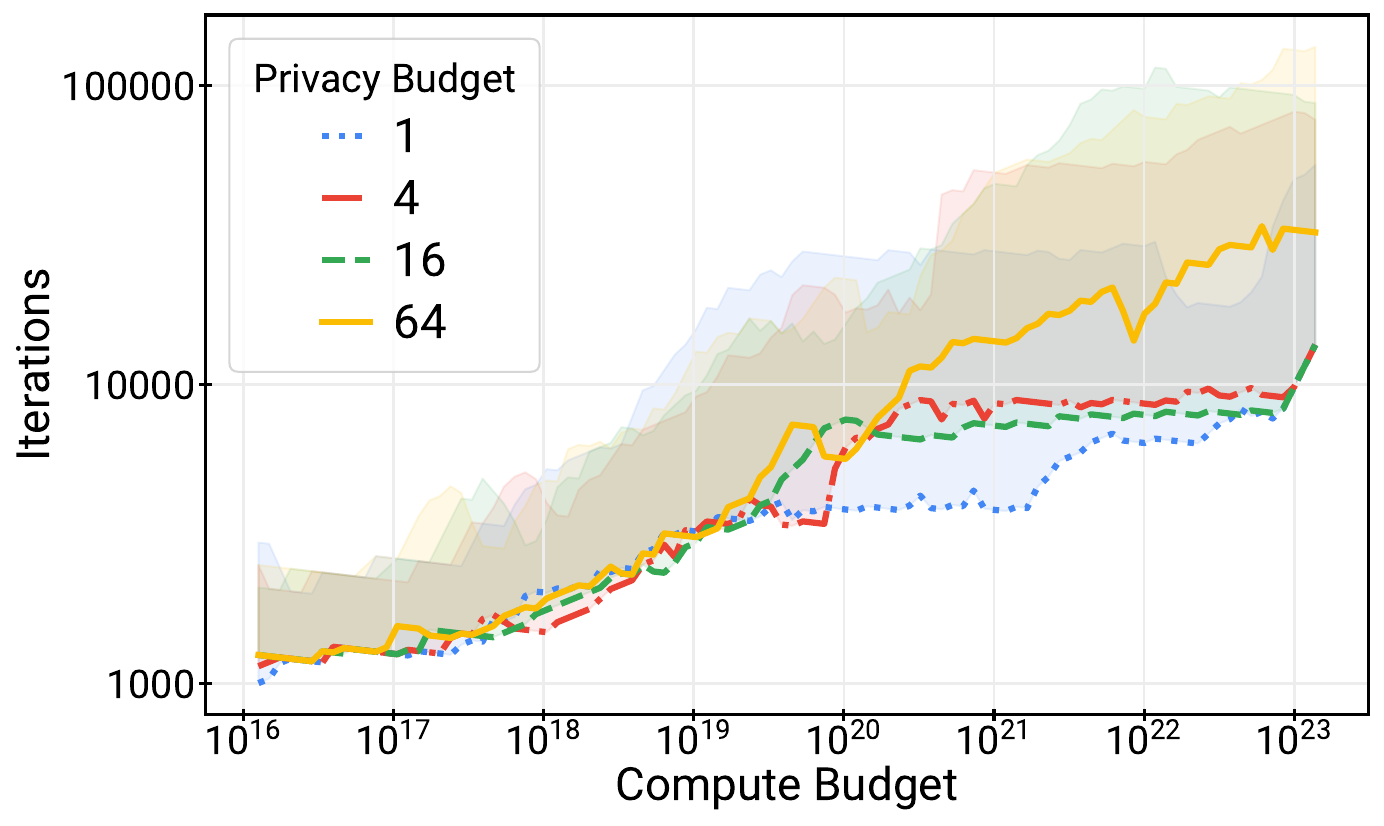}
\includegraphics[width=\textwidth]{plots/optimal_Iterations_100000000users.pdf}
\includegraphics[width=\textwidth]{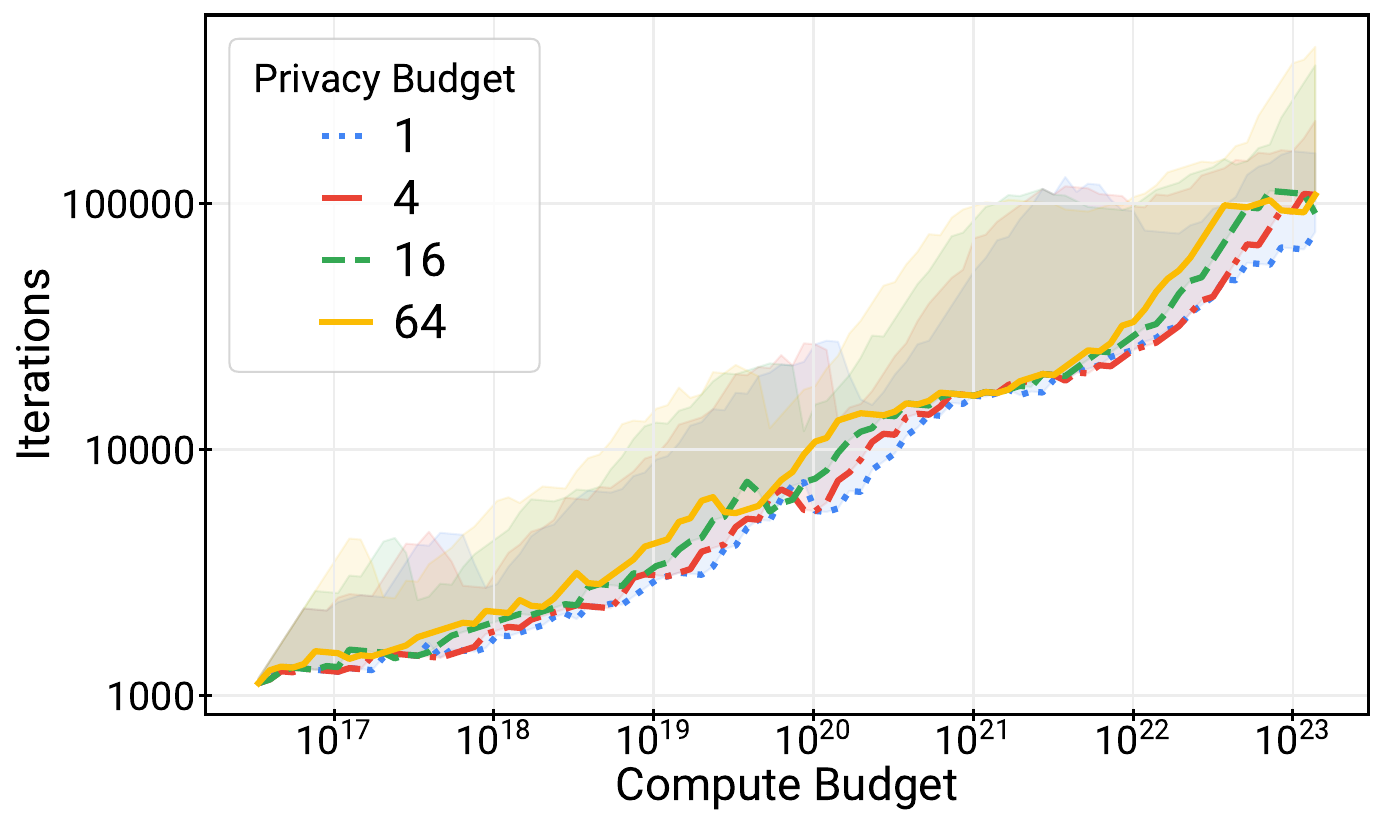}
\caption{Iterations}
\end{subfigure}
\caption{Compute optimal model-sizes, batch sizes, and iterations for varying privacy budgets and compute budgets, and data budgets. Each row of plots corresponds to a different data budget of $\numsamples=10^6, 10^7, 10^8$, and $10^9$ respectively. Each line corresponds to the minimum value of that hyper-parameter that achieves within 1\% of the optimal cross entropy across all constant-compute training configurations. The shaded region corresponds to the full range of possible values for that hyper-parameter that are optimal to within 1\%.} 
\label{fig:optimal_allocation_extended}
\end{figure*}

In this section, we extend the results from \cref{sec:compute_budget_allocation}, including results for more settings of the data budget, ranging from $\numsamples=10^6$ to $\numsamples=10^9$. The full results are shown in \cref{fig:optimal_allocation_extended}.  Our findings are qualitatively similar to the ones we identified in the main text across different data budgets, but the precise constants may differ. 

\subsection{Smoothing and Extrapolation} \label{sec:smoothing}

\begin{figure*}[ht!]
\begin{subfigure}{0.48\linewidth}
\includegraphics[width=\textwidth]{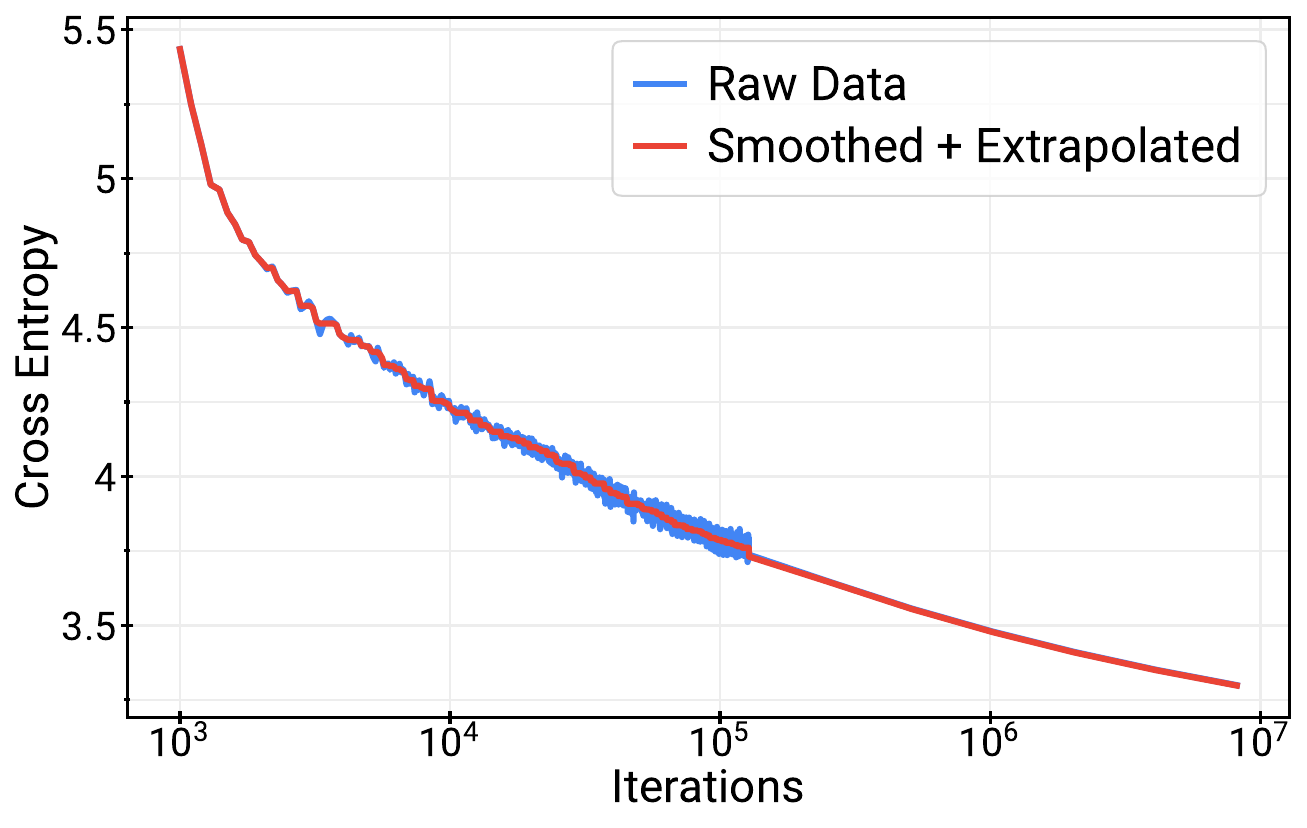}
\caption{\nbr $=0.5^{15}$}
\end{subfigure}
\begin{subfigure}{0.48\linewidth}
\includegraphics[width=\textwidth]{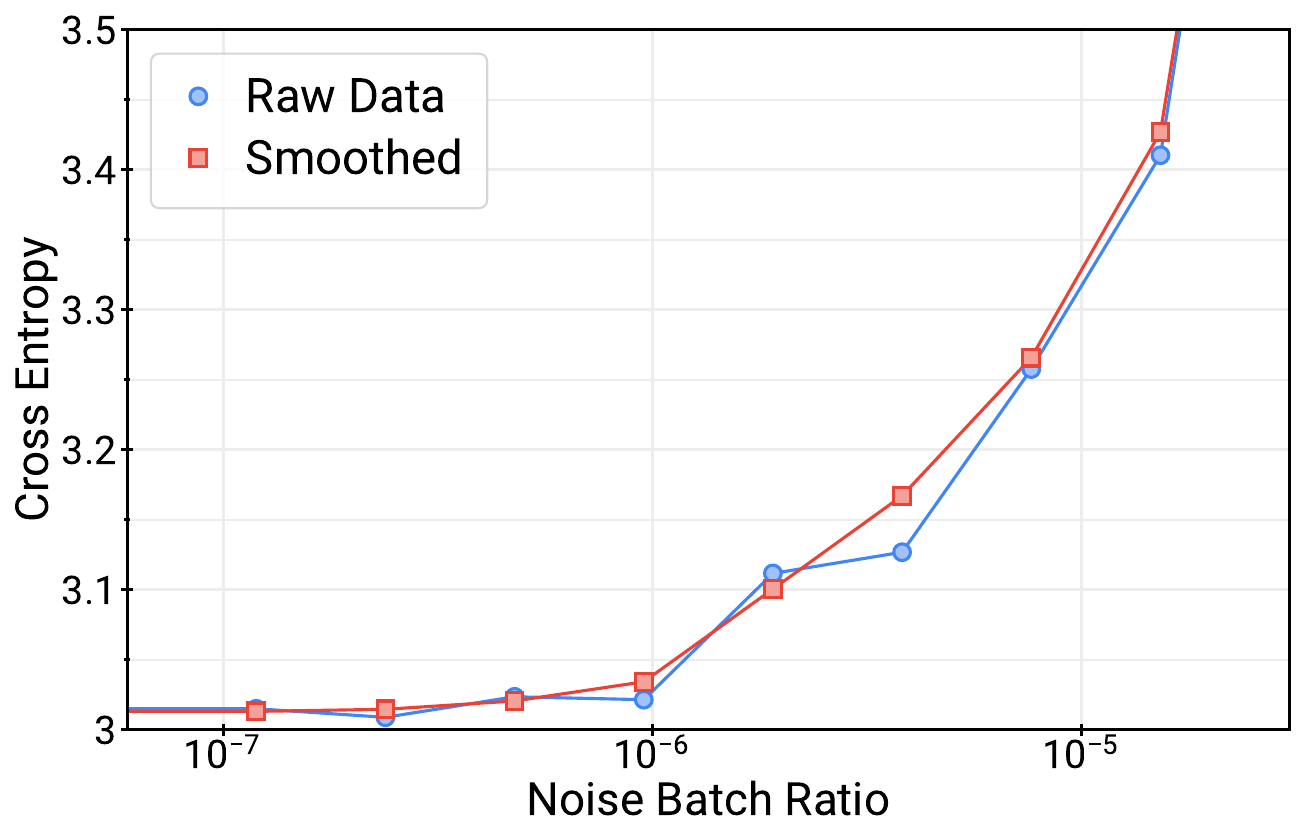}
\caption{$T=32000$}
\end{subfigure}
\caption{Demonstration of our semi-parametric smoothing on \berttiny.} \label{fig:smoothing}
\end{figure*}

In \cref{fig:smoothing} we visualize how our semi-parametric smoothing approach works. Since each raw measurement is an average cross entropy over $ 1024 \mul 100 $ examples, it is naturally a noisy estimate of the ``true'' cross entropy. Our smoothing strategy ensures the appropriate monotonicity properties are enforced, while matching the overall trend as closely as possible.

\section{Caveats on Privacy Calibration}
Throughout the work, we have assumed that hyperparameter choices for model training are made against a \textit{fixed} privacy budget. 
In particular, we assume the common scenario in which the model trainer fixes an $(\epsilon, \delta)$-budget and then utilises a \textit{privacy calibration} algorithm to choose DP-SGD hyperparameter combinations (sampling probability, training iterations and noise scale) which satisfy this privacy budget. 
Note that in the main manuscript, we express this choice in terms of the \nbr $\sigma$ and the number of iterations $\iters$, but this is merely a matter of notation.
As also noted in the preceding subsection, the choice of sampling probability (and thus the resulting batch size) play an important role in determining the final model's cross-entropy.
As described in the recent work of \citet{kaissis2024beyond}, calibrating against a fixed $(\epsilon, \delta)$-budget while varying DP-SGD hyperparameters must be done with care: 
In brief, one cannot assume that DP-SGD with different hyperparameters offers the same privacy guarantees \textit{despite} having the same nominal $(\epsilon, \delta)$-budget.
This is due to the fact that the privacy guarantees of DP-SGD can only be adequately expressed through a \textit{privacy profile}, that is, a collection of $(\epsilon, \delta(\epsilon))$ tuples. 
In simple terms, two DP-SGD algorithms can share an $(\epsilon, \delta)$-budget, that is, offer the same privacy guarantees for a specific $\delta$ while offering (sometimes drastically) different privacy guarantees at a different value of $\delta$.
As also described in the aforementioned work, varying the sampling rate (and thus batch size) has a drastic impact on this difference in privacy guarantees.
The authors of the aforementioned work thus recommend reporting the \textit{excess vulnerability} that DP-SGD algorithms incur with respect to each other when they replace one another in a workflow. 
We refer to the aforementioned work for technical details.
Here, we demonstrate that meaningful differences can indeed arise between models calibrated to satisfy the same $(\epsilon, \delta)$-budget.

\begin{figure*}[htbp]
    \centering
    \begin{subfigure}[b]{0.32\textwidth}
        \centering
        \includegraphics[width=\textwidth]{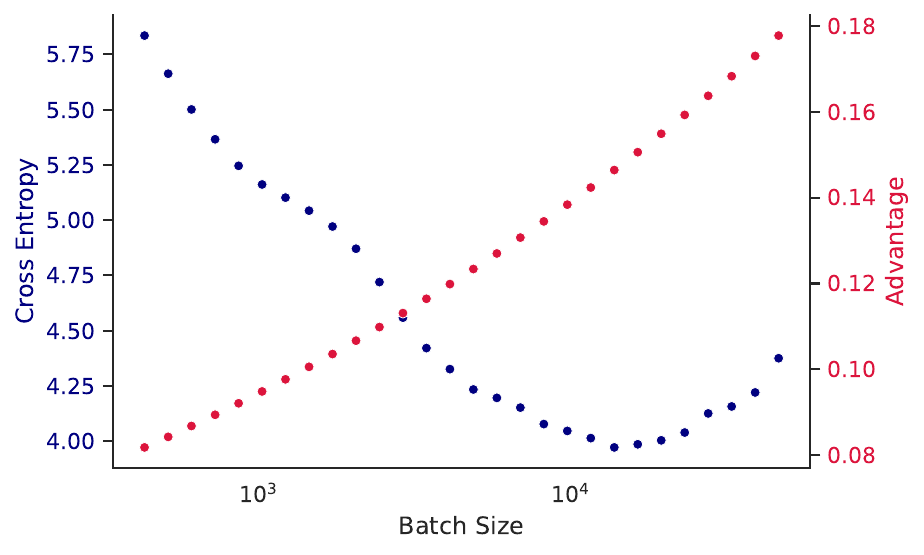}
        \caption*{(a)}
    \end{subfigure}
    \hfill
    \begin{subfigure}[b]{0.32\textwidth}
        \centering
        \includegraphics[width=\textwidth]{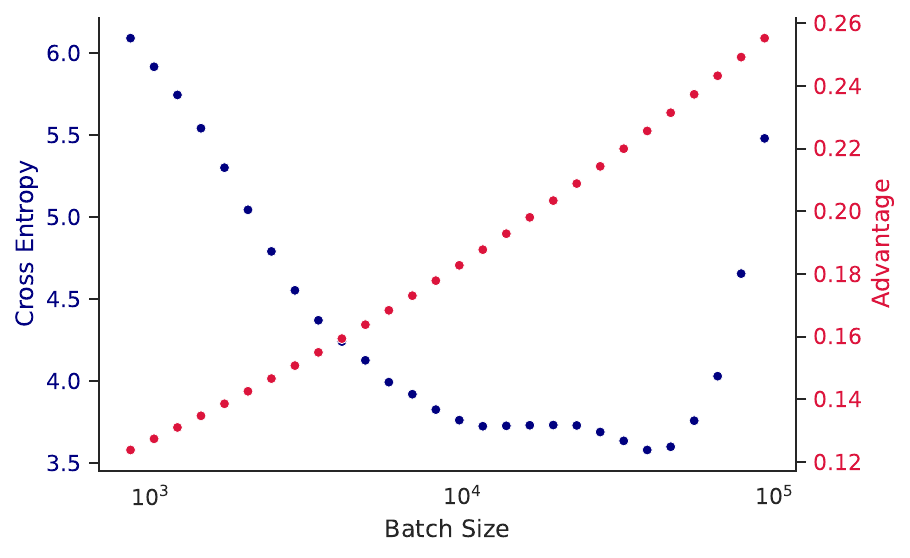}
        \caption*{(b)}
    \end{subfigure}
    \hfill
    \begin{subfigure}[b]{0.32\textwidth}
        \centering
        \includegraphics[width=\textwidth]{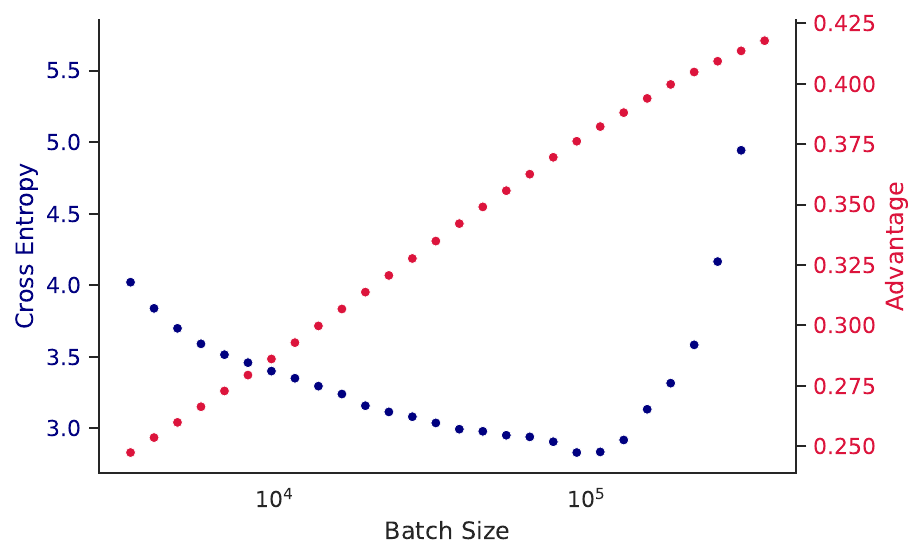}
        \caption*{(c)}
    \end{subfigure}
    \caption{
    Varying the batch size (horizontal axis, log-scale) has a drastic effect on excess vulnerability (measured as MIA advantage, {\textcolor{red}{red}}, right vertical axis) for models with a fixed compute budget and size and a fixed privacy budget of $(\epsilon, \delta) = (8, 10^{-8})$.
    (a): Compute budget: $6 \mul 10^{17}$, model size: $4\,000\,000$. (b) Compute budget: $6.3 \mul 10^{19}$, model size: $200\,000\,000$. (c) Compute budget: $2.5 \mul 10^{20}$, model size: $200\,000\,000$.
    The scaling-law-predicted cross-entropy is plotted on the left vertical axis in {\textcolor{blue}{blue}}.
    }
    \label{fig:excess-vulnerability}
\end{figure*}

Exemplarily, we fixed a privacy budget of $(\epsilon, \delta) = (8, 10^{-8})$ for specific fixed compute budgets and model sizes while varying the batch size (and adjusting the noise to maintain the privacy budget).
We then computed the scaling-law predicted cross-entropy and the vulnerability of the models against membership inference attack (MIA) adversaries measured in terms of MIA \textit{advantage} \citep{yeom2018privacy}.
We note that MIA advantage is a proxy metric for other attacks such as reconstruction attacks and is related to the $\Delta$-divergence which quantifies vulnerability as described in \citet{kaissis2024beyond}.
\cref{fig:excess-vulnerability} demonstrates the phenomenon.

Note that in all three cases, it is possible to achieve virtually the same cross-entropy (blue, left vertical axis) while controlling the MIA advantage by judiciously choosing the batch size.
Conversely, it is also possible to incur an unduly high vulnerability without a substantial decrease (or sometimes even an increase) in cross-entropy through a poor choice of batch size. 
As an auxiliary finding, we note that the relationship between cross-entropy and batch size follows the trend observed in \citet{de2022unlocking}.
In brief, there is a \textit{Pareto optimal} batch size beyond which both the cross-entropy \textit{and} the excess vulnerability can only become worse (larger).
We stress that the models shown here all satisfy the \textit{same nominal $(\epsilon, \delta)$-budget} but exhibit (substantial) differences in vulnerability against at least a subset of adversaries which may pass unnoticed if only reporting a single $(\epsilon, \delta)$-DP guarantee.
We thus recommend practitioners to monitor changes in excess vulnerability that may arise due to hyperparameter tuning and report them alongside the $(\epsilon, \delta)$-budget to which DP-SGD has been calibrated. 

\section{Parametric Scaling Laws}
\label{sec:parametric-sl}

Previous work on (non-private) LLM scaling laws use a fully parametric form to predict the cross entropy loss based on several key factors. For example, the ``Chinchilla'' scaling law~\citep{hoffmann2022training} can be parameterized as follows:
\begin{equation}
    \hat{L}(n_{\text{params}}, n_{\text{tokens}}) \triangleq E + \frac{A}{n_\text{params}^\alpha} + \frac{B}{n_\text{tokens}^\beta}
\end{equation}
In this section, we explore a similar methodology to fit a fully parametric form of scaling law in the setting of private training. Following the notation of this paper, we define a parametric form based on the following key factors: the model size $\modelsize$, the number of examples $\numsamples$ and the \nbr $\nbratio$. Note our notations are slightly different from \citet{hoffmann2022training}, and we use number of \emph{examples} instead of number of \emph{tokens} as it is a more relevant quantity in private training.

We consider several variations of parametric forms. The first one is a naive extension of the Chinchilla scaling law, by adding an additional term involving the \nbr:
\begin{equation}
    \hat{L}_\mathbf{1}(\modelsize, \numsamples, \nbratio) \triangleq E + \frac{A}{\modelsize^\alpha} + \frac{B}{\numsamples^\beta} + C\nbratio^{\gamma}
    \label{eq:parametric-sl-naive}
\end{equation}
We did not put $\nbratio^\gamma$ in the denominator because the loss increases with the \nbr. Following \citet{hoffmann2022training}, we estimate the coefficients $(E, A, B, C, \alpha, \beta, \gamma)$ by minimizing the Huber loss~\citep{huber1992robust} between the predicted and the observed loss using the L-BFGS algorithm~\citep{nocedal1980updating}, and we try multiple different initializations and choose the best fit. We restrict the curve fitting data to only the subsets of data points with more than $100,000$ training iterations, \nbr larger than $5\times 10^{-7}$, and ignore points with very high cross entropy loss ($> 8$).

\begin{figure}
    \centering
    \includegraphics[width=.5\linewidth]{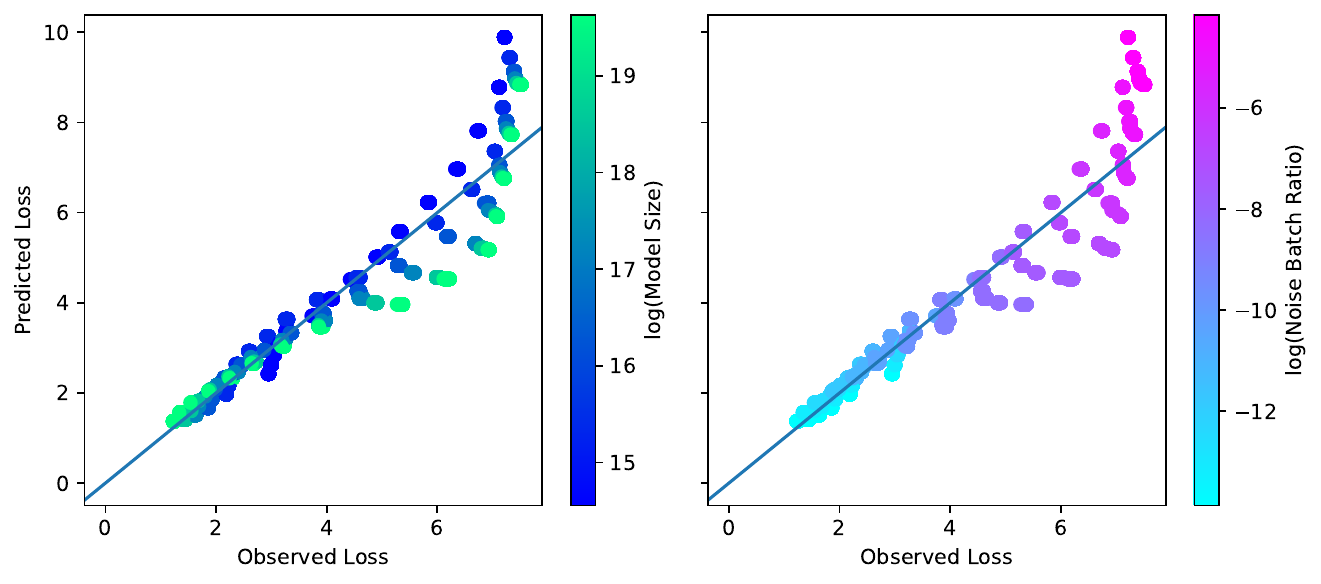}
    \caption{Parametric private scaling law of $\hat{L}_\mathbf{1}$ from \Cref{eq:parametric-sl-naive}. Optimal fit with $\alpha=0.71$, $\beta=12.87$, $\gamma=0.19$. The two pannels show the same plot of observed cross entropy loss against the predicted loss from the scaling law, except the data points are colored differerently, according to the model size and \nbr, respectively.}
    \label{fig:parametric-scaling-law-naive}
\end{figure}

\begin{figure}
    \centering
    \includegraphics[width=.5\linewidth]{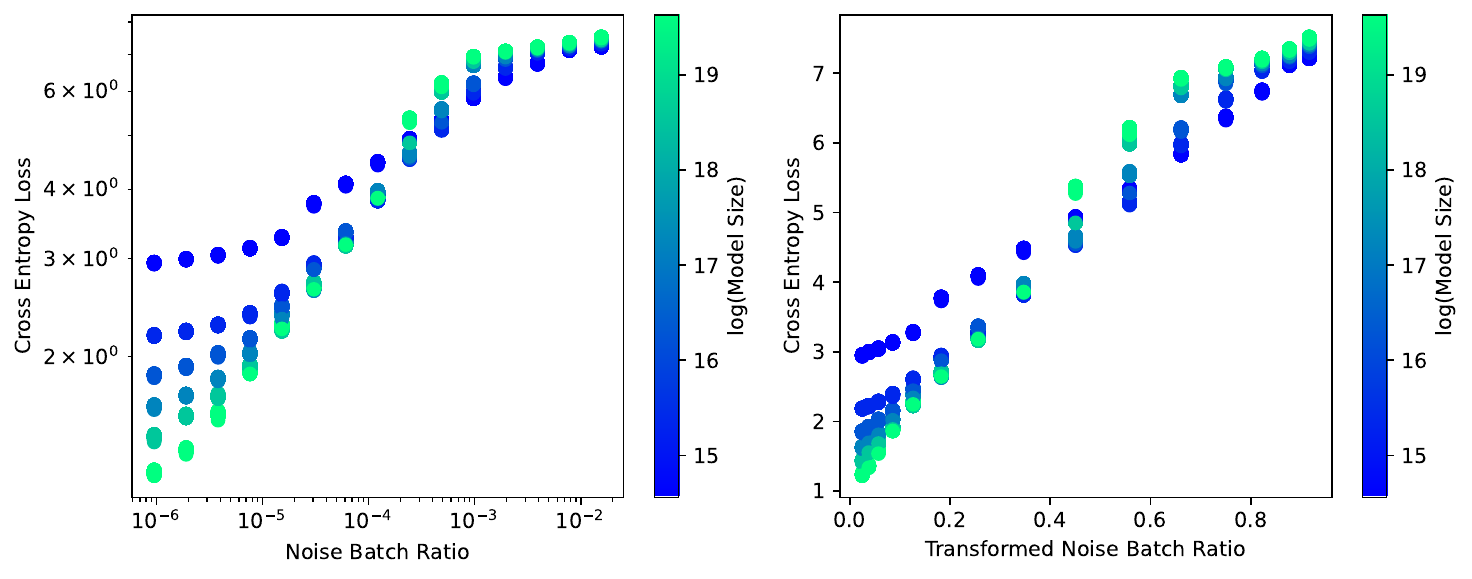}
    \caption{Relation between the \nbr and the cross entropy loss. \textbf{(left)} The data plotted in log-log scale. \textbf{(right)} The data plotted in linear scale, where the \nbr $\nbratio$ is transformed according to a simple rule in \Cref{eq:nbr-transform}.}
    \label{fig:parametric-scaling-law-nbr-transform}
\end{figure}

\Cref{fig:parametric-scaling-law-naive} visualize the optimal fit. We observe that the prediction is generally accurate for low loss value ranges. However, the prediction starts to diverge at high loss value ranges, corresponding to runs with high \nbr. This is partly due to the fact that the \nbr does not impact the loss in a log-linear fashion, as shown on the left panel of \Cref{fig:parametric-scaling-law-nbr-transform}. Therefore, the parametric form of \Cref{eq:parametric-sl-naive} cannot capture the relation accurately. Instead, we observe S-shaped curves in the log-log plot. To account for this, we apply a simple transform to the \nbr $\nbratio$:
\begin{equation}
    \nbratio_\looparrowright\triangleq
    \text{sigmoid}\left( \frac{\log(\nbratio) + 8}{1.6} \right)
    \label{eq:nbr-transform}
\end{equation}
The right panel of \Cref{fig:parametric-scaling-law-nbr-transform} shows an approximately linear relation after this transformation. Furthermore, we observe that the relation between the \nbr and the loss changes with the model sizes.

After incorporating those observations, we consider an alternative variant of private scaling law parameterization:
\begin{equation}
        \hat{L}_\mathbf{2}(\modelsize, \numsamples, \nbratio) \triangleq E + \frac{A}{\modelsize^\alpha} + \frac{B}{\numsamples^\beta} + \frac{C\nbratio_\looparrowright^{\gamma}}{\modelsize^{\alpha_2}}
    \label{eq:parametric-sl-sigmoid}
\end{equation}

\begin{figure}
    \centering
    \includegraphics[width=.5\linewidth]{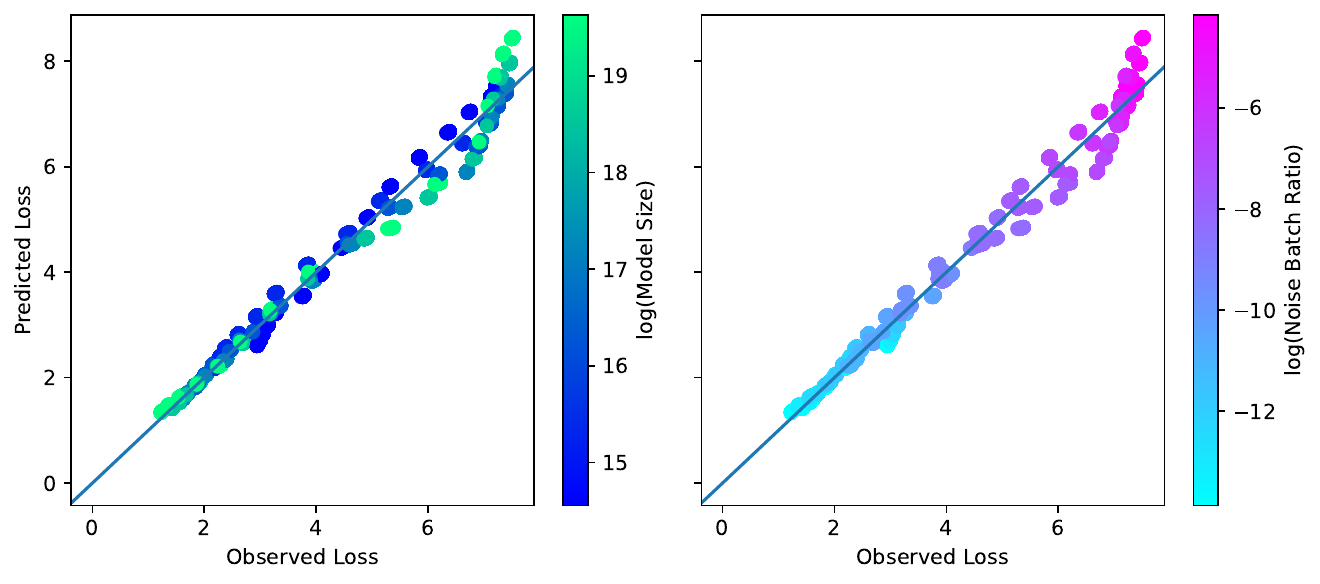}
    \caption{Parametric private scaling law of $\hat{L}_\mathbf{2}$ from \Cref{eq:parametric-sl-sigmoid}. Optimal fit with $\alpha=0.47$, $\beta=0.12$, $\gamma=0.95$, $\alpha_2=-0.07$. The two pannels show the same plot of observed cross entropy loss against the predicted loss from the scaling law, except the data points are colored differerently, according to the model size and \nbr, respectively.}
    \label{fig:parametric-scaling-law-sigmoid}
\end{figure}

The optimal fit according to this parameterization is shown in \Cref{fig:parametric-scaling-law-sigmoid}. We observe that the predicted loss matches with the observed loss better than the previous parameterization in \Cref{fig:parametric-scaling-law-naive}. 

\begin{figure}
    \centering
    \includegraphics[width=.5\linewidth]{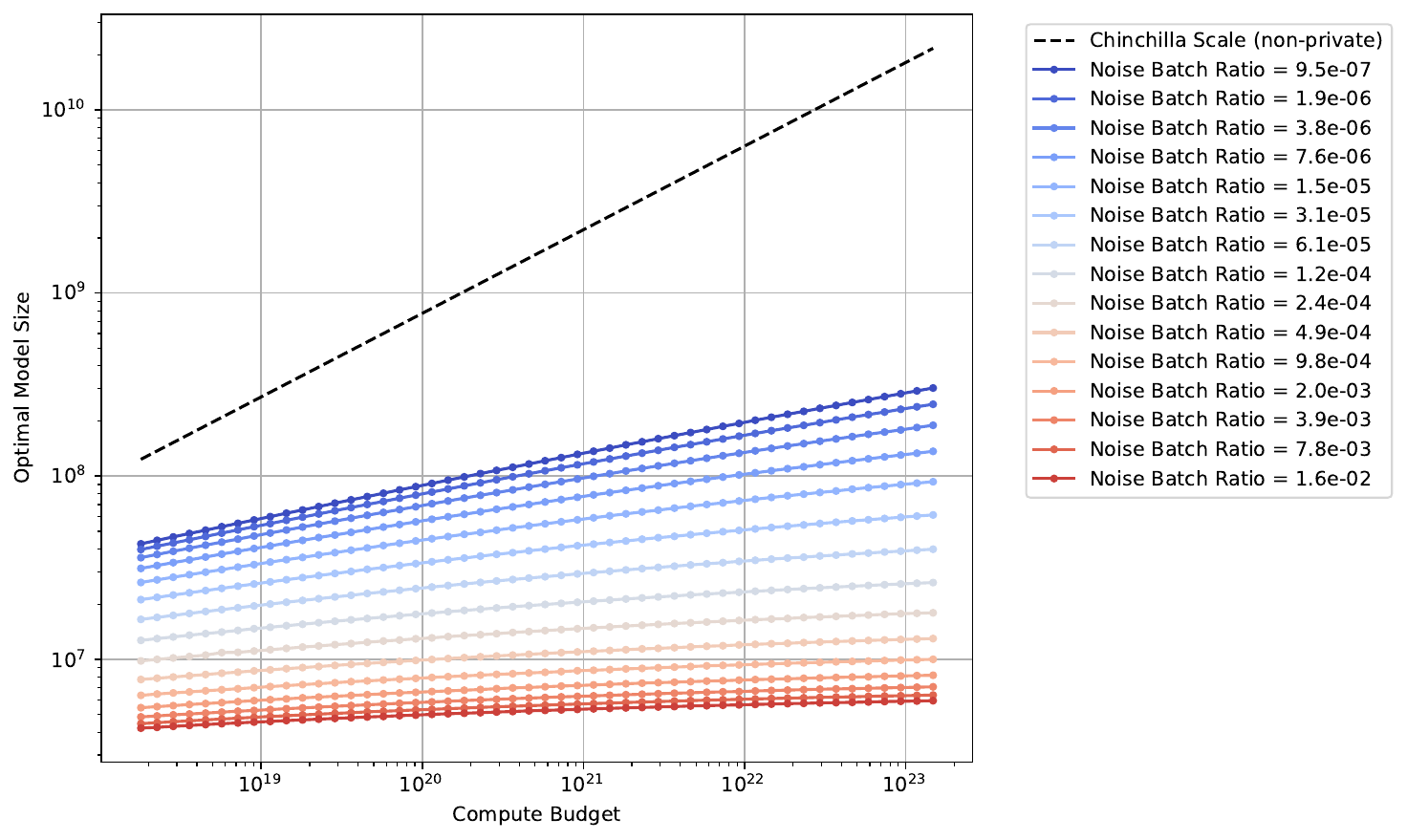}
    \caption{Optimal model sizes under according to the parametric private scaling law in \Cref{eq:parametric-sl-sigmoid}.}
    \label{fig:parametric-scaling-optimal-model-size}
\end{figure}

In the Chinchilla parameterization of scaling law for non-private LLMs, the optimal model size under a certain compute budget (approximately represented by $6n_\text{params}n_\text{tokens}$) can be directly solved and takes a power-law form~\citep[][Equation (4)]{hoffmann2022training}. In our case, the parameterization is more complicated, for a given compute budget and \nbr, we use \texttt{scipy.optimize.minimize\_scalar} to find the optimal model size that minimizes $\hat{L}_\mathbf{2}$. The results are plotted in \Cref{fig:parametric-scaling-optimal-model-size}. We observe that the slop is lower for curves with larger \nbr, indicating the challenges to scale model sizes under heavy DP noises. As the noise decreases, the curves shift up and the slopes increase, approaching towards the non-private Chinchilla scaling law shown in dashed line.

While a fully parametric scaling law can be easier to interpret and understand, as noted above, there is not a simple log-linear relation between the loss and the \nbr. Our sigmoid based transformation (and the coupling with the model size) improved the tightness of the fitting. But the transformation is not designed in a very principled way. As a result, we opt to use the semi-parametric fitting in \Cref{sec:exp_setup} in the main analysis of our results. We also leave the exploration of other alternative parametric fitting such as fitting a $\nbratio$-depending delta term on top of a non-private scaling law for future work.

\end{document}